%% file: main.tex
\documentclass[conference]{IEEEtran}
\IEEEoverridecommandlockouts
\usepackage{cite}
\usepackage{amsmath,amssymb,amsfonts}
\usepackage{algorithmic}
\usepackage{graphicx}
\usepackage{textcomp}
\usepackage{xcolor}
\usepackage{siunitx}
\usepackage{booktabs}
\usepackage{multirow}
\usepackage{soul}
\usepackage{balance}
\usepackage[top=0.75in, bottom=1in, right=0.625in, left=0.625in]{geometry}
\def\BibTeX{{\rm B\kern-.05em{\sc i\kern-.025em b}\kern-.08em
    T\kern-.1667em\lower.7ex\hbox{E}\kern-.125emX}}

\usepackage{pgfplots}
\pgfplotsset{compat=1.18}

\newcommand\inputpgf[2]{{
\let\pgfimageWithoutPath\pgfimage
\renewcommand{\pgfimage}[2][]{\pgfimageWithoutPath[##1]{#1/##2}}
\let\includegraphicsWithoutPath\includegraphics
\renewcommand{\includegraphics}[2][]{\includegraphicsWithoutPath[##1]{#1/##2}}
\input{#1/#2}
}}

\newcommand{\elia}[1]{{#1}}

\newcommand{\paolo}[1]{#1}

\begin{document}

\title{
Energy Minimization for Participatory Federated Learning in IoT Analyzed via Game Theory
}

\author{

\IEEEauthorblockN{Alessandro Buratto\IEEEauthorrefmark{1}, Elia Guerra\IEEEauthorrefmark{2}, Marco Miozzo\IEEEauthorrefmark{2}, Paolo Dini\IEEEauthorrefmark{2}, and Leonardo Badia\IEEEauthorrefmark{1}}

\IEEEauthorblockA{\IEEEauthorrefmark{1}Dept.\ of Information Engineering (DEI), University of Padova, Italy \\
email: \{alessandro.buratto.1@phd. , leonardo.badia@\}unipd.it}

\IEEEauthorblockA{\IEEEauthorrefmark{2}Centre Tecnol\`ogic de Telecomunicacions de Catalunya (CTTC/CERCA), Castelldefels, Spain \\
email: \{eguerra, marco.miozzo, paolo.dini\}@cttc.es
\vspace{-0.35cm}}
\thanks{This work was supported by the Italian PRIN project 2022PNRR ``DIGIT4CIRCLE,'' by the Spanish project PID2020-113832RB-C22(ORIGIN)/MCIN/AEI/10.13039/50110001103, European Union Horizon 2020 research and innovation programme under Grant Agreement No. 953775 (GREENEDGE) and the grant CHIST-ERA-20-SICT-004 (SONATA) by PCI2021-122043-2A/AEI/10.13039/501100011033}
}

\maketitle

\begin{abstract}
The Internet of Things requires intelligent decision making in many scenarios. To this end, resources available at the individual nodes for sensing or computing, or both, can be leveraged.
This results in approaches known as participatory sensing and federated learning, respectively.
We investigate the simultaneous implementation of both, through a distributed approach based on empowering local nodes with game theoretic decision making. A global objective of energy minimization is combined with the individual node's optimization of local expenditure for sensing and transmitting data over multiple learning rounds.
We present extensive evaluations of this technique, based on both a theoretical framework and experiments in a simulated network scenario with real data.
Such a distributed approach can reach a desired level of accuracy for federated learning without a centralized supervision of the data collector. 
However, depending on the weight attributed to the local costs of the single node, it may also result in a significantly high Price of Anarchy (from \(1.28\) onwards).
Thus, we argue for the need of incentive mechanisms, possibly based on Age of Information of the single nodes. 
\end{abstract}

\begin{IEEEkeywords}
Federated learning; Energy consumption; Participatory sensing; Internet of things; Game theory.
\end{IEEEkeywords}

\section{Introduction}
Participatory sensing, facilitated by the proliferation of mobile devices and the Internet of Things (IoT), has emerged as a promising paradigm for collecting large-scale data from end nodes and exploiting them for many applications \cite{guo2014participatory,mora21decentralised}. 
It implies a voluntary contribution of individuals to data provision, forming a distributed sensing network that harnesses user-generated data for diverse applications, such as environmental monitoring, urban planning, and healthcare \cite{michelusi2012operation,lanza2023urban,cisotto2018joint}. 

Federated learning \elia{(FL)} is a \paolo{distributed} approach to artificial intelligence, where devices collaboratively train a shared model with private data, so as to split the computational burden \cite{tran2019federated}. In context-aware applications, FL can be superimposed to sensing as a collaborative approach for efficient extraction of ambient information from data, leveraging the collective intelligence and computational power of a diverse set of devices to train machine learning models without a costly centralization \cite{cruz2020understanding}.

Leveraging the distribution of training tasks and shifting the computation from \paolo{cloud} servers to local \paolo{edge} devices reduces the energy consumption associated with transmitting large amounts of data to a central server for processing \cite{gindullina2020energy}. As the learning process takes place directly on \paolo{or close to} the participating devices, energy requirements are decreased \cite{song2019pushing}. 

Thus, we investigate a combined paradigm of data management, where both sensing and extrapolation of meaning from sensed data are performed by the end nodes, and the role of the network coordinator is to merge the local decision process. Such an integration of participatory sensing and FL can be analyzed through the lens of game theory \cite{buratto2023game,tu2022incentive}, whose main purpose is to study the strategic interactions among self-interested entities. This can shed light on the resulting system dynamics and highlight relevant trade-offs.

We consider a scenario, where $N$ end nodes are connected to a central data sink that coordinates their federation. Each end node has access to a \emph{private} dataset, either sensed directly from the environment, or assigned by the data sink as a partition of a global dataset, and partake in the learning process on a \emph{voluntary} basis \cite{dasari2020game,cong2020game}.
Their decision about whether to participate or not is iterated through multiple rounds with a probabilistic approach \cite{zhang2022communication}. 
We assume all nodes to behave identically and not exchange information with each other, thus they have identical and independently distributed (i.i.d.) probability $p_i$ to participate in each round, computed locally. This can be extended to correlated/communicating nodes along the lines of \cite{buratto2023optimizing}.

The common approach is to handle the participation in a centralized fashion at the sink \cite{luo2021cost}, which also assigns the data to the end nodes for local training and handles the federation by merging the resulting parameters. Even in this context, it would be convenient to alternate the participation of nodes over multiple rounds to obtain faster convergence of the FL, avoiding overfitting or entrapment, which would unnecessarily increase the duration of the training as well as cause overconsumption of energy 
\cite{sun2020energy,wang2022threats}. 
Thus, an optimal probability of participation for the single node can be computed, which minimizes the time to reach a target accuracy of the FL.

In a fully distributed and participatory context, which is the approach of choice if nodes perform their measurements locally \cite{mora21decentralised}, we can consider that nodes act based on their selfish objective, individually computed, minimizing a linear combination of duration and cost.
This implementation resounds as a \emph{Tragedy of the Commons} scenario \cite{prospero2021resource}, since in large IoT systems the end nodes are expected to realize that their individual contribution gives little benefit to the overall duration of the task, and opt for reduced action. This implies an overall suboptimal participation rate throughout the entire network, and a high Price of Anarchy (PoA) \cite{donahue2021optimality}.

In this spirit, game theory may allow us to exploit participatory sensing and FL to their fullest. Strategic interactions and incentives of participants can ensure effectiveness and energy sustainability of collaborative systems \cite{cappelletti2021game} and our study paves the way for designing less power-hungry data-driven applications that benefit from the collective intelligence of diverse participants.

The rest of this paper is organized as follows. In Section \ref{sec:relw}, we discuss related work. Section \ref{sec:system_model} describes the system model and discusses our proposed approach to combine federated learning with participatory sensing.
We show numerical results in Section \ref{sec:results} and we finally conclude in Section \ref{sec:concs}.

\section{Related Work}
\label{sec:relw}

Development of federated learning strategies and minimization of energy consumption are tight-knit goals for future networks exploiting the intelligence available at the end nodes \cite{gindullina2020energy}. For example, \cite{sun2020energy} proposes an online energy-aware dynamic
worker scheduling policy, which maximizes the average number
of workers scheduled for gradient update at each iteration under
a long-term energy constraint.
In \cite{yang2020energy}, the reasoning
about the energy consumption being related to the use computation resources is also extended to include the wireless communication exchanges.
Such proposals compare well with what we consider to be the optimal centralized allocation in the following.

At the same time, participatory sensing and federated learning are scenarios where the actions of individual agents mutually influence the outcome, and it makes sense to invoke an application of game theory \cite{dasari2020game,donahue2021optimality}.
In this context, the shared goal of the activity is at odds with the individual cost paid by the users, which can be declined in multiple ways, e.g., to security and privacy \cite{giaretta2021pds} of data owners.

We argue that the quintessential motivation behind distributed approaches to sensing and learning lies in optimizing the resource usage \cite{luo2021cost}, simultaneously aiming to save communication overhead and energy \cite{guerra2023cost}. 

In \cite{tu2022incentive}, the problem of recruiting participants for a federated learning is studied from a game theoretic perspective. The starting point is similar to what we argue in this paper, that is, rational data owners may be unwilling to participate in a collaborative learning process due to excessive resource consumption. Therefore, the authors propose some incentives for their participation designed through game theoretic mechanisms. However, the paper takes a high level perspective, where the motivation behind the individual nodes requiring incentives lies more in their selfishness towards an economic advantage. Instead, we consider this to be more directly related to energy consumption, which in our opinion gives a stronger justification even to well-meaning data owners.

The investigation of incentive design is also traditionally approached through auctions, capturing the supply and demand interaction \cite{cong2020game}. Similarly, \cite{buratto2023game} explores the Nash equilibria (NEs) of a participatory intervention of the nodes minimizing age of information (AoI).

Differently from these contributions, we argue that the problems of participation in federated learning are not just related to the interaction of data exchange and resource utilization, but more fundamentally rooted in the energy consumption \cite{song2019pushing}. To obtain a factual energy saving from a participatory paradigm, the objective must be shared by all nodes, and the efficiency of strategic choices must be evaluated from this standpoint, i.e., to see whether uncoordinated nodes can harmonize their operation, without unnecessary energy inefficiencies.

Despite the importance of participatory federated learning, this particular issue was scarcely addressed in the literature, which is why our contribution fills a gap and validate such an approach for IoT scenarios.

\section{System Model}
\label{sec:system_model}
We consider an FL scenario with a set \elia{$\mathcal{N} = \{1 \dots N\}$} of nodes that need to collaborate to the same learning task, communicating their model weights to a central receiver that acts as a sink. The receiver is in charge of collecting the weights from the participating nodes and to combine them, to obtain a global training model. 
\elia{In particular, we consider the FedAvg algorithm~\cite{mcmahan2017communication} where each node $i \in \mathcal{N}$ trains for $E$ local epochs with a loss function $\ell_i$ on its local dataset the global model and then shares the model updates with the central server via an IEEE 802.11ax wireless link. In the algorithm formulation the receiver, i.e, a central server, is in charge of selecting the subset of clients that participate in the current round.}

\elia{We give a special twist on this scenario,} giving to each node the ability to control its own probability $p_i$ to participate in the task. Differently from approaches where a globally optimal involvement is sought for the end nodes \cite{zhang2022communication,luo2021cost}, here the participation probability is computed by each node individually, based on local evaluations about the time for task completion and energy consumption. This probability is set a priori and cannot be changed later on. The task is successful when a target \elia{validation} accuracy is reached at the receiver's side. At every round, each node decides whether to participate in the learning task according to the local probability. Thus, only a subset of nodes \elia{$\mathcal{P}^t$} participates in learning round $t$.
\elia{Starting from the model in~\cite{guerra2023cost}, we evaluate the energy consumption of each node depending on the decision to participate in FL round $t$ or not.
If client $i$ participates, the energy spent for training is:}
\begin{equation}
    \mathcal{E}_{{\rm train}, i}^t = P_{{\rm hw},i}^t T_{{\rm train},i}^t,
    \label{eq:energy_train}
\end{equation}
\elia{where $P_{{\rm hw,i}}^t$ is the average power drained by the hardware, i.e., CPU, GPU, and DRAM, and $T_{{\rm train}, i}^t$ is the training process duration. The energy spent for communication is}
\begin{equation}
    \mathcal{E}_{\rm tx} = P_{\rm tx} T_{\rm tx},
    \label{energy_tx}
\end{equation}
where $P_{\rm tx}$ and $T_{\rm tx}$ are transmission power and time, respectively. Note that $\mathcal{E}_{\rm tx}$ is the same for each client in each FL round since the size of the model updates is constant during the whole learning process.
Finally, all participating nodes wait for the conclusion of the current FL round. Let $T_{\rm round}$  be the maximum round duration defined by the sink. All participating devices have to start the upload process within $T_{\rm round}$, otherwise their contribution will be discarded. The idle energy is computed as
\begin{equation}
    \mathcal{E}_{{\rm idle}, i}^t = P_{{\rm idle},i}(T_{\rm round}-T_{{\rm train}, i}^t),
    \label{eq:energy_idle}
\end{equation}
\elia{with $P_{{\rm idle}, i}$ the power consumed while idling and $T_{\rm round}-T_{{\rm train}, i}^t$ the idling time. The total energy is given by}
\begin{equation}
    \mathcal{E}_{i}^t = \mathcal{E}_{{\rm train}, i}^t + \mathcal{E}_{\rm tx} + \mathcal{E}_{{\rm idle}, i}^t.
\end{equation}
\elia{If a node $j$ does not join the current FL round its energy consumption is given by the idle energy} 
\begin{equation}
    \mathcal{E}_{j} = P_{{\rm idle}, j}T_{\rm round} 
\end{equation}
\elia{since the node is waiting for the conclusion of the round.}
\elia{The total energy consumption is}
\begin{equation}
    \mathcal{E}^t = \sum_{i \in \mathcal{P}^t} \mathcal{E}_{i}^t + \sum_{j \in \mathcal{N} \setminus \mathcal{P}^t} \mathcal{E}_{j}.
\end{equation}
\elia{The energy consumed for $d$ rounds is} 
\begin{equation}
    \mathcal{E} = \sum_{t = 1}^{d} \mathcal{E}^t.
    \label{eq:energy_run}
\end{equation}

The energy consumption is minimized for the lowest duration $d$ of the FL task, which is concluded when a target value for the validation accuracy is consistently met \cite{yang2020energy}. This is dependent on the mean number of nodes participating in a given round, as we will later argue in Section \ref{sec:results}. The probability that a given number of nodes are active at any given time follows a Poisson-Binomial distribution, because participation of each node is independent of the others. The learning task duration follows a Poisson-Binomial distribution $\mathcal{D} \sim PoiBin(p_i)$, $i{\in} \mathcal{N}$ and the duration value $d(k)$ is a function of the number of participating nodes $k$. It follows from probability theory that the first moment of this distribution is calculated as
\begin{equation}\label{eq:exp_value}
    \mathbb{E} [\mathcal{D}] = \sum_{i=0}^{N} d(i) \cdot P[m=i] \text{,}
\end{equation}
where $P[m=i]$ is the probability that exactly $i$ nodes participate, computed in closed form as \cite{fernandez2010closed}
\begin{IEEEeqnarray}{rCl}\label{eq:pmf}
    P[m] &=& \frac{ \displaystyle\sum_{n=0}^{N} \bigg\{ e^{\frac{-j2 \pi n m} {N+1}} \prod_{k=1}^{N} [p_{k}(e^{\frac{j2 \pi n}{N + 1}} {-} 1){+} 1] \bigg\}}
    {N+1} \text{.}
\end{IEEEeqnarray}

We formalize this distributed optimization task as a static game of complete information $\mathcal{G=\{\mathcal{N}, \mathcal{A}, \mathcal{U}\}}$, where $\mathcal{N}$ is the set of players, i.e., the nodes in the network, $\mathcal{A}$ is the set of actions, namely the participation probability $p_i \in [0,1]$ for each player $i$, and $\mathcal{U}$ is the set of the utilities for each player \cite{cappelletti2021game}. 
In order to promote client participation, we implement an incentive mechanism based on the expected AoI of the single node, computed as the ratio of the second order moment and two times the first order moment of the inter-participation time $Y$ as a function of the node's participation probability.
\begin{equation}\label{eq:aoi}
    \mathbb{E}[\delta_i] = \frac{\mathbb{E}[Y^2]}{2 \mathbb{E}[Y]} = \frac{1}{p_i} - \frac{1}{2} \text{.}
\end{equation}
We define the utility of each player $i$ as 
\begin{equation}\label{eq:utility}
    u_i = - \mathbb{E} [\mathcal{D}] - \gamma \log(\mathbb{E}[\delta_i]) - c  p_i \text{,}
\end{equation}
where $\mathbb{E} [\mathcal{D}]$ follows \eqref{eq:exp_value}, $\log(\mathbb{E}[\delta_i])$ is the natural logarithm of \eqref{eq:aoi}, $\gamma$ is the weight assigned to the incentive and $c$ is a cost factor taking into account the energy consumption for the node when it participates. \elia{This equation combines the two goals of the clients: minimizing the number of rounds to reach convergence and the associated participation cost \cite{michelusi2012operation}.} It furthermore rewards the clients when they decide to participate actively in the federated learning task.

The NE is found through a one-sided optimization of the utility, which implies that each player takes a \textit{best response} to the unchanged actions of the others \cite{buratto2023game}. Thus, we solve the system of differential equations
\begin{IEEEeqnarray}{rCl}\label{eq:ne}
    \frac{\partial u_i}{\partial p_i} = 0 \text{, } \quad i=1, \dots, N
\end{IEEEeqnarray}
which can be obtained in closed form from \eqref{eq:exp_value}, \eqref{eq:pmf} and \eqref{eq:aoi}. Due to symmetry, this will result in the same value $p$ for all nodes, i.e., $p = p_1 = \ldots = p_N$.

We further calculate the Price of Anarchy (PoA) to evaluate how much a decentralized solution deteriorates the optimal centralized one's performance. The PoA is calculated as \cite{prospero2021resource}
\begin{equation} \label{eq:poa}
    PoA = \frac{\max_{s \in \text{NE}}u(s)}{\min_{s \in \mathcal{S}} u(s)}
\end{equation}
where, at the numerator, we take strategy $s$ from the set of all NEs with the highest cost, and at the denominator, the optimal centralized strategy that minimizes the cost.

\begin{table}[t]
    \resizebox{\columnwidth}{!}{  
    \begin{tabular}{c|ccc}
        \toprule
           & \textbf{Parameter} & \textbf{Description} & \textbf{Value} \\
         \midrule
          \multirow{9}{*}{\rotatebox[origin=c]{90}{Fed. Learning }} 
          & $|w|$ & Number of model parameters & $\num{11181642}$ \\
          & $S_{w}$ & ResNet-18 model parameters size$\!\!\!\!$ & $44.73$ MB \\
          & $\eta$ & Learning rate & $0.01$ \\
          & $N$ & Number of total clients & $50$ \\
          & $E$ & Local epochs number & $5$ \\
          & $T_{\rm round}$ & Maximum training time & $10$s \\
          & $\ell_i$ & Local loss function & $\!\!\!\!$Sparse Cat.\ Crossentropy\\
          & $T_{\rm acc}$ & Target accuracy on CIFAR-10$\!\!\!\!$ & $0.73$\\
          & $P_{\rm idle}$ & Idle power consumption & $96.85$ W\\
         \midrule
         \multirow{17}{*}{\rotatebox[origin=c]{90}{Communication (IEEE 802.11ax)}} 
         & $P_{\rm{tx}}$ & Tx power for edge devices & $9$ dBm \\
         & $\sigma_{\text{leg}}$ & Legacy OFDM symbol duration & $4$ \textmu s \\
 		 & $N_{\rm sc}$ & Number of subcarriers ($20$ MHz)$\!\!\!\!$ & $234$ \\
		 & $N_{\rm ss}$ & Number of spatial streams & $1$ \\
         & $T_{\rm{e}}$ & Empty slot duration & $9$ \textmu s \\
		 & $T_{\rm{SIFS}}$ & SIFS duration & $16$ \textmu s \\ 
		 & $T_{\rm{DIFS}}$ & DIFS duration & $34$ \textmu s \\ 
		 & $T_{\rm{PHY}}$ & Preamble duration & $20$ \textmu s \\
		 & $T_{\rm{HE-SU}}$ & HE single-user field duration & $100$ \textmu s \\
		 & $L_{s}$ & Size OFDM symbol & $24$ bits \\ 
		 & $L_{\rm{RTS}}$ & Length of an RTS packet & $160$ bits \\ 
		 & $L_{\rm{CTS}}$ & Length of a CTS packet & $112$ bits \\
 		 & $L_{\rm{ACK}}$ & Length of an ACK packet & $240$ bits \\ 
		 & $L_{\rm{SF}}$ & Length of service field & $16$ bits \\ 
		 & $L_{\rm{MAC}}$ & Length of MAC header & $320$ bits \\
		 & $\text{CW}$ & Contention window (fixed) & $15$ \\
        \bottomrule
    \end{tabular}}
    \vspace{1mm}
    \caption{Simulation parameters.}
    \label{tab:sim_param}
    \vspace{-0.5cm}
\end{table}

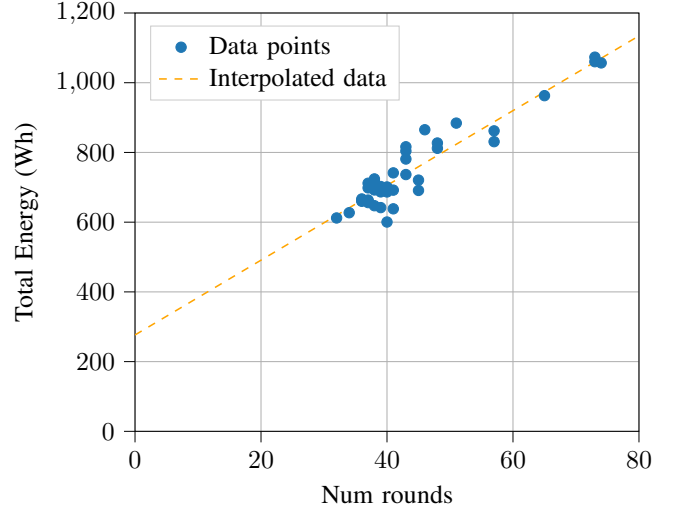
\begin{figure}[!t]
    \hspace{-0.5cm}
    \resizebox{1.06\columnwidth}{!}{
        \input{plots/scatterplot_with_interpolation.pgf}
    }
    \vspace{-0.6cm}
    \caption{Total energy spent $(\mathcal{E})$ vs. number of rounds to converge $(d)$.}
    \label{fig:rounds_vs_energy}
\end{figure}

\section{Performance Evaluation}
\label{sec:results}

\subsection{Experimental Setup}
\elia{We consider a scenario with $N=50$ nodes that collaboratively learn ResNet-18~\cite{he2016deep}, to correctly classify the images in the CIFAR-10~\cite{krizhevsky2009learning} dataset, containing $\num{50000}$ training samples that are randomly but fairly divided across all nodes. The input features are $32{\times}32$ colored images divided into $10$ classes. To evaluate the global model, we extract $7000$ samples as a validation set from the test partition of CIFAR-10. ResNet-18 $(|w|)$ has  $\num{11181642}$ trainable parameters, which require $44.73\text{MB}$ $(S_w)$ if stored as \emph{float32}. In every round, each client executes $E=5$ local epochs of Stochastic Gradient Descent using sparse categorical accuracy as local loss function $(\ell_i)$. Convergence is reached when the global model validation accuracy $(T_{\rm acc})$ is $\geq 0.73$ for $3$ consecutive rounds to avoid performance spikes. When the training is concluded, each node shares the model update with the central server using transmission power $(P_{\rm tx}) {=} 9~\text{dBm}$. Simulation parameters are reported in Table~\ref{tab:sim_param} and a comprehensive description of the communication model can be found in~\cite{guerra2023cost}. Experiments run on a server equipped with two Intel Xeon 6230, $188$ GB of RAM, and an RTX 2080 Ti.}

\begin{table}[t!]
    \begin{tabular}{c|rc|}
        \toprule
        $p_i$ & $\mathcal{E}$ & $d$ \\ \midrule
        0.100 & 1056.81 & 74 \\ 
        0.125 & 1060.25 & 73 \\ 
        0.130 & 830.90 & 57 \\ 
        0.150 & 1073.33 & 73 \\ 
        0.160 & 962.90 & 65 \\ 
        0.175 & 600.42 & 40 \\ 
        0.200 & 861.87 & 57 \\ 
        0.225 & 691.04 & 45 \\ 
        0.250 & 638.27 & 41 \\ 
        0.300 & 720.66 & 45 \\ 
        0.350 & 641.78 & 39 \\ 
        0.400 & 691.90 & 41 \\ 
        0.410 & 811.87 & 48 \\ 
        0.420 & 647.21 & 38 \\ 
        0.430 & 736.57 & 43 \\ 
        0.440 & 686.69 & 40 \\ 
        0.450 & 827.07 & 48 \\ 
        0.460 & 884.16 & 51 \\ 
        0.470 & 698.03 & 40 \\ 
        0.480 & 700.97 & 40 \\ 
        0.490 & 686.84 & 39 \\ 
        0.500 & 689.25 & 39 \\ 
        0.510 & 656.18 & 37 \\ 
        0.520 & 660.68 & 37 \\ 
        0.530 & 663.44 & 37 \\ 
        0.540 & 702.24 & 39 \\ 
        0.550 & 741.38 & 41 \\ 
        0.560 & 781.14 & 43 \\ 
        0.570 & 692.42 & 38 \\ 
        0.580 & 659.89 & 36 \\ 
        0.590 & 662.56 & 36 \\ 
        0.600 & 627.10 & 34 \\ 
        0.610 & 666.57 & 36 \\ 
        0.620 & 707.24 & 38 \\ 
        0.630 & 804.00 & 43 \\ 
        0.640 & 865.10 & 46 \\ 
        0.650 & 716.03 & 38 \\ 
        0.660 & 698.39 & 37 \\ 
        0.670 & 816.24 & 43 \\ 
        0.680 & 724.07 & 38 \\ 
        0.690 & 612.04 & 32 \\ 
        0.700 & 711.64 & 37 \\ 
        \bottomrule
\multicolumn{3}{c}{(a) One single seed}

    \end{tabular}
    \hfill
        \begin{tabular}{|c|crrr}
        \toprule
        $p_i$& $\overline{d}$ & $\sigma(d)$ & $\overline{\mathcal{E}}$ & $\sigma(\mathcal{E})$ \\ \midrule 
        0.100 & 74.50 & 11.47 & 1072.14 & 123.43 \\ 
        0.125 & 68.00 & 13.09 & 1005.97 & 140.49 \\ 
        0.130 & 56.00 & 5.29 & 862.84 & 60.19 \\ 
        0.150 & 62.50 & 8.81 & 950.26 & 100.14 \\ 
        0.160 & 57.25 & 6.13 & 887.80 & 61.31 \\ 
        0.175 & 51.00 & 9.42 & 797.18 & 145.67 \\ 
        0.200 & 51.00 & 4.55 & 816.96 & 37.86 \\ 
        0.225 & 45.50 & 3.70 & 747.44 & 54.52 \\ 
        0.250 & 51.00 & 9.56 & 803.96 & 132.64 \\ 
        0.300 & 46.75 & 2.75 & 768.25 & 41.50 \\ 
        0.350 & 43.00 & 5.23 & 724.40 & 73.21 \\ 
        0.400 & 43.25 & 2.22 & 734.25 & 33.22 \\ 
        0.410 & 44.50 & 5.32 & 758.88 & 62.29 \\ 
        0.420 & 42.75 & 4.11 & 725.76 & 59.45 \\ 
        0.430 & 42.75 & 3.30 & 734.69 & 35.41 \\ 
        0.440 & 43.00 & 4.08 & 732.95 & 49.07 \\ 
        0.450 & 43.50 & 4.43 & 751.96 & 61.11 \\ 
        0.460 & 42.75 & 5.56 & 750.14 & 89.77 \\ 
        0.470 & 39.50 & 3.11 & 698.25 & 33.15 \\ 
        0.480 & 39.25 & 6.70 & 696.30 & 71.74 \\ 
        0.490 & 40.67 & 2.89 & 709.99 & 33.48 \\ 
        0.500 & 40.00 & 0.82 & 704.10 & 11.11 \\ 
        0.510 & 41.75 & 3.30 & 719.96 & 43.71 \\ 
        0.520 & 42.50 & 7.33 & 729.13 & 81.90 \\ 
        0.530 & 40.00 & 3.16 & 703.01 & 37.23 \\ 
        0.540 & 41.75 & 4.27 & 726.11 & 44.34 \\ 
        0.550 & 39.50 & 2.65 & 706.41 & 35.12 \\ 
        0.560 & 40.25 & 2.99 & 719.03 & 48.51 \\ 
        0.570 & 40.50 & 4.43 & 712.93 & 46.15 \\ 
        0.580 & 46.25 & 14.15 & 771.83 & 152.41 \\ 
        0.590 & 39.00 & 2.58 & 694.74 & 27.70 \\ 
        0.600 & 39.00 & 4.24 & 691.24 & 51.19 \\ 
        0.610 & 37.75 & 2.87 & 682.34 & 30.05 \\ 
        0.620 & 39.75 & 5.56 & 708.59 & 58.31 \\ 
        0.630 & 37.75 & 3.50 & 697.93 & 70.71 \\ 
        0.640 & 39.75 & 5.91 & 726.61 & 102.68 \\ 
        0.650 & 39.00 & 2.16 & 702.75 & 23.75 \\ 
        0.660 & 40.75 & 4.99 & 719.79 & 48.48 \\ 
        0.670 & 40.00 & 4.69 & 725.12 & 75.90 \\ 
        0.680 & 41.25 & 4.03 & 728.89 & 36.60 \\ 
        0.690 & 37.50 & 3.87 & 676.75 & 45.17 \\ 
        0.700 & 38.25 & 5.50 & 696.29 & 59.19 \\  
\bottomrule
                \multicolumn{5}{c}{(b) Average results}
    \end{tabular}
    \vspace{1.5mm}
    \caption{Number of rounds $(d)$ and energy $(\mathcal{E})$ in Wh spent to converge with different participation probabilities.}
    \label{tab:results}
    \vspace{-3mm}
\end{table}

\subsection{Numerical Results} 
We now report the results of experiments and simulations.
\elia{To obtain a realistic model for FL from the perspective of game theory, we executed a wide array of simulations. First, we measured the energy consumption $(\mathcal{E})$ and the number of rounds to reach convergence for several participation probabilities $(p_i\in[0.1, 0.7])$ on the scenario described before. Table~\ref{tab:results}(a) shows the number of rounds to reach convergence $(d)$ and the total energy consumption $(\mathcal{E})$. To assess the training energy $(\mathcal{E}_{\rm train})$, we use Codecarbon~\cite{codecarbon} a Python library that estimates the hardware power consumption as per~\eqref{eq:energy_train}. The results show that low participation probabilities, i.e., $p_i \in [0.15, 0.3]$ do not lead to good performance. The best performance is achieved by $p_i = 0.69$ that converges in $32$ rounds, with $612.04$~Wh of energy spent. Fig.~\ref{fig:rounds_vs_energy} shows an approximately linear trend between $d$ and $\mathcal{E}$, justifying the assumption of Section~\ref{sec:system_model}.}

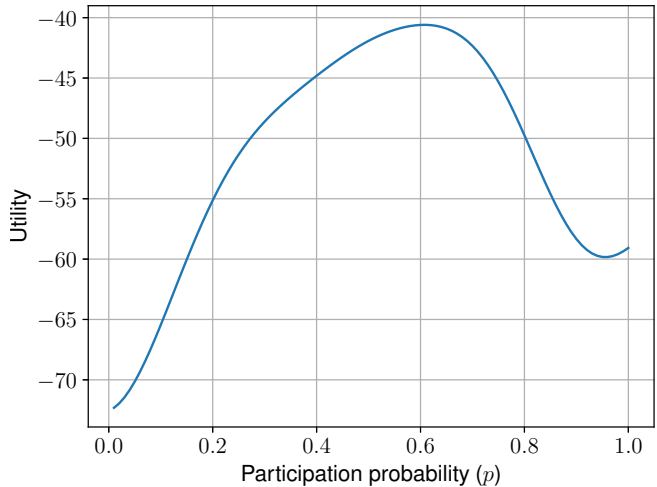
\begin{figure}[t]
    \centering
    \resizebox{\columnwidth}{!}{
        \input{plots/utility.pgf}
    }
    \vspace{-0.5cm}
    \caption{Utility from a fit of the FL simulation, applying \eqref{eq:utility} with $c=0$.}
    \label{fig:utility}
\end{figure}

\begin{figure}[t]
    \resizebox{\columnwidth}{!}{
        \includegraphics{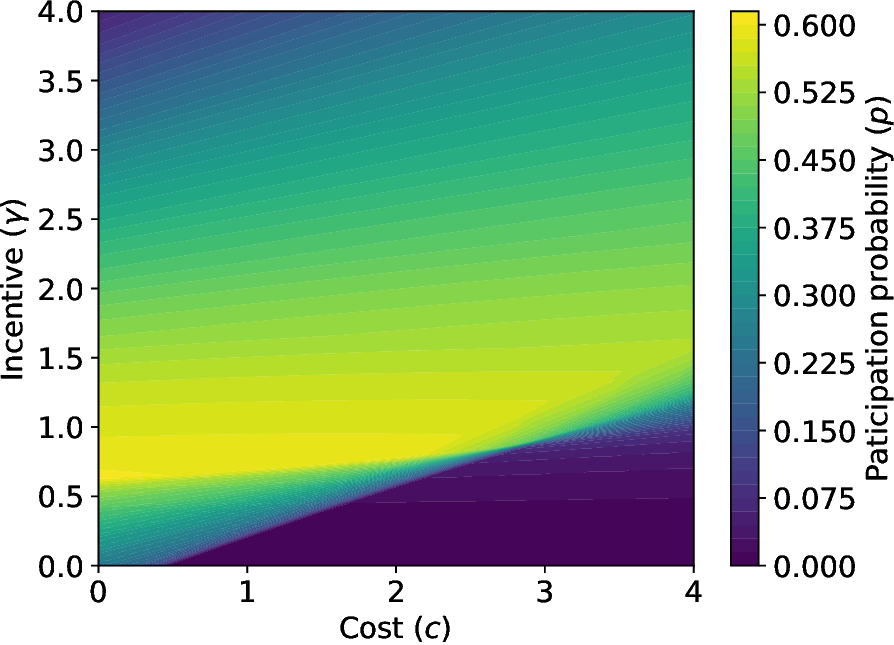}
    }
        \vspace{-0.5cm}
    \caption{NE solution of the participation probability for various cost factors $c$ and incentive weights $\gamma$.}
    \label{fig:3dne}
\end{figure}

\begin{figure}[t]
    \centering
    \resizebox{\columnwidth}{!}{
        \input{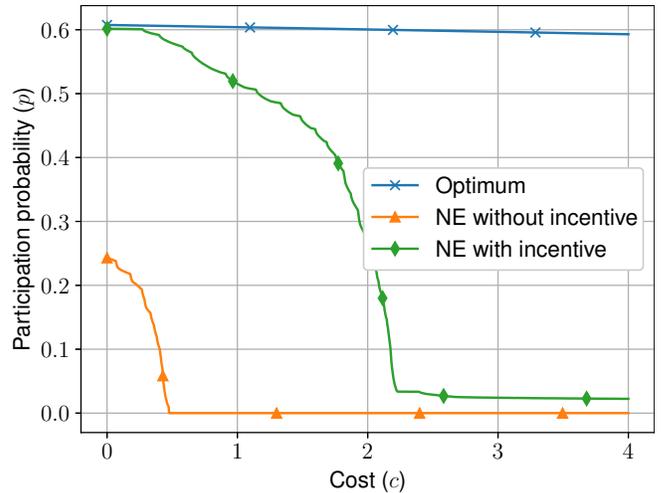}
    }
            \vspace{-0.6cm}
    \caption{Nodes' participation probability in the optimal centralized solution and at the NE with and without incentive.}
    \label{fig:participation_prob}
\end{figure}

\begin{figure}[t]
    \centering
    \resizebox{\columnwidth}{!}{
        \input{plots/utility_costv2.pgf}
    }
            \vspace{-0.7cm}
    \caption{Utility obtained by the optimal centralized nodes' participation probability and at the NE for various values of the $c$ parameter.}
    \label{fig:utility_cost}
\end{figure}
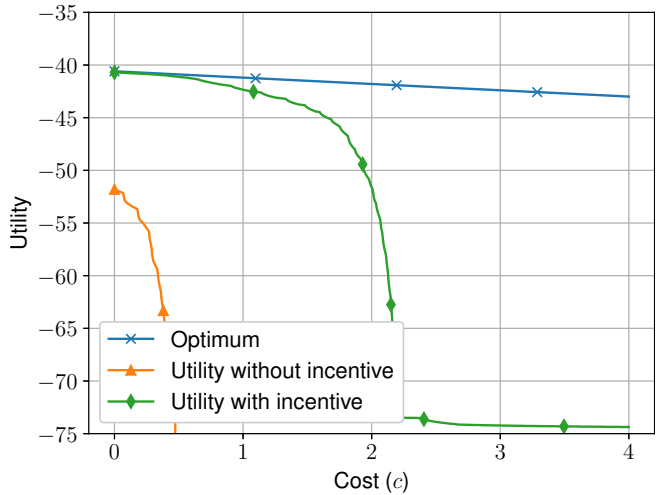

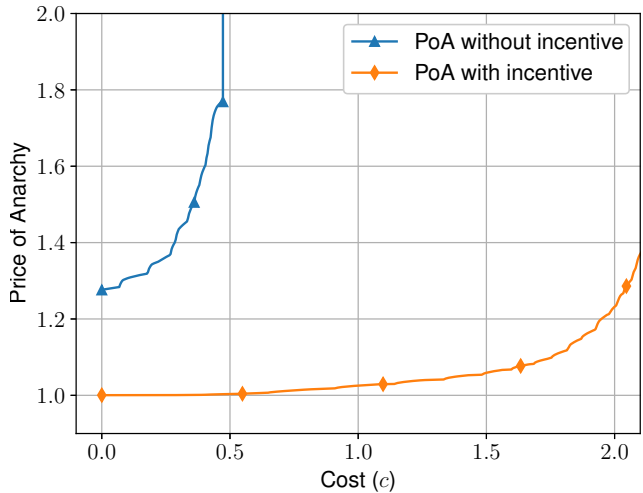
\begin{figure}[t]
    \centering
    \resizebox{0.98\columnwidth}{!}{
        \input{plots/poav2.pgf}
    }
    \caption{PoA as in (\ref{eq:poa}) for increasing values of the parameter $c$.}
            \vspace{-0.5cm}
    \label{fig:poa}
\end{figure}

\elia{To increase the generality of our results, we repeated the simulation with 
different random seeds. This influences both the decision of the nodes to join the current round (or not), and the initialization of the global model. Simulation runs are executed on the same hardware described before using four RTX2080 Ti in parallel. Due to Codecarbon limitations, it is not possible to distinguish the energy consumption of different processes running in parallel on the same machine, so the total energy is estimated from the linear interpolation shown in Fig.~\ref{fig:rounds_vs_energy}. Table~\ref{tab:results}(b) reports the average and standard deviation for the number of rounds and energy to converge.} 

To obtain results for the theoretical setup, we modeled the duration of the FL task using the data in Table~\ref{tab:results}(b). Specifically, we used a polynomial regression model to fit random points with a normal distribution as per mean and standard deviation of $d$ reported in the table. With said model, we calculated the utility as in Fig.~\ref{fig:utility} for $c{=}0$ and $\gamma{=}0$.

Fig.~\ref{fig:3dne} is a contour plot that shows the NE solution for $p$ with different values for the cost factor $c$ and the incentive weight $\gamma$. As expected, the nodes start to collaborate more by giving more weight to the incentive offsetting the cost they need to pay for the participation at the FL task. However for higher values of $\gamma$ it is evident that there is a tradeoff between the cost and the incentive. This is probably due to the shape of our utility function that penalizes the nodes if all of them participate in the task.
For the following plots we chose to fix $\gamma \approx 0.6$ as it is the value that obtains the highest participation probability.

Fig.~\ref{fig:participation_prob} shows the participation probability chosen by the nodes by global maximization of the utility and the NE as obtained by solving \eqref{eq:ne}. For $c=0$ the optimal participation probability is approximately $p=0.61$, while the NE without incentive obtains $p=0.24$. For increasing cost values this solution quickly falls to $p=0$. This is an evident \emph{Tragedy of the Commons} \cite{prospero2021resource}, since selfish users share a common resource, yet their individual participation has little influence on the payoff, so they do not cooperate to a full extent, damaging both the collective and ultimately their own interest.
The incentive mechanism based on AoI proves to be effective in improving nodes collaboration as the highest probability is $p=0.6$. This solution then quickly falls to lower participation as the incentive is not enough to offset the cost but it never reaches $p=0$.

Fig.~\ref{fig:utility_cost} shows the evolution of the utility obtained by the solutions. Here it is evident the sudden drop in the NE solution without incentive as soon as the participation probability approaches $0$. Conversely, the NE with incentive has a significant drop from the optimum, but then becomes stable.
By considering the PoA reported in Fig.~\ref{fig:poa} we can appreciate the importance of the incentive on the NE solution. Not having any kind of incentive obtains $PoA \simeq 1.28$ for $c=0$, this means that a distributed solution with these constraints has a $28\%$ loss over the performance of a centralized participation schedule. This becomes even worse for higher cost factors and eventually explodes to infinity. The NE with incentive based on AoI manages to obtain $PoA \approx 1$ and does not grow as rapidly as the previous case meaning that this solution allows transmissions even when they become more costly.

\section{Conclusions}
\label{sec:concs}

We explored a collaborative federated learning scenario, where individual nodes are in control of their participation to the learning task. The overall objective is to minimize the task duration and consequently the energy consumption. We tackled this challenge through the instruments of game theory \cite{dasari2020game} deriving the NE of the resulting allocation, and ultimately computing its energetic performance.

A fully distributed optimization is not feasible without an incentive mechanism in place, as in the best case there would be a $28\%$ performance drop with respect to a centralized solution.

Thus, we introduced an incentive mechanism for the nodes based on AoI that is able to offset the participation cost that they need to sustain \cite{tu2022incentive}. A future extension of this work may consider other more sophisticated incentive mechanisms.

\bibliographystyle{IEEEtran}
\bibliography{IEEEabrv,bibliography}

\end{document}

%% file: plots/scatterplot_with_interpolation.pgf
\begin{tikzpicture}

\definecolor{darkgray176}{RGB}{176,176,176}
\definecolor{lightgray204}{RGB}{204,204,204}
\definecolor{orange}{RGB}{255,165,0}
\definecolor{steelblue31119180}{RGB}{31,119,180}

\begin{axis}[
legend cell align={left},
legend style={
  fill opacity=0.8,
  draw opacity=1,
  text opacity=1,
  at={(0.03,0.97)},
  anchor=north west,
  draw=lightgray204
},
tick align=outside,
tick pos=left,
x grid style={darkgray176},
xlabel={Num rounds},
xmajorgrids,
xmin=0, xmax=80,
xtick style={color=black},
y grid style={darkgray176},
ylabel={Total Energy (Wh)},
ymajorgrids,
ymin=0, ymax=1200,
ytick style={color=black}
]
\addplot [draw=steelblue31119180, fill=steelblue31119180, mark=*, only marks]
table{%
x  y
74 1056.81
73 1060.25
57 830.9
73 1073.33
65 962.9
40 600.42
57 861.87
45 691.04
41 638.27
45 720.66
39 641.78
41 691.9
48 811.87
38 647.21
43 736.57
40 686.69
48 827.07
51 884.16
40 698.03
40 700.97
39 686.84
39 689.25
37 656.18
37 660.68
37 663.44
39 702.24
41 741.38
43 781.14
38 692.42
36 659.89
36 662.56
34 627.1
36 666.57
38 707.24
43 804
46 865.1
38 716.03
37 698.39
43 816.24
38 724.07
32 612.04
37 711.64
};
\addlegendentry{Data points}
\addplot [semithick, orange, dashed]
table {%
0 276.495467186938
0.808080808080808 285.16164267829
1.61616161616162 293.827818169642
2.42424242424242 302.493993660993
3.23232323232323 311.160169152345
4.04040404040404 319.826344643696
4.84848484848485 328.492520135048
5.65656565656566 337.158695626399
6.46464646464646 345.824871117751
7.27272727272727 354.491046609102
8.08080808080808 363.157222100454
8.88888888888889 371.823397591806
9.6969696969697 380.489573083157
10.5050505050505 389.155748574509
11.3131313131313 397.82192406586
12.1212121212121 406.488099557212
12.9292929292929 415.154275048563
13.7373737373737 423.820450539915
14.5454545454545 432.486626031266
15.3535353535354 441.152801522618
16.1616161616162 449.818977013969
16.969696969697 458.485152505321
17.7777777777778 467.151327996673
18.5858585858586 475.817503488024
19.3939393939394 484.483678979376
20.2020202020202 493.149854470727
21.010101010101 501.816029962079
21.8181818181818 510.48220545343
22.6262626262626 519.148380944782
23.4343434343434 527.814556436133
24.2424242424242 536.480731927485
25.050505050505 545.146907418836
25.8585858585859 553.813082910188
26.6666666666667 562.47925840154
27.4747474747475 571.145433892891
28.2828282828283 579.811609384243
29.0909090909091 588.477784875594
29.8989898989899 597.143960366946
30.7070707070707 605.810135858297
31.5151515151515 614.476311349649
32.3232323232323 623.142486841
33.1313131313131 631.808662332352
33.9393939393939 640.474837823704
34.7474747474747 649.141013315055
35.5555555555556 657.807188806407
36.3636363636364 666.473364297758
37.1717171717172 675.13953978911
37.979797979798 683.805715280461
38.7878787878788 692.471890771813
39.5959595959596 701.138066263164
40.4040404040404 709.804241754516
41.2121212121212 718.470417245867
42.020202020202 727.136592737219
42.8282828282828 735.802768228571
43.6363636363636 744.468943719922
44.4444444444444 753.135119211274
45.2525252525253 761.801294702625
46.0606060606061 770.467470193977
46.8686868686869 779.133645685328
47.6767676767677 787.79982117668
48.4848484848485 796.465996668031
49.2929292929293 805.132172159383
50.1010101010101 813.798347650735
50.9090909090909 822.464523142086
51.7171717171717 831.130698633438
52.5252525252525 839.796874124789
53.3333333333333 848.463049616141
54.1414141414141 857.129225107492
54.9494949494949 865.795400598844
55.7575757575758 874.461576090195
56.5656565656566 883.127751581547
57.3737373737374 891.793927072898
58.1818181818182 900.46010256425
58.989898989899 909.126278055601
59.7979797979798 917.792453546953
60.6060606060606 926.458629038305
61.4141414141414 935.124804529656
62.2222222222222 943.790980021008
63.030303030303 952.457155512359
63.8383838383838 961.123331003711
64.6464646464647 969.789506495062
65.4545454545455 978.455681986414
66.2626262626263 987.121857477765
67.0707070707071 995.788032969117
67.8787878787879 1004.45420846047
68.6868686868687 1013.12038395182
69.4949494949495 1021.78655944317
70.3030303030303 1030.45273493452
71.1111111111111 1039.11891042587
71.9191919191919 1047.78508591723
72.7272727272727 1056.45126140858
73.5353535353535 1065.11743689993
74.3434343434343 1073.78361239128
75.1515151515152 1082.44978788263
75.959595959596 1091.11596337398
76.7676767676768 1099.78213886534
77.5757575757576 1108.44831435669
78.3838383838384 1117.11448984804
79.1919191919192 1125.78066533939
80 1134.44684083074
};
\addlegendentry{Interpolated data}
\end{axis}

\end{tikzpicture}

%% file: plots/utility.pgf
\begingroup%
\makeatletter%
\begin{pgfpicture}%
\pgfpathrectangle{\pgfpointorigin}{\pgfqpoint{5.641942in}{4.226555in}}%
\pgfusepath{use as bounding box, clip}%
\begin{pgfscope}%
\pgfsetbuttcap%
\pgfsetmiterjoin%
\definecolor{currentfill}{rgb}{1.000000,1.000000,1.000000}%
\pgfsetfillcolor{currentfill}%
\pgfsetlinewidth{0.000000pt}%
\definecolor{currentstroke}{rgb}{1.000000,1.000000,1.000000}%
\pgfsetstrokecolor{currentstroke}%
\pgfsetdash{}{0pt}%
\pgfpathmoveto{\pgfqpoint{0.000000in}{0.000000in}}%
\pgfpathlineto{\pgfqpoint{5.641942in}{0.000000in}}%
\pgfpathlineto{\pgfqpoint{5.641942in}{4.226555in}}%
\pgfpathlineto{\pgfqpoint{0.000000in}{4.226555in}}%
\pgfpathlineto{\pgfqpoint{0.000000in}{0.000000in}}%
\pgfpathclose%
\pgfusepath{fill}%
\end{pgfscope}%
\begin{pgfscope}%
\pgfsetbuttcap%
\pgfsetmiterjoin%
\definecolor{currentfill}{rgb}{1.000000,1.000000,1.000000}%
\pgfsetfillcolor{currentfill}%
\pgfsetlinewidth{0.000000pt}%
\definecolor{currentstroke}{rgb}{0.000000,0.000000,0.000000}%
\pgfsetstrokecolor{currentstroke}%
\pgfsetstrokeopacity{0.000000}%
\pgfsetdash{}{0pt}%
\pgfpathmoveto{\pgfqpoint{0.681942in}{0.530555in}}%
\pgfpathlineto{\pgfqpoint{5.641942in}{0.530555in}}%
\pgfpathlineto{\pgfqpoint{5.641942in}{4.226555in}}%
\pgfpathlineto{\pgfqpoint{0.681942in}{4.226555in}}%
\pgfpathlineto{\pgfqpoint{0.681942in}{0.530555in}}%
\pgfpathclose%
\pgfusepath{fill}%
\end{pgfscope}%
\begin{pgfscope}%
\pgfpathrectangle{\pgfqpoint{0.681942in}{0.530555in}}{\pgfqpoint{4.960000in}{3.696000in}}%
\pgfusepath{clip}%
\pgfsetrectcap%
\pgfsetroundjoin%
\pgfsetlinewidth{0.803000pt}%
\definecolor{currentstroke}{rgb}{0.690196,0.690196,0.690196}%
\pgfsetstrokecolor{currentstroke}%
\pgfsetdash{}{0pt}%
\pgfpathmoveto{\pgfqpoint{0.861850in}{0.530555in}}%
\pgfpathlineto{\pgfqpoint{0.861850in}{4.226555in}}%
\pgfusepath{stroke}%
\end{pgfscope}%
\begin{pgfscope}%
\pgfsetbuttcap%
\pgfsetroundjoin%
\definecolor{currentfill}{rgb}{0.000000,0.000000,0.000000}%
\pgfsetfillcolor{currentfill}%
\pgfsetlinewidth{0.803000pt}%
\definecolor{currentstroke}{rgb}{0.000000,0.000000,0.000000}%
\pgfsetstrokecolor{currentstroke}%
\pgfsetdash{}{0pt}%
\pgfsys@defobject{currentmarker}{\pgfqpoint{0.000000in}{-0.048611in}}{\pgfqpoint{0.000000in}{0.000000in}}{%
\pgfpathmoveto{\pgfqpoint{0.000000in}{0.000000in}}%
\pgfpathlineto{\pgfqpoint{0.000000in}{-0.048611in}}%
\pgfusepath{stroke,fill}%
}%
\begin{pgfscope}%
\pgfsys@transformshift{0.861850in}{0.530555in}%
\pgfsys@useobject{currentmarker}{}%
\end{pgfscope}%
\end{pgfscope}%
\begin{pgfscope}%
\definecolor{textcolor}{rgb}{0.000000,0.000000,0.000000}%
\pgfsetstrokecolor{textcolor}%
\pgfsetfillcolor{textcolor}%
\pgftext[x=0.861850in,y=0.433333in,,top]{\color{textcolor}\sffamily\fontsize{14.000000}{16.800000}\selectfont \(\displaystyle {0.0}\)}%
\end{pgfscope}%
\begin{pgfscope}%
\pgfpathrectangle{\pgfqpoint{0.681942in}{0.530555in}}{\pgfqpoint{4.960000in}{3.696000in}}%
\pgfusepath{clip}%
\pgfsetrectcap%
\pgfsetroundjoin%
\pgfsetlinewidth{0.803000pt}%
\definecolor{currentstroke}{rgb}{0.690196,0.690196,0.690196}%
\pgfsetstrokecolor{currentstroke}%
\pgfsetdash{}{0pt}%
\pgfpathmoveto{\pgfqpoint{1.772778in}{0.530555in}}%
\pgfpathlineto{\pgfqpoint{1.772778in}{4.226555in}}%
\pgfusepath{stroke}%
\end{pgfscope}%
\begin{pgfscope}%
\pgfsetbuttcap%
\pgfsetroundjoin%
\definecolor{currentfill}{rgb}{0.000000,0.000000,0.000000}%
\pgfsetfillcolor{currentfill}%
\pgfsetlinewidth{0.803000pt}%
\definecolor{currentstroke}{rgb}{0.000000,0.000000,0.000000}%
\pgfsetstrokecolor{currentstroke}%
\pgfsetdash{}{0pt}%
\pgfsys@defobject{currentmarker}{\pgfqpoint{0.000000in}{-0.048611in}}{\pgfqpoint{0.000000in}{0.000000in}}{%
\pgfpathmoveto{\pgfqpoint{0.000000in}{0.000000in}}%
\pgfpathlineto{\pgfqpoint{0.000000in}{-0.048611in}}%
\pgfusepath{stroke,fill}%
}%
\begin{pgfscope}%
\pgfsys@transformshift{1.772778in}{0.530555in}%
\pgfsys@useobject{currentmarker}{}%
\end{pgfscope}%
\end{pgfscope}%
\begin{pgfscope}%
\definecolor{textcolor}{rgb}{0.000000,0.000000,0.000000}%
\pgfsetstrokecolor{textcolor}%
\pgfsetfillcolor{textcolor}%
\pgftext[x=1.772778in,y=0.433333in,,top]{\color{textcolor}\sffamily\fontsize{14.000000}{16.800000}\selectfont \(\displaystyle {0.2}\)}%
\end{pgfscope}%
\begin{pgfscope}%
\pgfpathrectangle{\pgfqpoint{0.681942in}{0.530555in}}{\pgfqpoint{4.960000in}{3.696000in}}%
\pgfusepath{clip}%
\pgfsetrectcap%
\pgfsetroundjoin%
\pgfsetlinewidth{0.803000pt}%
\definecolor{currentstroke}{rgb}{0.690196,0.690196,0.690196}%
\pgfsetstrokecolor{currentstroke}%
\pgfsetdash{}{0pt}%
\pgfpathmoveto{\pgfqpoint{2.683705in}{0.530555in}}%
\pgfpathlineto{\pgfqpoint{2.683705in}{4.226555in}}%
\pgfusepath{stroke}%
\end{pgfscope}%
\begin{pgfscope}%
\pgfsetbuttcap%
\pgfsetroundjoin%
\definecolor{currentfill}{rgb}{0.000000,0.000000,0.000000}%
\pgfsetfillcolor{currentfill}%
\pgfsetlinewidth{0.803000pt}%
\definecolor{currentstroke}{rgb}{0.000000,0.000000,0.000000}%
\pgfsetstrokecolor{currentstroke}%
\pgfsetdash{}{0pt}%
\pgfsys@defobject{currentmarker}{\pgfqpoint{0.000000in}{-0.048611in}}{\pgfqpoint{0.000000in}{0.000000in}}{%
\pgfpathmoveto{\pgfqpoint{0.000000in}{0.000000in}}%
\pgfpathlineto{\pgfqpoint{0.000000in}{-0.048611in}}%
\pgfusepath{stroke,fill}%
}%
\begin{pgfscope}%
\pgfsys@transformshift{2.683705in}{0.530555in}%
\pgfsys@useobject{currentmarker}{}%
\end{pgfscope}%
\end{pgfscope}%
\begin{pgfscope}%
\definecolor{textcolor}{rgb}{0.000000,0.000000,0.000000}%
\pgfsetstrokecolor{textcolor}%
\pgfsetfillcolor{textcolor}%
\pgftext[x=2.683705in,y=0.433333in,,top]{\color{textcolor}\sffamily\fontsize{14.000000}{16.800000}\selectfont \(\displaystyle {0.4}\)}%
\end{pgfscope}%
\begin{pgfscope}%
\pgfpathrectangle{\pgfqpoint{0.681942in}{0.530555in}}{\pgfqpoint{4.960000in}{3.696000in}}%
\pgfusepath{clip}%
\pgfsetrectcap%
\pgfsetroundjoin%
\pgfsetlinewidth{0.803000pt}%
\definecolor{currentstroke}{rgb}{0.690196,0.690196,0.690196}%
\pgfsetstrokecolor{currentstroke}%
\pgfsetdash{}{0pt}%
\pgfpathmoveto{\pgfqpoint{3.594632in}{0.530555in}}%
\pgfpathlineto{\pgfqpoint{3.594632in}{4.226555in}}%
\pgfusepath{stroke}%
\end{pgfscope}%
\begin{pgfscope}%
\pgfsetbuttcap%
\pgfsetroundjoin%
\definecolor{currentfill}{rgb}{0.000000,0.000000,0.000000}%
\pgfsetfillcolor{currentfill}%
\pgfsetlinewidth{0.803000pt}%
\definecolor{currentstroke}{rgb}{0.000000,0.000000,0.000000}%
\pgfsetstrokecolor{currentstroke}%
\pgfsetdash{}{0pt}%
\pgfsys@defobject{currentmarker}{\pgfqpoint{0.000000in}{-0.048611in}}{\pgfqpoint{0.000000in}{0.000000in}}{%
\pgfpathmoveto{\pgfqpoint{0.000000in}{0.000000in}}%
\pgfpathlineto{\pgfqpoint{0.000000in}{-0.048611in}}%
\pgfusepath{stroke,fill}%
}%
\begin{pgfscope}%
\pgfsys@transformshift{3.594632in}{0.530555in}%
\pgfsys@useobject{currentmarker}{}%
\end{pgfscope}%
\end{pgfscope}%
\begin{pgfscope}%
\definecolor{textcolor}{rgb}{0.000000,0.000000,0.000000}%
\pgfsetstrokecolor{textcolor}%
\pgfsetfillcolor{textcolor}%
\pgftext[x=3.594632in,y=0.433333in,,top]{\color{textcolor}\sffamily\fontsize{14.000000}{16.800000}\selectfont \(\displaystyle {0.6}\)}%
\end{pgfscope}%
\begin{pgfscope}%
\pgfpathrectangle{\pgfqpoint{0.681942in}{0.530555in}}{\pgfqpoint{4.960000in}{3.696000in}}%
\pgfusepath{clip}%
\pgfsetrectcap%
\pgfsetroundjoin%
\pgfsetlinewidth{0.803000pt}%
\definecolor{currentstroke}{rgb}{0.690196,0.690196,0.690196}%
\pgfsetstrokecolor{currentstroke}%
\pgfsetdash{}{0pt}%
\pgfpathmoveto{\pgfqpoint{4.505560in}{0.530555in}}%
\pgfpathlineto{\pgfqpoint{4.505560in}{4.226555in}}%
\pgfusepath{stroke}%
\end{pgfscope}%
\begin{pgfscope}%
\pgfsetbuttcap%
\pgfsetroundjoin%
\definecolor{currentfill}{rgb}{0.000000,0.000000,0.000000}%
\pgfsetfillcolor{currentfill}%
\pgfsetlinewidth{0.803000pt}%
\definecolor{currentstroke}{rgb}{0.000000,0.000000,0.000000}%
\pgfsetstrokecolor{currentstroke}%
\pgfsetdash{}{0pt}%
\pgfsys@defobject{currentmarker}{\pgfqpoint{0.000000in}{-0.048611in}}{\pgfqpoint{0.000000in}{0.000000in}}{%
\pgfpathmoveto{\pgfqpoint{0.000000in}{0.000000in}}%
\pgfpathlineto{\pgfqpoint{0.000000in}{-0.048611in}}%
\pgfusepath{stroke,fill}%
}%
\begin{pgfscope}%
\pgfsys@transformshift{4.505560in}{0.530555in}%
\pgfsys@useobject{currentmarker}{}%
\end{pgfscope}%
\end{pgfscope}%
\begin{pgfscope}%
\definecolor{textcolor}{rgb}{0.000000,0.000000,0.000000}%
\pgfsetstrokecolor{textcolor}%
\pgfsetfillcolor{textcolor}%
\pgftext[x=4.505560in,y=0.433333in,,top]{\color{textcolor}\sffamily\fontsize{14.000000}{16.800000}\selectfont \(\displaystyle {0.8}\)}%
\end{pgfscope}%
\begin{pgfscope}%
\pgfpathrectangle{\pgfqpoint{0.681942in}{0.530555in}}{\pgfqpoint{4.960000in}{3.696000in}}%
\pgfusepath{clip}%
\pgfsetrectcap%
\pgfsetroundjoin%
\pgfsetlinewidth{0.803000pt}%
\definecolor{currentstroke}{rgb}{0.690196,0.690196,0.690196}%
\pgfsetstrokecolor{currentstroke}%
\pgfsetdash{}{0pt}%
\pgfpathmoveto{\pgfqpoint{5.416487in}{0.530555in}}%
\pgfpathlineto{\pgfqpoint{5.416487in}{4.226555in}}%
\pgfusepath{stroke}%
\end{pgfscope}%
\begin{pgfscope}%
\pgfsetbuttcap%
\pgfsetroundjoin%
\definecolor{currentfill}{rgb}{0.000000,0.000000,0.000000}%
\pgfsetfillcolor{currentfill}%
\pgfsetlinewidth{0.803000pt}%
\definecolor{currentstroke}{rgb}{0.000000,0.000000,0.000000}%
\pgfsetstrokecolor{currentstroke}%
\pgfsetdash{}{0pt}%
\pgfsys@defobject{currentmarker}{\pgfqpoint{0.000000in}{-0.048611in}}{\pgfqpoint{0.000000in}{0.000000in}}{%
\pgfpathmoveto{\pgfqpoint{0.000000in}{0.000000in}}%
\pgfpathlineto{\pgfqpoint{0.000000in}{-0.048611in}}%
\pgfusepath{stroke,fill}%
}%
\begin{pgfscope}%
\pgfsys@transformshift{5.416487in}{0.530555in}%
\pgfsys@useobject{currentmarker}{}%
\end{pgfscope}%
\end{pgfscope}%
\begin{pgfscope}%
\definecolor{textcolor}{rgb}{0.000000,0.000000,0.000000}%
\pgfsetstrokecolor{textcolor}%
\pgfsetfillcolor{textcolor}%
\pgftext[x=5.416487in,y=0.433333in,,top]{\color{textcolor}\sffamily\fontsize{14.000000}{16.800000}\selectfont \(\displaystyle {1.0}\)}%
\end{pgfscope}%
\begin{pgfscope}%
\definecolor{textcolor}{rgb}{0.000000,0.000000,0.000000}%
\pgfsetstrokecolor{textcolor}%
\pgfsetfillcolor{textcolor}%
\pgftext[x=3.161942in,y=0.200000in,,top]{\color{textcolor}\sffamily\fontsize{14.000000}{16.800000}\selectfont Participation probability (\(\displaystyle p\))}%
\end{pgfscope}%
\begin{pgfscope}%
\pgfpathrectangle{\pgfqpoint{0.681942in}{0.530555in}}{\pgfqpoint{4.960000in}{3.696000in}}%
\pgfusepath{clip}%
\pgfsetrectcap%
\pgfsetroundjoin%
\pgfsetlinewidth{0.803000pt}%
\definecolor{currentstroke}{rgb}{0.690196,0.690196,0.690196}%
\pgfsetstrokecolor{currentstroke}%
\pgfsetdash{}{0pt}%
\pgfpathmoveto{\pgfqpoint{0.681942in}{0.944230in}}%
\pgfpathlineto{\pgfqpoint{5.641942in}{0.944230in}}%
\pgfusepath{stroke}%
\end{pgfscope}%
\begin{pgfscope}%
\pgfsetbuttcap%
\pgfsetroundjoin%
\definecolor{currentfill}{rgb}{0.000000,0.000000,0.000000}%
\pgfsetfillcolor{currentfill}%
\pgfsetlinewidth{0.803000pt}%
\definecolor{currentstroke}{rgb}{0.000000,0.000000,0.000000}%
\pgfsetstrokecolor{currentstroke}%
\pgfsetdash{}{0pt}%
\pgfsys@defobject{currentmarker}{\pgfqpoint{-0.048611in}{0.000000in}}{\pgfqpoint{-0.000000in}{0.000000in}}{%
\pgfpathmoveto{\pgfqpoint{-0.000000in}{0.000000in}}%
\pgfpathlineto{\pgfqpoint{-0.048611in}{0.000000in}}%
\pgfusepath{stroke,fill}%
}%
\begin{pgfscope}%
\pgfsys@transformshift{0.681942in}{0.944230in}%
\pgfsys@useobject{currentmarker}{}%
\end{pgfscope}%
\end{pgfscope}%
\begin{pgfscope}%
\definecolor{textcolor}{rgb}{0.000000,0.000000,0.000000}%
\pgfsetstrokecolor{textcolor}%
\pgfsetfillcolor{textcolor}%
\pgftext[x=0.233333in, y=0.874786in, left, base]{\color{textcolor}\sffamily\fontsize{14.000000}{16.800000}\selectfont \(\displaystyle {\ensuremath{-}70}\)}%
\end{pgfscope}%
\begin{pgfscope}%
\pgfpathrectangle{\pgfqpoint{0.681942in}{0.530555in}}{\pgfqpoint{4.960000in}{3.696000in}}%
\pgfusepath{clip}%
\pgfsetrectcap%
\pgfsetroundjoin%
\pgfsetlinewidth{0.803000pt}%
\definecolor{currentstroke}{rgb}{0.690196,0.690196,0.690196}%
\pgfsetstrokecolor{currentstroke}%
\pgfsetdash{}{0pt}%
\pgfpathmoveto{\pgfqpoint{0.681942in}{1.473767in}}%
\pgfpathlineto{\pgfqpoint{5.641942in}{1.473767in}}%
\pgfusepath{stroke}%
\end{pgfscope}%
\begin{pgfscope}%
\pgfsetbuttcap%
\pgfsetroundjoin%
\definecolor{currentfill}{rgb}{0.000000,0.000000,0.000000}%
\pgfsetfillcolor{currentfill}%
\pgfsetlinewidth{0.803000pt}%
\definecolor{currentstroke}{rgb}{0.000000,0.000000,0.000000}%
\pgfsetstrokecolor{currentstroke}%
\pgfsetdash{}{0pt}%
\pgfsys@defobject{currentmarker}{\pgfqpoint{-0.048611in}{0.000000in}}{\pgfqpoint{-0.000000in}{0.000000in}}{%
\pgfpathmoveto{\pgfqpoint{-0.000000in}{0.000000in}}%
\pgfpathlineto{\pgfqpoint{-0.048611in}{0.000000in}}%
\pgfusepath{stroke,fill}%
}%
\begin{pgfscope}%
\pgfsys@transformshift{0.681942in}{1.473767in}%
\pgfsys@useobject{currentmarker}{}%
\end{pgfscope}%
\end{pgfscope}%
\begin{pgfscope}%
\definecolor{textcolor}{rgb}{0.000000,0.000000,0.000000}%
\pgfsetstrokecolor{textcolor}%
\pgfsetfillcolor{textcolor}%
\pgftext[x=0.233333in, y=1.404323in, left, base]{\color{textcolor}\sffamily\fontsize{14.000000}{16.800000}\selectfont \(\displaystyle {\ensuremath{-}65}\)}%
\end{pgfscope}%
\begin{pgfscope}%
\pgfpathrectangle{\pgfqpoint{0.681942in}{0.530555in}}{\pgfqpoint{4.960000in}{3.696000in}}%
\pgfusepath{clip}%
\pgfsetrectcap%
\pgfsetroundjoin%
\pgfsetlinewidth{0.803000pt}%
\definecolor{currentstroke}{rgb}{0.690196,0.690196,0.690196}%
\pgfsetstrokecolor{currentstroke}%
\pgfsetdash{}{0pt}%
\pgfpathmoveto{\pgfqpoint{0.681942in}{2.003305in}}%
\pgfpathlineto{\pgfqpoint{5.641942in}{2.003305in}}%
\pgfusepath{stroke}%
\end{pgfscope}%
\begin{pgfscope}%
\pgfsetbuttcap%
\pgfsetroundjoin%
\definecolor{currentfill}{rgb}{0.000000,0.000000,0.000000}%
\pgfsetfillcolor{currentfill}%
\pgfsetlinewidth{0.803000pt}%
\definecolor{currentstroke}{rgb}{0.000000,0.000000,0.000000}%
\pgfsetstrokecolor{currentstroke}%
\pgfsetdash{}{0pt}%
\pgfsys@defobject{currentmarker}{\pgfqpoint{-0.048611in}{0.000000in}}{\pgfqpoint{-0.000000in}{0.000000in}}{%
\pgfpathmoveto{\pgfqpoint{-0.000000in}{0.000000in}}%
\pgfpathlineto{\pgfqpoint{-0.048611in}{0.000000in}}%
\pgfusepath{stroke,fill}%
}%
\begin{pgfscope}%
\pgfsys@transformshift{0.681942in}{2.003305in}%
\pgfsys@useobject{currentmarker}{}%
\end{pgfscope}%
\end{pgfscope}%
\begin{pgfscope}%
\definecolor{textcolor}{rgb}{0.000000,0.000000,0.000000}%
\pgfsetstrokecolor{textcolor}%
\pgfsetfillcolor{textcolor}%
\pgftext[x=0.233333in, y=1.933861in, left, base]{\color{textcolor}\sffamily\fontsize{14.000000}{16.800000}\selectfont \(\displaystyle {\ensuremath{-}60}\)}%
\end{pgfscope}%
\begin{pgfscope}%
\pgfpathrectangle{\pgfqpoint{0.681942in}{0.530555in}}{\pgfqpoint{4.960000in}{3.696000in}}%
\pgfusepath{clip}%
\pgfsetrectcap%
\pgfsetroundjoin%
\pgfsetlinewidth{0.803000pt}%
\definecolor{currentstroke}{rgb}{0.690196,0.690196,0.690196}%
\pgfsetstrokecolor{currentstroke}%
\pgfsetdash{}{0pt}%
\pgfpathmoveto{\pgfqpoint{0.681942in}{2.532842in}}%
\pgfpathlineto{\pgfqpoint{5.641942in}{2.532842in}}%
\pgfusepath{stroke}%
\end{pgfscope}%
\begin{pgfscope}%
\pgfsetbuttcap%
\pgfsetroundjoin%
\definecolor{currentfill}{rgb}{0.000000,0.000000,0.000000}%
\pgfsetfillcolor{currentfill}%
\pgfsetlinewidth{0.803000pt}%
\definecolor{currentstroke}{rgb}{0.000000,0.000000,0.000000}%
\pgfsetstrokecolor{currentstroke}%
\pgfsetdash{}{0pt}%
\pgfsys@defobject{currentmarker}{\pgfqpoint{-0.048611in}{0.000000in}}{\pgfqpoint{-0.000000in}{0.000000in}}{%
\pgfpathmoveto{\pgfqpoint{-0.000000in}{0.000000in}}%
\pgfpathlineto{\pgfqpoint{-0.048611in}{0.000000in}}%
\pgfusepath{stroke,fill}%
}%
\begin{pgfscope}%
\pgfsys@transformshift{0.681942in}{2.532842in}%
\pgfsys@useobject{currentmarker}{}%
\end{pgfscope}%
\end{pgfscope}%
\begin{pgfscope}%
\definecolor{textcolor}{rgb}{0.000000,0.000000,0.000000}%
\pgfsetstrokecolor{textcolor}%
\pgfsetfillcolor{textcolor}%
\pgftext[x=0.233333in, y=2.463398in, left, base]{\color{textcolor}\sffamily\fontsize{14.000000}{16.800000}\selectfont \(\displaystyle {\ensuremath{-}55}\)}%
\end{pgfscope}%
\begin{pgfscope}%
\pgfpathrectangle{\pgfqpoint{0.681942in}{0.530555in}}{\pgfqpoint{4.960000in}{3.696000in}}%
\pgfusepath{clip}%
\pgfsetrectcap%
\pgfsetroundjoin%
\pgfsetlinewidth{0.803000pt}%
\definecolor{currentstroke}{rgb}{0.690196,0.690196,0.690196}%
\pgfsetstrokecolor{currentstroke}%
\pgfsetdash{}{0pt}%
\pgfpathmoveto{\pgfqpoint{0.681942in}{3.062380in}}%
\pgfpathlineto{\pgfqpoint{5.641942in}{3.062380in}}%
\pgfusepath{stroke}%
\end{pgfscope}%
\begin{pgfscope}%
\pgfsetbuttcap%
\pgfsetroundjoin%
\definecolor{currentfill}{rgb}{0.000000,0.000000,0.000000}%
\pgfsetfillcolor{currentfill}%
\pgfsetlinewidth{0.803000pt}%
\definecolor{currentstroke}{rgb}{0.000000,0.000000,0.000000}%
\pgfsetstrokecolor{currentstroke}%
\pgfsetdash{}{0pt}%
\pgfsys@defobject{currentmarker}{\pgfqpoint{-0.048611in}{0.000000in}}{\pgfqpoint{-0.000000in}{0.000000in}}{%
\pgfpathmoveto{\pgfqpoint{-0.000000in}{0.000000in}}%
\pgfpathlineto{\pgfqpoint{-0.048611in}{0.000000in}}%
\pgfusepath{stroke,fill}%
}%
\begin{pgfscope}%
\pgfsys@transformshift{0.681942in}{3.062380in}%
\pgfsys@useobject{currentmarker}{}%
\end{pgfscope}%
\end{pgfscope}%
\begin{pgfscope}%
\definecolor{textcolor}{rgb}{0.000000,0.000000,0.000000}%
\pgfsetstrokecolor{textcolor}%
\pgfsetfillcolor{textcolor}%
\pgftext[x=0.233333in, y=2.992936in, left, base]{\color{textcolor}\sffamily\fontsize{14.000000}{16.800000}\selectfont \(\displaystyle {\ensuremath{-}50}\)}%
\end{pgfscope}%
\begin{pgfscope}%
\pgfpathrectangle{\pgfqpoint{0.681942in}{0.530555in}}{\pgfqpoint{4.960000in}{3.696000in}}%
\pgfusepath{clip}%
\pgfsetrectcap%
\pgfsetroundjoin%
\pgfsetlinewidth{0.803000pt}%
\definecolor{currentstroke}{rgb}{0.690196,0.690196,0.690196}%
\pgfsetstrokecolor{currentstroke}%
\pgfsetdash{}{0pt}%
\pgfpathmoveto{\pgfqpoint{0.681942in}{3.591917in}}%
\pgfpathlineto{\pgfqpoint{5.641942in}{3.591917in}}%
\pgfusepath{stroke}%
\end{pgfscope}%
\begin{pgfscope}%
\pgfsetbuttcap%
\pgfsetroundjoin%
\definecolor{currentfill}{rgb}{0.000000,0.000000,0.000000}%
\pgfsetfillcolor{currentfill}%
\pgfsetlinewidth{0.803000pt}%
\definecolor{currentstroke}{rgb}{0.000000,0.000000,0.000000}%
\pgfsetstrokecolor{currentstroke}%
\pgfsetdash{}{0pt}%
\pgfsys@defobject{currentmarker}{\pgfqpoint{-0.048611in}{0.000000in}}{\pgfqpoint{-0.000000in}{0.000000in}}{%
\pgfpathmoveto{\pgfqpoint{-0.000000in}{0.000000in}}%
\pgfpathlineto{\pgfqpoint{-0.048611in}{0.000000in}}%
\pgfusepath{stroke,fill}%
}%
\begin{pgfscope}%
\pgfsys@transformshift{0.681942in}{3.591917in}%
\pgfsys@useobject{currentmarker}{}%
\end{pgfscope}%
\end{pgfscope}%
\begin{pgfscope}%
\definecolor{textcolor}{rgb}{0.000000,0.000000,0.000000}%
\pgfsetstrokecolor{textcolor}%
\pgfsetfillcolor{textcolor}%
\pgftext[x=0.233333in, y=3.522473in, left, base]{\color{textcolor}\sffamily\fontsize{14.000000}{16.800000}\selectfont \(\displaystyle {\ensuremath{-}45}\)}%
\end{pgfscope}%
\begin{pgfscope}%
\pgfpathrectangle{\pgfqpoint{0.681942in}{0.530555in}}{\pgfqpoint{4.960000in}{3.696000in}}%
\pgfusepath{clip}%
\pgfsetrectcap%
\pgfsetroundjoin%
\pgfsetlinewidth{0.803000pt}%
\definecolor{currentstroke}{rgb}{0.690196,0.690196,0.690196}%
\pgfsetstrokecolor{currentstroke}%
\pgfsetdash{}{0pt}%
\pgfpathmoveto{\pgfqpoint{0.681942in}{4.121455in}}%
\pgfpathlineto{\pgfqpoint{5.641942in}{4.121455in}}%
\pgfusepath{stroke}%
\end{pgfscope}%
\begin{pgfscope}%
\pgfsetbuttcap%
\pgfsetroundjoin%
\definecolor{currentfill}{rgb}{0.000000,0.000000,0.000000}%
\pgfsetfillcolor{currentfill}%
\pgfsetlinewidth{0.803000pt}%
\definecolor{currentstroke}{rgb}{0.000000,0.000000,0.000000}%
\pgfsetstrokecolor{currentstroke}%
\pgfsetdash{}{0pt}%
\pgfsys@defobject{currentmarker}{\pgfqpoint{-0.048611in}{0.000000in}}{\pgfqpoint{-0.000000in}{0.000000in}}{%
\pgfpathmoveto{\pgfqpoint{-0.000000in}{0.000000in}}%
\pgfpathlineto{\pgfqpoint{-0.048611in}{0.000000in}}%
\pgfusepath{stroke,fill}%
}%
\begin{pgfscope}%
\pgfsys@transformshift{0.681942in}{4.121455in}%
\pgfsys@useobject{currentmarker}{}%
\end{pgfscope}%
\end{pgfscope}%
\begin{pgfscope}%
\definecolor{textcolor}{rgb}{0.000000,0.000000,0.000000}%
\pgfsetstrokecolor{textcolor}%
\pgfsetfillcolor{textcolor}%
\pgftext[x=0.233333in, y=4.052010in, left, base]{\color{textcolor}\sffamily\fontsize{14.000000}{16.800000}\selectfont \(\displaystyle {\ensuremath{-}40}\)}%
\end{pgfscope}%
\begin{pgfscope}%
\definecolor{textcolor}{rgb}{0.000000,0.000000,0.000000}%
\pgfsetstrokecolor{textcolor}%
\pgfsetfillcolor{textcolor}%
\pgftext[x=0.177777in,y=2.378555in,,bottom,rotate=90.000000]{\color{textcolor}\sffamily\fontsize{14.000000}{16.800000}\selectfont Utility}%
\end{pgfscope}%
\begin{pgfscope}%
\pgfpathrectangle{\pgfqpoint{0.681942in}{0.530555in}}{\pgfqpoint{4.960000in}{3.696000in}}%
\pgfusepath{clip}%
\pgfsetrectcap%
\pgfsetroundjoin%
\pgfsetlinewidth{1.505625pt}%
\definecolor{currentstroke}{rgb}{0.121569,0.466667,0.705882}%
\pgfsetstrokecolor{currentstroke}%
\pgfsetdash{}{0pt}%
\pgfpathmoveto{\pgfqpoint{0.907396in}{0.698555in}}%
\pgfpathlineto{\pgfqpoint{0.952943in}{0.737167in}}%
\pgfpathlineto{\pgfqpoint{0.998489in}{0.789132in}}%
\pgfpathlineto{\pgfqpoint{1.044036in}{0.853393in}}%
\pgfpathlineto{\pgfqpoint{1.089582in}{0.928759in}}%
\pgfpathlineto{\pgfqpoint{1.135128in}{1.013938in}}%
\pgfpathlineto{\pgfqpoint{1.180675in}{1.107572in}}%
\pgfpathlineto{\pgfqpoint{1.226221in}{1.208271in}}%
\pgfpathlineto{\pgfqpoint{1.271767in}{1.314637in}}%
\pgfpathlineto{\pgfqpoint{1.317314in}{1.425296in}}%
\pgfpathlineto{\pgfqpoint{1.362860in}{1.538919in}}%
\pgfpathlineto{\pgfqpoint{1.408407in}{1.654243in}}%
\pgfpathlineto{\pgfqpoint{1.453953in}{1.770086in}}%
\pgfpathlineto{\pgfqpoint{1.499499in}{1.885362in}}%
\pgfpathlineto{\pgfqpoint{1.545046in}{1.999094in}}%
\pgfpathlineto{\pgfqpoint{1.590592in}{2.110415in}}%
\pgfpathlineto{\pgfqpoint{1.636138in}{2.218583in}}%
\pgfpathlineto{\pgfqpoint{1.681685in}{2.322975in}}%
\pgfpathlineto{\pgfqpoint{1.727231in}{2.423093in}}%
\pgfpathlineto{\pgfqpoint{1.772778in}{2.518560in}}%
\pgfpathlineto{\pgfqpoint{1.818324in}{2.609118in}}%
\pgfpathlineto{\pgfqpoint{1.863870in}{2.694622in}}%
\pgfpathlineto{\pgfqpoint{1.909417in}{2.775033in}}%
\pgfpathlineto{\pgfqpoint{1.954963in}{2.850410in}}%
\pgfpathlineto{\pgfqpoint{2.000509in}{2.920898in}}%
\pgfpathlineto{\pgfqpoint{2.046056in}{2.986720in}}%
\pgfpathlineto{\pgfqpoint{2.091602in}{3.048161in}}%
\pgfpathlineto{\pgfqpoint{2.137149in}{3.105558in}}%
\pgfpathlineto{\pgfqpoint{2.182695in}{3.159282in}}%
\pgfpathlineto{\pgfqpoint{2.228241in}{3.209728in}}%
\pgfpathlineto{\pgfqpoint{2.273788in}{3.257299in}}%
\pgfpathlineto{\pgfqpoint{2.319334in}{3.302391in}}%
\pgfpathlineto{\pgfqpoint{2.364880in}{3.345386in}}%
\pgfpathlineto{\pgfqpoint{2.410427in}{3.386633in}}%
\pgfpathlineto{\pgfqpoint{2.455973in}{3.426446in}}%
\pgfpathlineto{\pgfqpoint{2.501520in}{3.465090in}}%
\pgfpathlineto{\pgfqpoint{2.547066in}{3.502783in}}%
\pgfpathlineto{\pgfqpoint{2.592612in}{3.539685in}}%
\pgfpathlineto{\pgfqpoint{2.638159in}{3.575902in}}%
\pgfpathlineto{\pgfqpoint{2.683705in}{3.611489in}}%
\pgfpathlineto{\pgfqpoint{2.729251in}{3.646449in}}%
\pgfpathlineto{\pgfqpoint{2.774798in}{3.680741in}}%
\pgfpathlineto{\pgfqpoint{2.820344in}{3.714291in}}%
\pgfpathlineto{\pgfqpoint{2.865891in}{3.746991in}}%
\pgfpathlineto{\pgfqpoint{2.911437in}{3.778718in}}%
\pgfpathlineto{\pgfqpoint{2.956983in}{3.809333in}}%
\pgfpathlineto{\pgfqpoint{3.002530in}{3.838698in}}%
\pgfpathlineto{\pgfqpoint{3.048076in}{3.866677in}}%
\pgfpathlineto{\pgfqpoint{3.093622in}{3.893140in}}%
\pgfpathlineto{\pgfqpoint{3.139169in}{3.917976in}}%
\pgfpathlineto{\pgfqpoint{3.184715in}{3.941083in}}%
\pgfpathlineto{\pgfqpoint{3.230262in}{3.962377in}}%
\pgfpathlineto{\pgfqpoint{3.275808in}{3.981781in}}%
\pgfpathlineto{\pgfqpoint{3.321354in}{3.999227in}}%
\pgfpathlineto{\pgfqpoint{3.366901in}{4.014645in}}%
\pgfpathlineto{\pgfqpoint{3.412447in}{4.027958in}}%
\pgfpathlineto{\pgfqpoint{3.457993in}{4.039071in}}%
\pgfpathlineto{\pgfqpoint{3.503540in}{4.047862in}}%
\pgfpathlineto{\pgfqpoint{3.549086in}{4.054178in}}%
\pgfpathlineto{\pgfqpoint{3.594632in}{4.057823in}}%
\pgfpathlineto{\pgfqpoint{3.640179in}{4.058555in}}%
\pgfpathlineto{\pgfqpoint{3.685725in}{4.056083in}}%
\pgfpathlineto{\pgfqpoint{3.731272in}{4.050070in}}%
\pgfpathlineto{\pgfqpoint{3.776818in}{4.040134in}}%
\pgfpathlineto{\pgfqpoint{3.822364in}{4.025855in}}%
\pgfpathlineto{\pgfqpoint{3.867911in}{4.006792in}}%
\pgfpathlineto{\pgfqpoint{3.913457in}{3.982490in}}%
\pgfpathlineto{\pgfqpoint{3.959003in}{3.952503in}}%
\pgfpathlineto{\pgfqpoint{4.004550in}{3.916414in}}%
\pgfpathlineto{\pgfqpoint{4.050096in}{3.873855in}}%
\pgfpathlineto{\pgfqpoint{4.095643in}{3.824532in}}%
\pgfpathlineto{\pgfqpoint{4.141189in}{3.768246in}}%
\pgfpathlineto{\pgfqpoint{4.186735in}{3.704919in}}%
\pgfpathlineto{\pgfqpoint{4.232282in}{3.634606in}}%
\pgfpathlineto{\pgfqpoint{4.277828in}{3.557519in}}%
\pgfpathlineto{\pgfqpoint{4.323374in}{3.474034in}}%
\pgfpathlineto{\pgfqpoint{4.368921in}{3.384696in}}%
\pgfpathlineto{\pgfqpoint{4.414467in}{3.290225in}}%
\pgfpathlineto{\pgfqpoint{4.460014in}{3.191505in}}%
\pgfpathlineto{\pgfqpoint{4.505560in}{3.089575in}}%
\pgfpathlineto{\pgfqpoint{4.551106in}{2.985608in}}%
\pgfpathlineto{\pgfqpoint{4.596653in}{2.880888in}}%
\pgfpathlineto{\pgfqpoint{4.642199in}{2.776779in}}%
\pgfpathlineto{\pgfqpoint{4.687745in}{2.674691in}}%
\pgfpathlineto{\pgfqpoint{4.733292in}{2.576045in}}%
\pgfpathlineto{\pgfqpoint{4.778838in}{2.482227in}}%
\pgfpathlineto{\pgfqpoint{4.824385in}{2.394556in}}%
\pgfpathlineto{\pgfqpoint{4.869931in}{2.314239in}}%
\pgfpathlineto{\pgfqpoint{4.915477in}{2.242339in}}%
\pgfpathlineto{\pgfqpoint{4.961024in}{2.179741in}}%
\pgfpathlineto{\pgfqpoint{5.006570in}{2.127125in}}%
\pgfpathlineto{\pgfqpoint{5.052116in}{2.084948in}}%
\pgfpathlineto{\pgfqpoint{5.097663in}{2.053428in}}%
\pgfpathlineto{\pgfqpoint{5.143209in}{2.032544in}}%
\pgfpathlineto{\pgfqpoint{5.188756in}{2.022034in}}%
\pgfpathlineto{\pgfqpoint{5.234302in}{2.021411in}}%
\pgfpathlineto{\pgfqpoint{5.279848in}{2.029980in}}%
\pgfpathlineto{\pgfqpoint{5.325395in}{2.046865in}}%
\pgfpathlineto{\pgfqpoint{5.370941in}{2.071039in}}%
\pgfpathlineto{\pgfqpoint{5.416487in}{2.101363in}}%
\pgfusepath{stroke}%
\end{pgfscope}%
\begin{pgfscope}%
\pgfsetrectcap%
\pgfsetmiterjoin%
\pgfsetlinewidth{0.803000pt}%
\definecolor{currentstroke}{rgb}{0.000000,0.000000,0.000000}%
\pgfsetstrokecolor{currentstroke}%
\pgfsetdash{}{0pt}%
\pgfpathmoveto{\pgfqpoint{0.681942in}{0.530555in}}%
\pgfpathlineto{\pgfqpoint{0.681942in}{4.226555in}}%
\pgfusepath{stroke}%
\end{pgfscope}%
\begin{pgfscope}%
\pgfsetrectcap%
\pgfsetmiterjoin%
\pgfsetlinewidth{0.803000pt}%
\definecolor{currentstroke}{rgb}{0.000000,0.000000,0.000000}%
\pgfsetstrokecolor{currentstroke}%
\pgfsetdash{}{0pt}%
\pgfpathmoveto{\pgfqpoint{5.641942in}{0.530555in}}%
\pgfpathlineto{\pgfqpoint{5.641942in}{4.226555in}}%
\pgfusepath{stroke}%
\end{pgfscope}%
\begin{pgfscope}%
\pgfsetrectcap%
\pgfsetmiterjoin%
\pgfsetlinewidth{0.803000pt}%
\definecolor{currentstroke}{rgb}{0.000000,0.000000,0.000000}%
\pgfsetstrokecolor{currentstroke}%
\pgfsetdash{}{0pt}%
\pgfpathmoveto{\pgfqpoint{0.681942in}{0.530555in}}%
\pgfpathlineto{\pgfqpoint{5.641942in}{0.530555in}}%
\pgfusepath{stroke}%
\end{pgfscope}%
\begin{pgfscope}%
\pgfsetrectcap%
\pgfsetmiterjoin%
\pgfsetlinewidth{0.803000pt}%
\definecolor{currentstroke}{rgb}{0.000000,0.000000,0.000000}%
\pgfsetstrokecolor{currentstroke}%
\pgfsetdash{}{0pt}%
\pgfpathmoveto{\pgfqpoint{0.681942in}{4.226555in}}%
\pgfpathlineto{\pgfqpoint{5.641942in}{4.226555in}}%
\pgfusepath{stroke}%
\end{pgfscope}%
\end{pgfpicture}%
\makeatother%
\endgroup%

%% file: plots/utility_costv2.pgf
\begingroup%
\makeatletter%
\begin{pgfpicture}%
\pgfpathrectangle{\pgfpointorigin}{\pgfqpoint{5.641942in}{4.295999in}}%
\pgfusepath{use as bounding box, clip}%
\begin{pgfscope}%
\pgfsetbuttcap%
\pgfsetmiterjoin%
\definecolor{currentfill}{rgb}{1.000000,1.000000,1.000000}%
\pgfsetfillcolor{currentfill}%
\pgfsetlinewidth{0.000000pt}%
\definecolor{currentstroke}{rgb}{1.000000,1.000000,1.000000}%
\pgfsetstrokecolor{currentstroke}%
\pgfsetdash{}{0pt}%
\pgfpathmoveto{\pgfqpoint{0.000000in}{0.000000in}}%
\pgfpathlineto{\pgfqpoint{5.641942in}{0.000000in}}%
\pgfpathlineto{\pgfqpoint{5.641942in}{4.295999in}}%
\pgfpathlineto{\pgfqpoint{0.000000in}{4.295999in}}%
\pgfpathlineto{\pgfqpoint{0.000000in}{0.000000in}}%
\pgfpathclose%
\pgfusepath{fill}%
\end{pgfscope}%
\begin{pgfscope}%
\pgfsetbuttcap%
\pgfsetmiterjoin%
\definecolor{currentfill}{rgb}{1.000000,1.000000,1.000000}%
\pgfsetfillcolor{currentfill}%
\pgfsetlinewidth{0.000000pt}%
\definecolor{currentstroke}{rgb}{0.000000,0.000000,0.000000}%
\pgfsetstrokecolor{currentstroke}%
\pgfsetstrokeopacity{0.000000}%
\pgfsetdash{}{0pt}%
\pgfpathmoveto{\pgfqpoint{0.681942in}{0.530555in}}%
\pgfpathlineto{\pgfqpoint{5.641942in}{0.530555in}}%
\pgfpathlineto{\pgfqpoint{5.641942in}{4.226555in}}%
\pgfpathlineto{\pgfqpoint{0.681942in}{4.226555in}}%
\pgfpathlineto{\pgfqpoint{0.681942in}{0.530555in}}%
\pgfpathclose%
\pgfusepath{fill}%
\end{pgfscope}%
\begin{pgfscope}%
\pgfpathrectangle{\pgfqpoint{0.681942in}{0.530555in}}{\pgfqpoint{4.960000in}{3.696000in}}%
\pgfusepath{clip}%
\pgfsetrectcap%
\pgfsetroundjoin%
\pgfsetlinewidth{0.803000pt}%
\definecolor{currentstroke}{rgb}{0.690196,0.690196,0.690196}%
\pgfsetstrokecolor{currentstroke}%
\pgfsetdash{}{0pt}%
\pgfpathmoveto{\pgfqpoint{0.907396in}{0.530555in}}%
\pgfpathlineto{\pgfqpoint{0.907396in}{4.226555in}}%
\pgfusepath{stroke}%
\end{pgfscope}%
\begin{pgfscope}%
\pgfsetbuttcap%
\pgfsetroundjoin%
\definecolor{currentfill}{rgb}{0.000000,0.000000,0.000000}%
\pgfsetfillcolor{currentfill}%
\pgfsetlinewidth{0.803000pt}%
\definecolor{currentstroke}{rgb}{0.000000,0.000000,0.000000}%
\pgfsetstrokecolor{currentstroke}%
\pgfsetdash{}{0pt}%
\pgfsys@defobject{currentmarker}{\pgfqpoint{0.000000in}{-0.048611in}}{\pgfqpoint{0.000000in}{0.000000in}}{%
\pgfpathmoveto{\pgfqpoint{0.000000in}{0.000000in}}%
\pgfpathlineto{\pgfqpoint{0.000000in}{-0.048611in}}%
\pgfusepath{stroke,fill}%
}%
\begin{pgfscope}%
\pgfsys@transformshift{0.907396in}{0.530555in}%
\pgfsys@useobject{currentmarker}{}%
\end{pgfscope}%
\end{pgfscope}%
\begin{pgfscope}%
\definecolor{textcolor}{rgb}{0.000000,0.000000,0.000000}%
\pgfsetstrokecolor{textcolor}%
\pgfsetfillcolor{textcolor}%
\pgftext[x=0.907396in,y=0.433333in,,top]{\color{textcolor}\sffamily\fontsize{14.000000}{16.800000}\selectfont \(\displaystyle {0}\)}%
\end{pgfscope}%
\begin{pgfscope}%
\pgfpathrectangle{\pgfqpoint{0.681942in}{0.530555in}}{\pgfqpoint{4.960000in}{3.696000in}}%
\pgfusepath{clip}%
\pgfsetrectcap%
\pgfsetroundjoin%
\pgfsetlinewidth{0.803000pt}%
\definecolor{currentstroke}{rgb}{0.690196,0.690196,0.690196}%
\pgfsetstrokecolor{currentstroke}%
\pgfsetdash{}{0pt}%
\pgfpathmoveto{\pgfqpoint{2.034669in}{0.530555in}}%
\pgfpathlineto{\pgfqpoint{2.034669in}{4.226555in}}%
\pgfusepath{stroke}%
\end{pgfscope}%
\begin{pgfscope}%
\pgfsetbuttcap%
\pgfsetroundjoin%
\definecolor{currentfill}{rgb}{0.000000,0.000000,0.000000}%
\pgfsetfillcolor{currentfill}%
\pgfsetlinewidth{0.803000pt}%
\definecolor{currentstroke}{rgb}{0.000000,0.000000,0.000000}%
\pgfsetstrokecolor{currentstroke}%
\pgfsetdash{}{0pt}%
\pgfsys@defobject{currentmarker}{\pgfqpoint{0.000000in}{-0.048611in}}{\pgfqpoint{0.000000in}{0.000000in}}{%
\pgfpathmoveto{\pgfqpoint{0.000000in}{0.000000in}}%
\pgfpathlineto{\pgfqpoint{0.000000in}{-0.048611in}}%
\pgfusepath{stroke,fill}%
}%
\begin{pgfscope}%
\pgfsys@transformshift{2.034669in}{0.530555in}%
\pgfsys@useobject{currentmarker}{}%
\end{pgfscope}%
\end{pgfscope}%
\begin{pgfscope}%
\definecolor{textcolor}{rgb}{0.000000,0.000000,0.000000}%
\pgfsetstrokecolor{textcolor}%
\pgfsetfillcolor{textcolor}%
\pgftext[x=2.034669in,y=0.433333in,,top]{\color{textcolor}\sffamily\fontsize{14.000000}{16.800000}\selectfont \(\displaystyle {1}\)}%
\end{pgfscope}%
\begin{pgfscope}%
\pgfpathrectangle{\pgfqpoint{0.681942in}{0.530555in}}{\pgfqpoint{4.960000in}{3.696000in}}%
\pgfusepath{clip}%
\pgfsetrectcap%
\pgfsetroundjoin%
\pgfsetlinewidth{0.803000pt}%
\definecolor{currentstroke}{rgb}{0.690196,0.690196,0.690196}%
\pgfsetstrokecolor{currentstroke}%
\pgfsetdash{}{0pt}%
\pgfpathmoveto{\pgfqpoint{3.161942in}{0.530555in}}%
\pgfpathlineto{\pgfqpoint{3.161942in}{4.226555in}}%
\pgfusepath{stroke}%
\end{pgfscope}%
\begin{pgfscope}%
\pgfsetbuttcap%
\pgfsetroundjoin%
\definecolor{currentfill}{rgb}{0.000000,0.000000,0.000000}%
\pgfsetfillcolor{currentfill}%
\pgfsetlinewidth{0.803000pt}%
\definecolor{currentstroke}{rgb}{0.000000,0.000000,0.000000}%
\pgfsetstrokecolor{currentstroke}%
\pgfsetdash{}{0pt}%
\pgfsys@defobject{currentmarker}{\pgfqpoint{0.000000in}{-0.048611in}}{\pgfqpoint{0.000000in}{0.000000in}}{%
\pgfpathmoveto{\pgfqpoint{0.000000in}{0.000000in}}%
\pgfpathlineto{\pgfqpoint{0.000000in}{-0.048611in}}%
\pgfusepath{stroke,fill}%
}%
\begin{pgfscope}%
\pgfsys@transformshift{3.161942in}{0.530555in}%
\pgfsys@useobject{currentmarker}{}%
\end{pgfscope}%
\end{pgfscope}%
\begin{pgfscope}%
\definecolor{textcolor}{rgb}{0.000000,0.000000,0.000000}%
\pgfsetstrokecolor{textcolor}%
\pgfsetfillcolor{textcolor}%
\pgftext[x=3.161942in,y=0.433333in,,top]{\color{textcolor}\sffamily\fontsize{14.000000}{16.800000}\selectfont \(\displaystyle {2}\)}%
\end{pgfscope}%
\begin{pgfscope}%
\pgfpathrectangle{\pgfqpoint{0.681942in}{0.530555in}}{\pgfqpoint{4.960000in}{3.696000in}}%
\pgfusepath{clip}%
\pgfsetrectcap%
\pgfsetroundjoin%
\pgfsetlinewidth{0.803000pt}%
\definecolor{currentstroke}{rgb}{0.690196,0.690196,0.690196}%
\pgfsetstrokecolor{currentstroke}%
\pgfsetdash{}{0pt}%
\pgfpathmoveto{\pgfqpoint{4.289215in}{0.530555in}}%
\pgfpathlineto{\pgfqpoint{4.289215in}{4.226555in}}%
\pgfusepath{stroke}%
\end{pgfscope}%
\begin{pgfscope}%
\pgfsetbuttcap%
\pgfsetroundjoin%
\definecolor{currentfill}{rgb}{0.000000,0.000000,0.000000}%
\pgfsetfillcolor{currentfill}%
\pgfsetlinewidth{0.803000pt}%
\definecolor{currentstroke}{rgb}{0.000000,0.000000,0.000000}%
\pgfsetstrokecolor{currentstroke}%
\pgfsetdash{}{0pt}%
\pgfsys@defobject{currentmarker}{\pgfqpoint{0.000000in}{-0.048611in}}{\pgfqpoint{0.000000in}{0.000000in}}{%
\pgfpathmoveto{\pgfqpoint{0.000000in}{0.000000in}}%
\pgfpathlineto{\pgfqpoint{0.000000in}{-0.048611in}}%
\pgfusepath{stroke,fill}%
}%
\begin{pgfscope}%
\pgfsys@transformshift{4.289215in}{0.530555in}%
\pgfsys@useobject{currentmarker}{}%
\end{pgfscope}%
\end{pgfscope}%
\begin{pgfscope}%
\definecolor{textcolor}{rgb}{0.000000,0.000000,0.000000}%
\pgfsetstrokecolor{textcolor}%
\pgfsetfillcolor{textcolor}%
\pgftext[x=4.289215in,y=0.433333in,,top]{\color{textcolor}\sffamily\fontsize{14.000000}{16.800000}\selectfont \(\displaystyle {3}\)}%
\end{pgfscope}%
\begin{pgfscope}%
\pgfpathrectangle{\pgfqpoint{0.681942in}{0.530555in}}{\pgfqpoint{4.960000in}{3.696000in}}%
\pgfusepath{clip}%
\pgfsetrectcap%
\pgfsetroundjoin%
\pgfsetlinewidth{0.803000pt}%
\definecolor{currentstroke}{rgb}{0.690196,0.690196,0.690196}%
\pgfsetstrokecolor{currentstroke}%
\pgfsetdash{}{0pt}%
\pgfpathmoveto{\pgfqpoint{5.416487in}{0.530555in}}%
\pgfpathlineto{\pgfqpoint{5.416487in}{4.226555in}}%
\pgfusepath{stroke}%
\end{pgfscope}%
\begin{pgfscope}%
\pgfsetbuttcap%
\pgfsetroundjoin%
\definecolor{currentfill}{rgb}{0.000000,0.000000,0.000000}%
\pgfsetfillcolor{currentfill}%
\pgfsetlinewidth{0.803000pt}%
\definecolor{currentstroke}{rgb}{0.000000,0.000000,0.000000}%
\pgfsetstrokecolor{currentstroke}%
\pgfsetdash{}{0pt}%
\pgfsys@defobject{currentmarker}{\pgfqpoint{0.000000in}{-0.048611in}}{\pgfqpoint{0.000000in}{0.000000in}}{%
\pgfpathmoveto{\pgfqpoint{0.000000in}{0.000000in}}%
\pgfpathlineto{\pgfqpoint{0.000000in}{-0.048611in}}%
\pgfusepath{stroke,fill}%
}%
\begin{pgfscope}%
\pgfsys@transformshift{5.416487in}{0.530555in}%
\pgfsys@useobject{currentmarker}{}%
\end{pgfscope}%
\end{pgfscope}%
\begin{pgfscope}%
\definecolor{textcolor}{rgb}{0.000000,0.000000,0.000000}%
\pgfsetstrokecolor{textcolor}%
\pgfsetfillcolor{textcolor}%
\pgftext[x=5.416487in,y=0.433333in,,top]{\color{textcolor}\sffamily\fontsize{14.000000}{16.800000}\selectfont \(\displaystyle {4}\)}%
\end{pgfscope}%
\begin{pgfscope}%
\definecolor{textcolor}{rgb}{0.000000,0.000000,0.000000}%
\pgfsetstrokecolor{textcolor}%
\pgfsetfillcolor{textcolor}%
\pgftext[x=3.161942in,y=0.200000in,,top]{\color{textcolor}\sffamily\fontsize{14.000000}{16.800000}\selectfont Cost (\(\displaystyle c\))}%
\end{pgfscope}%
\begin{pgfscope}%
\pgfpathrectangle{\pgfqpoint{0.681942in}{0.530555in}}{\pgfqpoint{4.960000in}{3.696000in}}%
\pgfusepath{clip}%
\pgfsetrectcap%
\pgfsetroundjoin%
\pgfsetlinewidth{0.803000pt}%
\definecolor{currentstroke}{rgb}{0.690196,0.690196,0.690196}%
\pgfsetstrokecolor{currentstroke}%
\pgfsetdash{}{0pt}%
\pgfpathmoveto{\pgfqpoint{0.681942in}{0.530555in}}%
\pgfpathlineto{\pgfqpoint{5.641942in}{0.530555in}}%
\pgfusepath{stroke}%
\end{pgfscope}%
\begin{pgfscope}%
\pgfsetbuttcap%
\pgfsetroundjoin%
\definecolor{currentfill}{rgb}{0.000000,0.000000,0.000000}%
\pgfsetfillcolor{currentfill}%
\pgfsetlinewidth{0.803000pt}%
\definecolor{currentstroke}{rgb}{0.000000,0.000000,0.000000}%
\pgfsetstrokecolor{currentstroke}%
\pgfsetdash{}{0pt}%
\pgfsys@defobject{currentmarker}{\pgfqpoint{-0.048611in}{0.000000in}}{\pgfqpoint{-0.000000in}{0.000000in}}{%
\pgfpathmoveto{\pgfqpoint{-0.000000in}{0.000000in}}%
\pgfpathlineto{\pgfqpoint{-0.048611in}{0.000000in}}%
\pgfusepath{stroke,fill}%
}%
\begin{pgfscope}%
\pgfsys@transformshift{0.681942in}{0.530555in}%
\pgfsys@useobject{currentmarker}{}%
\end{pgfscope}%
\end{pgfscope}%
\begin{pgfscope}%
\definecolor{textcolor}{rgb}{0.000000,0.000000,0.000000}%
\pgfsetstrokecolor{textcolor}%
\pgfsetfillcolor{textcolor}%
\pgftext[x=0.233333in, y=0.461111in, left, base]{\color{textcolor}\sffamily\fontsize{14.000000}{16.800000}\selectfont \(\displaystyle {\ensuremath{-}75}\)}%
\end{pgfscope}%
\begin{pgfscope}%
\pgfpathrectangle{\pgfqpoint{0.681942in}{0.530555in}}{\pgfqpoint{4.960000in}{3.696000in}}%
\pgfusepath{clip}%
\pgfsetrectcap%
\pgfsetroundjoin%
\pgfsetlinewidth{0.803000pt}%
\definecolor{currentstroke}{rgb}{0.690196,0.690196,0.690196}%
\pgfsetstrokecolor{currentstroke}%
\pgfsetdash{}{0pt}%
\pgfpathmoveto{\pgfqpoint{0.681942in}{0.992555in}}%
\pgfpathlineto{\pgfqpoint{5.641942in}{0.992555in}}%
\pgfusepath{stroke}%
\end{pgfscope}%
\begin{pgfscope}%
\pgfsetbuttcap%
\pgfsetroundjoin%
\definecolor{currentfill}{rgb}{0.000000,0.000000,0.000000}%
\pgfsetfillcolor{currentfill}%
\pgfsetlinewidth{0.803000pt}%
\definecolor{currentstroke}{rgb}{0.000000,0.000000,0.000000}%
\pgfsetstrokecolor{currentstroke}%
\pgfsetdash{}{0pt}%
\pgfsys@defobject{currentmarker}{\pgfqpoint{-0.048611in}{0.000000in}}{\pgfqpoint{-0.000000in}{0.000000in}}{%
\pgfpathmoveto{\pgfqpoint{-0.000000in}{0.000000in}}%
\pgfpathlineto{\pgfqpoint{-0.048611in}{0.000000in}}%
\pgfusepath{stroke,fill}%
}%
\begin{pgfscope}%
\pgfsys@transformshift{0.681942in}{0.992555in}%
\pgfsys@useobject{currentmarker}{}%
\end{pgfscope}%
\end{pgfscope}%
\begin{pgfscope}%
\definecolor{textcolor}{rgb}{0.000000,0.000000,0.000000}%
\pgfsetstrokecolor{textcolor}%
\pgfsetfillcolor{textcolor}%
\pgftext[x=0.233333in, y=0.923111in, left, base]{\color{textcolor}\sffamily\fontsize{14.000000}{16.800000}\selectfont \(\displaystyle {\ensuremath{-}70}\)}%
\end{pgfscope}%
\begin{pgfscope}%
\pgfpathrectangle{\pgfqpoint{0.681942in}{0.530555in}}{\pgfqpoint{4.960000in}{3.696000in}}%
\pgfusepath{clip}%
\pgfsetrectcap%
\pgfsetroundjoin%
\pgfsetlinewidth{0.803000pt}%
\definecolor{currentstroke}{rgb}{0.690196,0.690196,0.690196}%
\pgfsetstrokecolor{currentstroke}%
\pgfsetdash{}{0pt}%
\pgfpathmoveto{\pgfqpoint{0.681942in}{1.454555in}}%
\pgfpathlineto{\pgfqpoint{5.641942in}{1.454555in}}%
\pgfusepath{stroke}%
\end{pgfscope}%
\begin{pgfscope}%
\pgfsetbuttcap%
\pgfsetroundjoin%
\definecolor{currentfill}{rgb}{0.000000,0.000000,0.000000}%
\pgfsetfillcolor{currentfill}%
\pgfsetlinewidth{0.803000pt}%
\definecolor{currentstroke}{rgb}{0.000000,0.000000,0.000000}%
\pgfsetstrokecolor{currentstroke}%
\pgfsetdash{}{0pt}%
\pgfsys@defobject{currentmarker}{\pgfqpoint{-0.048611in}{0.000000in}}{\pgfqpoint{-0.000000in}{0.000000in}}{%
\pgfpathmoveto{\pgfqpoint{-0.000000in}{0.000000in}}%
\pgfpathlineto{\pgfqpoint{-0.048611in}{0.000000in}}%
\pgfusepath{stroke,fill}%
}%
\begin{pgfscope}%
\pgfsys@transformshift{0.681942in}{1.454555in}%
\pgfsys@useobject{currentmarker}{}%
\end{pgfscope}%
\end{pgfscope}%
\begin{pgfscope}%
\definecolor{textcolor}{rgb}{0.000000,0.000000,0.000000}%
\pgfsetstrokecolor{textcolor}%
\pgfsetfillcolor{textcolor}%
\pgftext[x=0.233333in, y=1.385111in, left, base]{\color{textcolor}\sffamily\fontsize{14.000000}{16.800000}\selectfont \(\displaystyle {\ensuremath{-}65}\)}%
\end{pgfscope}%
\begin{pgfscope}%
\pgfpathrectangle{\pgfqpoint{0.681942in}{0.530555in}}{\pgfqpoint{4.960000in}{3.696000in}}%
\pgfusepath{clip}%
\pgfsetrectcap%
\pgfsetroundjoin%
\pgfsetlinewidth{0.803000pt}%
\definecolor{currentstroke}{rgb}{0.690196,0.690196,0.690196}%
\pgfsetstrokecolor{currentstroke}%
\pgfsetdash{}{0pt}%
\pgfpathmoveto{\pgfqpoint{0.681942in}{1.916555in}}%
\pgfpathlineto{\pgfqpoint{5.641942in}{1.916555in}}%
\pgfusepath{stroke}%
\end{pgfscope}%
\begin{pgfscope}%
\pgfsetbuttcap%
\pgfsetroundjoin%
\definecolor{currentfill}{rgb}{0.000000,0.000000,0.000000}%
\pgfsetfillcolor{currentfill}%
\pgfsetlinewidth{0.803000pt}%
\definecolor{currentstroke}{rgb}{0.000000,0.000000,0.000000}%
\pgfsetstrokecolor{currentstroke}%
\pgfsetdash{}{0pt}%
\pgfsys@defobject{currentmarker}{\pgfqpoint{-0.048611in}{0.000000in}}{\pgfqpoint{-0.000000in}{0.000000in}}{%
\pgfpathmoveto{\pgfqpoint{-0.000000in}{0.000000in}}%
\pgfpathlineto{\pgfqpoint{-0.048611in}{0.000000in}}%
\pgfusepath{stroke,fill}%
}%
\begin{pgfscope}%
\pgfsys@transformshift{0.681942in}{1.916555in}%
\pgfsys@useobject{currentmarker}{}%
\end{pgfscope}%
\end{pgfscope}%
\begin{pgfscope}%
\definecolor{textcolor}{rgb}{0.000000,0.000000,0.000000}%
\pgfsetstrokecolor{textcolor}%
\pgfsetfillcolor{textcolor}%
\pgftext[x=0.233333in, y=1.847111in, left, base]{\color{textcolor}\sffamily\fontsize{14.000000}{16.800000}\selectfont \(\displaystyle {\ensuremath{-}60}\)}%
\end{pgfscope}%
\begin{pgfscope}%
\pgfpathrectangle{\pgfqpoint{0.681942in}{0.530555in}}{\pgfqpoint{4.960000in}{3.696000in}}%
\pgfusepath{clip}%
\pgfsetrectcap%
\pgfsetroundjoin%
\pgfsetlinewidth{0.803000pt}%
\definecolor{currentstroke}{rgb}{0.690196,0.690196,0.690196}%
\pgfsetstrokecolor{currentstroke}%
\pgfsetdash{}{0pt}%
\pgfpathmoveto{\pgfqpoint{0.681942in}{2.378555in}}%
\pgfpathlineto{\pgfqpoint{5.641942in}{2.378555in}}%
\pgfusepath{stroke}%
\end{pgfscope}%
\begin{pgfscope}%
\pgfsetbuttcap%
\pgfsetroundjoin%
\definecolor{currentfill}{rgb}{0.000000,0.000000,0.000000}%
\pgfsetfillcolor{currentfill}%
\pgfsetlinewidth{0.803000pt}%
\definecolor{currentstroke}{rgb}{0.000000,0.000000,0.000000}%
\pgfsetstrokecolor{currentstroke}%
\pgfsetdash{}{0pt}%
\pgfsys@defobject{currentmarker}{\pgfqpoint{-0.048611in}{0.000000in}}{\pgfqpoint{-0.000000in}{0.000000in}}{%
\pgfpathmoveto{\pgfqpoint{-0.000000in}{0.000000in}}%
\pgfpathlineto{\pgfqpoint{-0.048611in}{0.000000in}}%
\pgfusepath{stroke,fill}%
}%
\begin{pgfscope}%
\pgfsys@transformshift{0.681942in}{2.378555in}%
\pgfsys@useobject{currentmarker}{}%
\end{pgfscope}%
\end{pgfscope}%
\begin{pgfscope}%
\definecolor{textcolor}{rgb}{0.000000,0.000000,0.000000}%
\pgfsetstrokecolor{textcolor}%
\pgfsetfillcolor{textcolor}%
\pgftext[x=0.233333in, y=2.309111in, left, base]{\color{textcolor}\sffamily\fontsize{14.000000}{16.800000}\selectfont \(\displaystyle {\ensuremath{-}55}\)}%
\end{pgfscope}%
\begin{pgfscope}%
\pgfpathrectangle{\pgfqpoint{0.681942in}{0.530555in}}{\pgfqpoint{4.960000in}{3.696000in}}%
\pgfusepath{clip}%
\pgfsetrectcap%
\pgfsetroundjoin%
\pgfsetlinewidth{0.803000pt}%
\definecolor{currentstroke}{rgb}{0.690196,0.690196,0.690196}%
\pgfsetstrokecolor{currentstroke}%
\pgfsetdash{}{0pt}%
\pgfpathmoveto{\pgfqpoint{0.681942in}{2.840555in}}%
\pgfpathlineto{\pgfqpoint{5.641942in}{2.840555in}}%
\pgfusepath{stroke}%
\end{pgfscope}%
\begin{pgfscope}%
\pgfsetbuttcap%
\pgfsetroundjoin%
\definecolor{currentfill}{rgb}{0.000000,0.000000,0.000000}%
\pgfsetfillcolor{currentfill}%
\pgfsetlinewidth{0.803000pt}%
\definecolor{currentstroke}{rgb}{0.000000,0.000000,0.000000}%
\pgfsetstrokecolor{currentstroke}%
\pgfsetdash{}{0pt}%
\pgfsys@defobject{currentmarker}{\pgfqpoint{-0.048611in}{0.000000in}}{\pgfqpoint{-0.000000in}{0.000000in}}{%
\pgfpathmoveto{\pgfqpoint{-0.000000in}{0.000000in}}%
\pgfpathlineto{\pgfqpoint{-0.048611in}{0.000000in}}%
\pgfusepath{stroke,fill}%
}%
\begin{pgfscope}%
\pgfsys@transformshift{0.681942in}{2.840555in}%
\pgfsys@useobject{currentmarker}{}%
\end{pgfscope}%
\end{pgfscope}%
\begin{pgfscope}%
\definecolor{textcolor}{rgb}{0.000000,0.000000,0.000000}%
\pgfsetstrokecolor{textcolor}%
\pgfsetfillcolor{textcolor}%
\pgftext[x=0.233333in, y=2.771111in, left, base]{\color{textcolor}\sffamily\fontsize{14.000000}{16.800000}\selectfont \(\displaystyle {\ensuremath{-}50}\)}%
\end{pgfscope}%
\begin{pgfscope}%
\pgfpathrectangle{\pgfqpoint{0.681942in}{0.530555in}}{\pgfqpoint{4.960000in}{3.696000in}}%
\pgfusepath{clip}%
\pgfsetrectcap%
\pgfsetroundjoin%
\pgfsetlinewidth{0.803000pt}%
\definecolor{currentstroke}{rgb}{0.690196,0.690196,0.690196}%
\pgfsetstrokecolor{currentstroke}%
\pgfsetdash{}{0pt}%
\pgfpathmoveto{\pgfqpoint{0.681942in}{3.302555in}}%
\pgfpathlineto{\pgfqpoint{5.641942in}{3.302555in}}%
\pgfusepath{stroke}%
\end{pgfscope}%
\begin{pgfscope}%
\pgfsetbuttcap%
\pgfsetroundjoin%
\definecolor{currentfill}{rgb}{0.000000,0.000000,0.000000}%
\pgfsetfillcolor{currentfill}%
\pgfsetlinewidth{0.803000pt}%
\definecolor{currentstroke}{rgb}{0.000000,0.000000,0.000000}%
\pgfsetstrokecolor{currentstroke}%
\pgfsetdash{}{0pt}%
\pgfsys@defobject{currentmarker}{\pgfqpoint{-0.048611in}{0.000000in}}{\pgfqpoint{-0.000000in}{0.000000in}}{%
\pgfpathmoveto{\pgfqpoint{-0.000000in}{0.000000in}}%
\pgfpathlineto{\pgfqpoint{-0.048611in}{0.000000in}}%
\pgfusepath{stroke,fill}%
}%
\begin{pgfscope}%
\pgfsys@transformshift{0.681942in}{3.302555in}%
\pgfsys@useobject{currentmarker}{}%
\end{pgfscope}%
\end{pgfscope}%
\begin{pgfscope}%
\definecolor{textcolor}{rgb}{0.000000,0.000000,0.000000}%
\pgfsetstrokecolor{textcolor}%
\pgfsetfillcolor{textcolor}%
\pgftext[x=0.233333in, y=3.233111in, left, base]{\color{textcolor}\sffamily\fontsize{14.000000}{16.800000}\selectfont \(\displaystyle {\ensuremath{-}45}\)}%
\end{pgfscope}%
\begin{pgfscope}%
\pgfpathrectangle{\pgfqpoint{0.681942in}{0.530555in}}{\pgfqpoint{4.960000in}{3.696000in}}%
\pgfusepath{clip}%
\pgfsetrectcap%
\pgfsetroundjoin%
\pgfsetlinewidth{0.803000pt}%
\definecolor{currentstroke}{rgb}{0.690196,0.690196,0.690196}%
\pgfsetstrokecolor{currentstroke}%
\pgfsetdash{}{0pt}%
\pgfpathmoveto{\pgfqpoint{0.681942in}{3.764555in}}%
\pgfpathlineto{\pgfqpoint{5.641942in}{3.764555in}}%
\pgfusepath{stroke}%
\end{pgfscope}%
\begin{pgfscope}%
\pgfsetbuttcap%
\pgfsetroundjoin%
\definecolor{currentfill}{rgb}{0.000000,0.000000,0.000000}%
\pgfsetfillcolor{currentfill}%
\pgfsetlinewidth{0.803000pt}%
\definecolor{currentstroke}{rgb}{0.000000,0.000000,0.000000}%
\pgfsetstrokecolor{currentstroke}%
\pgfsetdash{}{0pt}%
\pgfsys@defobject{currentmarker}{\pgfqpoint{-0.048611in}{0.000000in}}{\pgfqpoint{-0.000000in}{0.000000in}}{%
\pgfpathmoveto{\pgfqpoint{-0.000000in}{0.000000in}}%
\pgfpathlineto{\pgfqpoint{-0.048611in}{0.000000in}}%
\pgfusepath{stroke,fill}%
}%
\begin{pgfscope}%
\pgfsys@transformshift{0.681942in}{3.764555in}%
\pgfsys@useobject{currentmarker}{}%
\end{pgfscope}%
\end{pgfscope}%
\begin{pgfscope}%
\definecolor{textcolor}{rgb}{0.000000,0.000000,0.000000}%
\pgfsetstrokecolor{textcolor}%
\pgfsetfillcolor{textcolor}%
\pgftext[x=0.233333in, y=3.695111in, left, base]{\color{textcolor}\sffamily\fontsize{14.000000}{16.800000}\selectfont \(\displaystyle {\ensuremath{-}40}\)}%
\end{pgfscope}%
\begin{pgfscope}%
\pgfpathrectangle{\pgfqpoint{0.681942in}{0.530555in}}{\pgfqpoint{4.960000in}{3.696000in}}%
\pgfusepath{clip}%
\pgfsetrectcap%
\pgfsetroundjoin%
\pgfsetlinewidth{0.803000pt}%
\definecolor{currentstroke}{rgb}{0.690196,0.690196,0.690196}%
\pgfsetstrokecolor{currentstroke}%
\pgfsetdash{}{0pt}%
\pgfpathmoveto{\pgfqpoint{0.681942in}{4.226555in}}%
\pgfpathlineto{\pgfqpoint{5.641942in}{4.226555in}}%
\pgfusepath{stroke}%
\end{pgfscope}%
\begin{pgfscope}%
\pgfsetbuttcap%
\pgfsetroundjoin%
\definecolor{currentfill}{rgb}{0.000000,0.000000,0.000000}%
\pgfsetfillcolor{currentfill}%
\pgfsetlinewidth{0.803000pt}%
\definecolor{currentstroke}{rgb}{0.000000,0.000000,0.000000}%
\pgfsetstrokecolor{currentstroke}%
\pgfsetdash{}{0pt}%
\pgfsys@defobject{currentmarker}{\pgfqpoint{-0.048611in}{0.000000in}}{\pgfqpoint{-0.000000in}{0.000000in}}{%
\pgfpathmoveto{\pgfqpoint{-0.000000in}{0.000000in}}%
\pgfpathlineto{\pgfqpoint{-0.048611in}{0.000000in}}%
\pgfusepath{stroke,fill}%
}%
\begin{pgfscope}%
\pgfsys@transformshift{0.681942in}{4.226555in}%
\pgfsys@useobject{currentmarker}{}%
\end{pgfscope}%
\end{pgfscope}%
\begin{pgfscope}%
\definecolor{textcolor}{rgb}{0.000000,0.000000,0.000000}%
\pgfsetstrokecolor{textcolor}%
\pgfsetfillcolor{textcolor}%
\pgftext[x=0.233333in, y=4.157111in, left, base]{\color{textcolor}\sffamily\fontsize{14.000000}{16.800000}\selectfont \(\displaystyle {\ensuremath{-}35}\)}%
\end{pgfscope}%
\begin{pgfscope}%
\definecolor{textcolor}{rgb}{0.000000,0.000000,0.000000}%
\pgfsetstrokecolor{textcolor}%
\pgfsetfillcolor{textcolor}%
\pgftext[x=0.177777in,y=2.378555in,,bottom,rotate=90.000000]{\color{textcolor}\sffamily\fontsize{14.000000}{16.800000}\selectfont Utility}%
\end{pgfscope}%
\begin{pgfscope}%
\pgfpathrectangle{\pgfqpoint{0.681942in}{0.530555in}}{\pgfqpoint{4.960000in}{3.696000in}}%
\pgfusepath{clip}%
\pgfsetrectcap%
\pgfsetroundjoin%
\pgfsetlinewidth{1.505625pt}%
\definecolor{currentstroke}{rgb}{0.121569,0.466667,0.705882}%
\pgfsetstrokecolor{currentstroke}%
\pgfsetdash{}{0pt}%
\pgfpathmoveto{\pgfqpoint{0.907396in}{3.709769in}}%
\pgfpathlineto{\pgfqpoint{3.859294in}{3.563907in}}%
\pgfpathlineto{\pgfqpoint{5.416487in}{3.487894in}}%
\pgfpathlineto{\pgfqpoint{5.416487in}{3.487894in}}%
\pgfusepath{stroke}%
\end{pgfscope}%
\begin{pgfscope}%
\pgfpathrectangle{\pgfqpoint{0.681942in}{0.530555in}}{\pgfqpoint{4.960000in}{3.696000in}}%
\pgfusepath{clip}%
\pgfsetbuttcap%
\pgfsetroundjoin%
\definecolor{currentfill}{rgb}{0.121569,0.466667,0.705882}%
\pgfsetfillcolor{currentfill}%
\pgfsetlinewidth{1.003750pt}%
\definecolor{currentstroke}{rgb}{0.121569,0.466667,0.705882}%
\pgfsetstrokecolor{currentstroke}%
\pgfsetdash{}{0pt}%
\pgfsys@defobject{currentmarker}{\pgfqpoint{-0.041667in}{-0.041667in}}{\pgfqpoint{0.041667in}{0.041667in}}{%
\pgfpathmoveto{\pgfqpoint{-0.041667in}{-0.041667in}}%
\pgfpathlineto{\pgfqpoint{0.041667in}{0.041667in}}%
\pgfpathmoveto{\pgfqpoint{-0.041667in}{0.041667in}}%
\pgfpathlineto{\pgfqpoint{0.041667in}{-0.041667in}}%
\pgfusepath{stroke,fill}%
}%
\begin{pgfscope}%
\pgfsys@transformshift{0.907396in}{3.709769in}%
\pgfsys@useobject{currentmarker}{}%
\end{pgfscope}%
\begin{pgfscope}%
\pgfsys@transformshift{2.144124in}{3.648386in}%
\pgfsys@useobject{currentmarker}{}%
\end{pgfscope}%
\begin{pgfscope}%
\pgfsys@transformshift{3.380852in}{3.587394in}%
\pgfsys@useobject{currentmarker}{}%
\end{pgfscope}%
\begin{pgfscope}%
\pgfsys@transformshift{4.613066in}{3.527029in}%
\pgfsys@useobject{currentmarker}{}%
\end{pgfscope}%
\end{pgfscope}%
\begin{pgfscope}%
\pgfpathrectangle{\pgfqpoint{0.681942in}{0.530555in}}{\pgfqpoint{4.960000in}{3.696000in}}%
\pgfusepath{clip}%
\pgfsetrectcap%
\pgfsetroundjoin%
\pgfsetlinewidth{1.505625pt}%
\definecolor{currentstroke}{rgb}{1.000000,0.498039,0.054902}%
\pgfsetstrokecolor{currentstroke}%
\pgfsetdash{}{0pt}%
\pgfpathmoveto{\pgfqpoint{0.907396in}{2.674324in}}%
\pgfpathlineto{\pgfqpoint{0.916424in}{2.670434in}}%
\pgfpathlineto{\pgfqpoint{0.925451in}{2.665284in}}%
\pgfpathlineto{\pgfqpoint{0.979614in}{2.642806in}}%
\pgfpathlineto{\pgfqpoint{0.984128in}{2.641444in}}%
\pgfpathlineto{\pgfqpoint{0.988641in}{2.613552in}}%
\pgfpathlineto{\pgfqpoint{0.993155in}{2.592933in}}%
\pgfpathlineto{\pgfqpoint{0.997669in}{2.578026in}}%
\pgfpathlineto{\pgfqpoint{1.002182in}{2.569770in}}%
\pgfpathlineto{\pgfqpoint{1.015723in}{2.556452in}}%
\pgfpathlineto{\pgfqpoint{1.029264in}{2.545080in}}%
\pgfpathlineto{\pgfqpoint{1.074400in}{2.517405in}}%
\pgfpathlineto{\pgfqpoint{1.105995in}{2.501733in}}%
\pgfpathlineto{\pgfqpoint{1.110509in}{2.485626in}}%
\pgfpathlineto{\pgfqpoint{1.115022in}{2.452863in}}%
\pgfpathlineto{\pgfqpoint{1.124050in}{2.417643in}}%
\pgfpathlineto{\pgfqpoint{1.128563in}{2.407492in}}%
\pgfpathlineto{\pgfqpoint{1.137590in}{2.397194in}}%
\pgfpathlineto{\pgfqpoint{1.155645in}{2.380353in}}%
\pgfpathlineto{\pgfqpoint{1.196267in}{2.324052in}}%
\pgfpathlineto{\pgfqpoint{1.205294in}{2.313206in}}%
\pgfpathlineto{\pgfqpoint{1.209808in}{2.300702in}}%
\pgfpathlineto{\pgfqpoint{1.214322in}{2.253418in}}%
\pgfpathlineto{\pgfqpoint{1.223349in}{2.205723in}}%
\pgfpathlineto{\pgfqpoint{1.227862in}{2.188081in}}%
\pgfpathlineto{\pgfqpoint{1.232376in}{2.157804in}}%
\pgfpathlineto{\pgfqpoint{1.236890in}{2.107104in}}%
\pgfpathlineto{\pgfqpoint{1.245917in}{2.052526in}}%
\pgfpathlineto{\pgfqpoint{1.254944in}{2.029953in}}%
\pgfpathlineto{\pgfqpoint{1.268485in}{2.000535in}}%
\pgfpathlineto{\pgfqpoint{1.282026in}{1.974686in}}%
\pgfpathlineto{\pgfqpoint{1.286539in}{1.942041in}}%
\pgfpathlineto{\pgfqpoint{1.291053in}{1.896508in}}%
\pgfpathlineto{\pgfqpoint{1.300080in}{1.842633in}}%
\pgfpathlineto{\pgfqpoint{1.309107in}{1.800155in}}%
\pgfpathlineto{\pgfqpoint{1.313621in}{1.783285in}}%
\pgfpathlineto{\pgfqpoint{1.318135in}{1.726630in}}%
\pgfpathlineto{\pgfqpoint{1.327162in}{1.658749in}}%
\pgfpathlineto{\pgfqpoint{1.336189in}{1.610785in}}%
\pgfpathlineto{\pgfqpoint{1.345216in}{1.514363in}}%
\pgfpathlineto{\pgfqpoint{1.354243in}{1.455287in}}%
\pgfpathlineto{\pgfqpoint{1.363271in}{1.411431in}}%
\pgfpathlineto{\pgfqpoint{1.367784in}{1.339605in}}%
\pgfpathlineto{\pgfqpoint{1.372298in}{1.296440in}}%
\pgfpathlineto{\pgfqpoint{1.376811in}{1.217166in}}%
\pgfpathlineto{\pgfqpoint{1.385839in}{1.135674in}}%
\pgfpathlineto{\pgfqpoint{1.390352in}{1.038764in}}%
\pgfpathlineto{\pgfqpoint{1.394866in}{0.966783in}}%
\pgfpathlineto{\pgfqpoint{1.399379in}{0.918969in}}%
\pgfpathlineto{\pgfqpoint{1.408407in}{0.858517in}}%
\pgfpathlineto{\pgfqpoint{1.412920in}{0.835955in}}%
\pgfpathlineto{\pgfqpoint{1.421947in}{0.807986in}}%
\pgfpathlineto{\pgfqpoint{1.430975in}{0.787628in}}%
\pgfpathlineto{\pgfqpoint{1.435488in}{0.781951in}}%
\pgfpathlineto{\pgfqpoint{1.440002in}{0.778862in}}%
\pgfpathlineto{\pgfqpoint{1.440015in}{0.520555in}}%
\pgfpathlineto{\pgfqpoint{1.440015in}{0.520555in}}%
\pgfusepath{stroke}%
\end{pgfscope}%
\begin{pgfscope}%
\pgfpathrectangle{\pgfqpoint{0.681942in}{0.530555in}}{\pgfqpoint{4.960000in}{3.696000in}}%
\pgfusepath{clip}%
\pgfsetbuttcap%
\pgfsetmiterjoin%
\definecolor{currentfill}{rgb}{1.000000,0.498039,0.054902}%
\pgfsetfillcolor{currentfill}%
\pgfsetlinewidth{1.003750pt}%
\definecolor{currentstroke}{rgb}{1.000000,0.498039,0.054902}%
\pgfsetstrokecolor{currentstroke}%
\pgfsetdash{}{0pt}%
\pgfsys@defobject{currentmarker}{\pgfqpoint{-0.041667in}{-0.041667in}}{\pgfqpoint{0.041667in}{0.041667in}}{%
\pgfpathmoveto{\pgfqpoint{0.000000in}{0.041667in}}%
\pgfpathlineto{\pgfqpoint{-0.041667in}{-0.041667in}}%
\pgfpathlineto{\pgfqpoint{0.041667in}{-0.041667in}}%
\pgfpathlineto{\pgfqpoint{0.000000in}{0.041667in}}%
\pgfpathclose%
\pgfusepath{stroke,fill}%
}%
\begin{pgfscope}%
\pgfsys@transformshift{0.907396in}{2.674324in}%
\pgfsys@useobject{currentmarker}{}%
\end{pgfscope}%
\begin{pgfscope}%
\pgfsys@transformshift{1.336189in}{1.610785in}%
\pgfsys@useobject{currentmarker}{}%
\end{pgfscope}%
\begin{pgfscope}%
\pgfsys@transformshift{1.440002in}{0.778862in}%
\pgfsys@useobject{currentmarker}{}%
\end{pgfscope}%
\begin{pgfscope}%
\pgfsys@transformshift{1.444515in}{-84.939445in}%
\pgfsys@useobject{currentmarker}{}%
\end{pgfscope}%
\begin{pgfscope}%
\pgfsys@transformshift{1.498679in}{-84.939445in}%
\pgfsys@useobject{currentmarker}{}%
\end{pgfscope}%
\begin{pgfscope}%
\pgfsys@transformshift{2.735406in}{-84.939445in}%
\pgfsys@useobject{currentmarker}{}%
\end{pgfscope}%
\begin{pgfscope}%
\pgfsys@transformshift{3.972134in}{-84.939445in}%
\pgfsys@useobject{currentmarker}{}%
\end{pgfscope}%
\begin{pgfscope}%
\pgfsys@transformshift{5.208862in}{-84.939445in}%
\pgfsys@useobject{currentmarker}{}%
\end{pgfscope}%
\end{pgfscope}%
\begin{pgfscope}%
\pgfpathrectangle{\pgfqpoint{0.681942in}{0.530555in}}{\pgfqpoint{4.960000in}{3.696000in}}%
\pgfusepath{clip}%
\pgfsetrectcap%
\pgfsetroundjoin%
\pgfsetlinewidth{1.505625pt}%
\definecolor{currentstroke}{rgb}{0.172549,0.627451,0.172549}%
\pgfsetstrokecolor{currentstroke}%
\pgfsetdash{}{0pt}%
\pgfpathmoveto{\pgfqpoint{0.907396in}{3.700375in}}%
\pgfpathlineto{\pgfqpoint{1.227862in}{3.683622in}}%
\pgfpathlineto{\pgfqpoint{1.304594in}{3.677946in}}%
\pgfpathlineto{\pgfqpoint{1.358757in}{3.674310in}}%
\pgfpathlineto{\pgfqpoint{1.376811in}{3.671135in}}%
\pgfpathlineto{\pgfqpoint{1.512219in}{3.656615in}}%
\pgfpathlineto{\pgfqpoint{1.548328in}{3.653217in}}%
\pgfpathlineto{\pgfqpoint{1.566383in}{3.650232in}}%
\pgfpathlineto{\pgfqpoint{1.602492in}{3.644337in}}%
\pgfpathlineto{\pgfqpoint{1.638600in}{3.639264in}}%
\pgfpathlineto{\pgfqpoint{1.647628in}{3.635021in}}%
\pgfpathlineto{\pgfqpoint{1.674709in}{3.627428in}}%
\pgfpathlineto{\pgfqpoint{1.706304in}{3.619904in}}%
\pgfpathlineto{\pgfqpoint{1.760468in}{3.608438in}}%
\pgfpathlineto{\pgfqpoint{1.828172in}{3.595659in}}%
\pgfpathlineto{\pgfqpoint{1.931985in}{3.580952in}}%
\pgfpathlineto{\pgfqpoint{1.945526in}{3.572996in}}%
\pgfpathlineto{\pgfqpoint{1.963580in}{3.566130in}}%
\pgfpathlineto{\pgfqpoint{2.017743in}{3.551996in}}%
\pgfpathlineto{\pgfqpoint{2.094474in}{3.535108in}}%
\pgfpathlineto{\pgfqpoint{2.103502in}{3.533542in}}%
\pgfpathlineto{\pgfqpoint{2.166692in}{3.526307in}}%
\pgfpathlineto{\pgfqpoint{2.193774in}{3.523768in}}%
\pgfpathlineto{\pgfqpoint{2.207315in}{3.513314in}}%
\pgfpathlineto{\pgfqpoint{2.234396in}{3.501971in}}%
\pgfpathlineto{\pgfqpoint{2.275019in}{3.489539in}}%
\pgfpathlineto{\pgfqpoint{2.293073in}{3.485219in}}%
\pgfpathlineto{\pgfqpoint{2.306614in}{3.481731in}}%
\pgfpathlineto{\pgfqpoint{2.329182in}{3.477402in}}%
\pgfpathlineto{\pgfqpoint{2.374318in}{3.471918in}}%
\pgfpathlineto{\pgfqpoint{2.410427in}{3.467699in}}%
\pgfpathlineto{\pgfqpoint{2.419454in}{3.457935in}}%
\pgfpathlineto{\pgfqpoint{2.432995in}{3.449002in}}%
\pgfpathlineto{\pgfqpoint{2.460076in}{3.437415in}}%
\pgfpathlineto{\pgfqpoint{2.473617in}{3.432666in}}%
\pgfpathlineto{\pgfqpoint{2.518753in}{3.419077in}}%
\pgfpathlineto{\pgfqpoint{2.577430in}{3.411208in}}%
\pgfpathlineto{\pgfqpoint{2.581944in}{3.406653in}}%
\pgfpathlineto{\pgfqpoint{2.586457in}{3.399978in}}%
\pgfpathlineto{\pgfqpoint{2.595485in}{3.392009in}}%
\pgfpathlineto{\pgfqpoint{2.631593in}{3.372711in}}%
\pgfpathlineto{\pgfqpoint{2.654161in}{3.363975in}}%
\pgfpathlineto{\pgfqpoint{2.667702in}{3.357984in}}%
\pgfpathlineto{\pgfqpoint{2.708325in}{3.351528in}}%
\pgfpathlineto{\pgfqpoint{2.712838in}{3.341616in}}%
\pgfpathlineto{\pgfqpoint{2.721866in}{3.331722in}}%
\pgfpathlineto{\pgfqpoint{2.735406in}{3.321290in}}%
\pgfpathlineto{\pgfqpoint{2.748947in}{3.313358in}}%
\pgfpathlineto{\pgfqpoint{2.794083in}{3.292354in}}%
\pgfpathlineto{\pgfqpoint{2.803110in}{3.290871in}}%
\pgfpathlineto{\pgfqpoint{2.807624in}{3.288081in}}%
\pgfpathlineto{\pgfqpoint{2.812138in}{3.276450in}}%
\pgfpathlineto{\pgfqpoint{2.821165in}{3.266355in}}%
\pgfpathlineto{\pgfqpoint{2.834706in}{3.255442in}}%
\pgfpathlineto{\pgfqpoint{2.852760in}{3.243707in}}%
\pgfpathlineto{\pgfqpoint{2.866301in}{3.235693in}}%
\pgfpathlineto{\pgfqpoint{2.879842in}{3.225515in}}%
\pgfpathlineto{\pgfqpoint{2.884355in}{3.222619in}}%
\pgfpathlineto{\pgfqpoint{2.888869in}{3.209289in}}%
\pgfpathlineto{\pgfqpoint{2.897896in}{3.194469in}}%
\pgfpathlineto{\pgfqpoint{2.911437in}{3.179075in}}%
\pgfpathlineto{\pgfqpoint{2.929491in}{3.162470in}}%
\pgfpathlineto{\pgfqpoint{2.938519in}{3.154908in}}%
\pgfpathlineto{\pgfqpoint{2.952059in}{3.144628in}}%
\pgfpathlineto{\pgfqpoint{2.970114in}{3.081260in}}%
\pgfpathlineto{\pgfqpoint{2.988168in}{3.054504in}}%
\pgfpathlineto{\pgfqpoint{3.006223in}{3.033064in}}%
\pgfpathlineto{\pgfqpoint{3.015250in}{3.024466in}}%
\pgfpathlineto{\pgfqpoint{3.019763in}{3.005675in}}%
\pgfpathlineto{\pgfqpoint{3.028791in}{2.987119in}}%
\pgfpathlineto{\pgfqpoint{3.042331in}{2.966961in}}%
\pgfpathlineto{\pgfqpoint{3.073927in}{2.931319in}}%
\pgfpathlineto{\pgfqpoint{3.078440in}{2.921873in}}%
\pgfpathlineto{\pgfqpoint{3.087467in}{2.867362in}}%
\pgfpathlineto{\pgfqpoint{3.091981in}{2.847663in}}%
\pgfpathlineto{\pgfqpoint{3.096495in}{2.833608in}}%
\pgfpathlineto{\pgfqpoint{3.110035in}{2.806394in}}%
\pgfpathlineto{\pgfqpoint{3.128090in}{2.776148in}}%
\pgfpathlineto{\pgfqpoint{3.137117in}{2.767328in}}%
\pgfpathlineto{\pgfqpoint{3.150658in}{2.718810in}}%
\pgfpathlineto{\pgfqpoint{3.164199in}{2.687076in}}%
\pgfpathlineto{\pgfqpoint{3.168712in}{2.677835in}}%
\pgfpathlineto{\pgfqpoint{3.182253in}{2.599843in}}%
\pgfpathlineto{\pgfqpoint{3.186767in}{2.579917in}}%
\pgfpathlineto{\pgfqpoint{3.191280in}{2.566971in}}%
\pgfpathlineto{\pgfqpoint{3.200308in}{2.551873in}}%
\pgfpathlineto{\pgfqpoint{3.204821in}{2.517330in}}%
\pgfpathlineto{\pgfqpoint{3.213848in}{2.478748in}}%
\pgfpathlineto{\pgfqpoint{3.222876in}{2.450538in}}%
\pgfpathlineto{\pgfqpoint{3.231903in}{2.423719in}}%
\pgfpathlineto{\pgfqpoint{3.236416in}{2.411506in}}%
\pgfpathlineto{\pgfqpoint{3.245444in}{2.335849in}}%
\pgfpathlineto{\pgfqpoint{3.254471in}{2.293845in}}%
\pgfpathlineto{\pgfqpoint{3.263498in}{2.208646in}}%
\pgfpathlineto{\pgfqpoint{3.272525in}{2.157774in}}%
\pgfpathlineto{\pgfqpoint{3.286066in}{2.105347in}}%
\pgfpathlineto{\pgfqpoint{3.299607in}{1.976927in}}%
\pgfpathlineto{\pgfqpoint{3.304120in}{1.951492in}}%
\pgfpathlineto{\pgfqpoint{3.308634in}{1.879708in}}%
\pgfpathlineto{\pgfqpoint{3.317661in}{1.797306in}}%
\pgfpathlineto{\pgfqpoint{3.322175in}{1.765427in}}%
\pgfpathlineto{\pgfqpoint{3.326689in}{1.743360in}}%
\pgfpathlineto{\pgfqpoint{3.331202in}{1.662456in}}%
\pgfpathlineto{\pgfqpoint{3.335716in}{1.618518in}}%
\pgfpathlineto{\pgfqpoint{3.349257in}{1.430937in}}%
\pgfpathlineto{\pgfqpoint{3.362797in}{1.166554in}}%
\pgfpathlineto{\pgfqpoint{3.367311in}{1.053838in}}%
\pgfpathlineto{\pgfqpoint{3.376338in}{0.921536in}}%
\pgfpathlineto{\pgfqpoint{3.385365in}{0.823559in}}%
\pgfpathlineto{\pgfqpoint{3.394393in}{0.758443in}}%
\pgfpathlineto{\pgfqpoint{3.403420in}{0.714850in}}%
\pgfpathlineto{\pgfqpoint{3.407933in}{0.695238in}}%
\pgfpathlineto{\pgfqpoint{3.412447in}{0.680973in}}%
\pgfpathlineto{\pgfqpoint{3.416961in}{0.672513in}}%
\pgfpathlineto{\pgfqpoint{3.421474in}{0.669701in}}%
\pgfpathlineto{\pgfqpoint{3.561396in}{0.668618in}}%
\pgfpathlineto{\pgfqpoint{3.570423in}{0.667896in}}%
\pgfpathlineto{\pgfqpoint{3.606532in}{0.667797in}}%
\pgfpathlineto{\pgfqpoint{3.620073in}{0.657373in}}%
\pgfpathlineto{\pgfqpoint{3.638127in}{0.651145in}}%
\pgfpathlineto{\pgfqpoint{3.647154in}{0.647705in}}%
\pgfpathlineto{\pgfqpoint{3.656182in}{0.647000in}}%
\pgfpathlineto{\pgfqpoint{3.665209in}{0.644253in}}%
\pgfpathlineto{\pgfqpoint{3.669722in}{0.644242in}}%
\pgfpathlineto{\pgfqpoint{3.687777in}{0.639448in}}%
\pgfpathlineto{\pgfqpoint{3.787076in}{0.625083in}}%
\pgfpathlineto{\pgfqpoint{3.800617in}{0.624384in}}%
\pgfpathlineto{\pgfqpoint{3.836726in}{0.619631in}}%
\pgfpathlineto{\pgfqpoint{3.845753in}{0.618945in}}%
\pgfpathlineto{\pgfqpoint{3.850267in}{0.617604in}}%
\pgfpathlineto{\pgfqpoint{3.854780in}{0.617594in}}%
\pgfpathlineto{\pgfqpoint{3.859294in}{0.616254in}}%
\pgfpathlineto{\pgfqpoint{3.868321in}{0.616899in}}%
\pgfpathlineto{\pgfqpoint{3.886375in}{0.614867in}}%
\pgfpathlineto{\pgfqpoint{3.890889in}{0.614857in}}%
\pgfpathlineto{\pgfqpoint{3.899916in}{0.612846in}}%
\pgfpathlineto{\pgfqpoint{3.904430in}{0.612837in}}%
\pgfpathlineto{\pgfqpoint{3.908943in}{0.611500in}}%
\pgfpathlineto{\pgfqpoint{3.922484in}{0.610809in}}%
\pgfpathlineto{\pgfqpoint{3.936025in}{0.609456in}}%
\pgfpathlineto{\pgfqpoint{3.945052in}{0.610100in}}%
\pgfpathlineto{\pgfqpoint{3.954080in}{0.609419in}}%
\pgfpathlineto{\pgfqpoint{3.981161in}{0.608701in}}%
\pgfpathlineto{\pgfqpoint{3.990188in}{0.608021in}}%
\pgfpathlineto{\pgfqpoint{4.044352in}{0.607250in}}%
\pgfpathlineto{\pgfqpoint{4.053379in}{0.606571in}}%
\pgfpathlineto{\pgfqpoint{4.116569in}{0.605783in}}%
\pgfpathlineto{\pgfqpoint{4.125597in}{0.605104in}}%
\pgfpathlineto{\pgfqpoint{4.166219in}{0.604363in}}%
\pgfpathlineto{\pgfqpoint{4.175246in}{0.603686in}}%
\pgfpathlineto{\pgfqpoint{4.184273in}{0.604327in}}%
\pgfpathlineto{\pgfqpoint{4.220382in}{0.603596in}}%
\pgfpathlineto{\pgfqpoint{4.229409in}{0.602920in}}%
\pgfpathlineto{\pgfqpoint{4.301627in}{0.602118in}}%
\pgfpathlineto{\pgfqpoint{4.310654in}{0.601443in}}%
\pgfpathlineto{\pgfqpoint{4.391899in}{0.600626in}}%
\pgfpathlineto{\pgfqpoint{4.400926in}{0.599951in}}%
\pgfpathlineto{\pgfqpoint{4.491198in}{0.599119in}}%
\pgfpathlineto{\pgfqpoint{4.500226in}{0.598446in}}%
\pgfpathlineto{\pgfqpoint{4.595011in}{0.597607in}}%
\pgfpathlineto{\pgfqpoint{4.604039in}{0.596935in}}%
\pgfpathlineto{\pgfqpoint{4.707851in}{0.596082in}}%
\pgfpathlineto{\pgfqpoint{4.716879in}{0.595411in}}%
\pgfpathlineto{\pgfqpoint{4.725906in}{0.596048in}}%
\pgfpathlineto{\pgfqpoint{4.784583in}{0.595282in}}%
\pgfpathlineto{\pgfqpoint{4.793610in}{0.594612in}}%
\pgfpathlineto{\pgfqpoint{4.910964in}{0.593737in}}%
\pgfpathlineto{\pgfqpoint{4.915477in}{0.593076in}}%
\pgfpathlineto{\pgfqpoint{4.924505in}{0.593712in}}%
\pgfpathlineto{\pgfqpoint{4.996722in}{0.592924in}}%
\pgfpathlineto{\pgfqpoint{5.005749in}{0.592256in}}%
\pgfpathlineto{\pgfqpoint{5.186294in}{0.591268in}}%
\pgfpathlineto{\pgfqpoint{5.195321in}{0.590601in}}%
\pgfpathlineto{\pgfqpoint{5.335243in}{0.589694in}}%
\pgfpathlineto{\pgfqpoint{5.344270in}{0.589028in}}%
\pgfpathlineto{\pgfqpoint{5.416487in}{0.588896in}}%
\pgfpathlineto{\pgfqpoint{5.416487in}{0.588896in}}%
\pgfusepath{stroke}%
\end{pgfscope}%
\begin{pgfscope}%
\pgfpathrectangle{\pgfqpoint{0.681942in}{0.530555in}}{\pgfqpoint{4.960000in}{3.696000in}}%
\pgfusepath{clip}%
\pgfsetbuttcap%
\pgfsetmiterjoin%
\definecolor{currentfill}{rgb}{0.172549,0.627451,0.172549}%
\pgfsetfillcolor{currentfill}%
\pgfsetlinewidth{1.003750pt}%
\definecolor{currentstroke}{rgb}{0.172549,0.627451,0.172549}%
\pgfsetstrokecolor{currentstroke}%
\pgfsetdash{}{0pt}%
\pgfsys@defobject{currentmarker}{\pgfqpoint{-0.035355in}{-0.058926in}}{\pgfqpoint{0.035355in}{0.058926in}}{%
\pgfpathmoveto{\pgfqpoint{-0.000000in}{-0.058926in}}%
\pgfpathlineto{\pgfqpoint{0.035355in}{0.000000in}}%
\pgfpathlineto{\pgfqpoint{0.000000in}{0.058926in}}%
\pgfpathlineto{\pgfqpoint{-0.035355in}{0.000000in}}%
\pgfpathlineto{\pgfqpoint{-0.000000in}{-0.058926in}}%
\pgfpathclose%
\pgfusepath{stroke,fill}%
}%
\begin{pgfscope}%
\pgfsys@transformshift{0.907396in}{3.700375in}%
\pgfsys@useobject{currentmarker}{}%
\end{pgfscope}%
\begin{pgfscope}%
\pgfsys@transformshift{2.126070in}{3.530608in}%
\pgfsys@useobject{currentmarker}{}%
\end{pgfscope}%
\begin{pgfscope}%
\pgfsys@transformshift{3.082954in}{2.893890in}%
\pgfsys@useobject{currentmarker}{}%
\end{pgfscope}%
\begin{pgfscope}%
\pgfsys@transformshift{3.331202in}{1.662456in}%
\pgfsys@useobject{currentmarker}{}%
\end{pgfscope}%
\begin{pgfscope}%
\pgfsys@transformshift{3.620073in}{0.657373in}%
\pgfsys@useobject{currentmarker}{}%
\end{pgfscope}%
\begin{pgfscope}%
\pgfsys@transformshift{4.847773in}{0.594509in}%
\pgfsys@useobject{currentmarker}{}%
\end{pgfscope}%
\end{pgfscope}%
\begin{pgfscope}%
\pgfsetrectcap%
\pgfsetmiterjoin%
\pgfsetlinewidth{0.803000pt}%
\definecolor{currentstroke}{rgb}{0.000000,0.000000,0.000000}%
\pgfsetstrokecolor{currentstroke}%
\pgfsetdash{}{0pt}%
\pgfpathmoveto{\pgfqpoint{0.681942in}{0.530555in}}%
\pgfpathlineto{\pgfqpoint{0.681942in}{4.226555in}}%
\pgfusepath{stroke}%
\end{pgfscope}%
\begin{pgfscope}%
\pgfsetrectcap%
\pgfsetmiterjoin%
\pgfsetlinewidth{0.803000pt}%
\definecolor{currentstroke}{rgb}{0.000000,0.000000,0.000000}%
\pgfsetstrokecolor{currentstroke}%
\pgfsetdash{}{0pt}%
\pgfpathmoveto{\pgfqpoint{5.641942in}{0.530555in}}%
\pgfpathlineto{\pgfqpoint{5.641942in}{4.226555in}}%
\pgfusepath{stroke}%
\end{pgfscope}%
\begin{pgfscope}%
\pgfsetrectcap%
\pgfsetmiterjoin%
\pgfsetlinewidth{0.803000pt}%
\definecolor{currentstroke}{rgb}{0.000000,0.000000,0.000000}%
\pgfsetstrokecolor{currentstroke}%
\pgfsetdash{}{0pt}%
\pgfpathmoveto{\pgfqpoint{0.681942in}{0.530555in}}%
\pgfpathlineto{\pgfqpoint{5.641942in}{0.530555in}}%
\pgfusepath{stroke}%
\end{pgfscope}%
\begin{pgfscope}%
\pgfsetrectcap%
\pgfsetmiterjoin%
\pgfsetlinewidth{0.803000pt}%
\definecolor{currentstroke}{rgb}{0.000000,0.000000,0.000000}%
\pgfsetstrokecolor{currentstroke}%
\pgfsetdash{}{0pt}%
\pgfpathmoveto{\pgfqpoint{0.681942in}{4.226555in}}%
\pgfpathlineto{\pgfqpoint{5.641942in}{4.226555in}}%
\pgfusepath{stroke}%
\end{pgfscope}%
\begin{pgfscope}%
\pgfsetbuttcap%
\pgfsetmiterjoin%
\definecolor{currentfill}{rgb}{1.000000,1.000000,1.000000}%
\pgfsetfillcolor{currentfill}%
\pgfsetfillopacity{0.800000}%
\pgfsetlinewidth{1.003750pt}%
\definecolor{currentstroke}{rgb}{0.800000,0.800000,0.800000}%
\pgfsetstrokecolor{currentstroke}%
\pgfsetstrokeopacity{0.800000}%
\pgfsetdash{}{0pt}%
\pgfpathmoveto{\pgfqpoint{0.818053in}{0.627777in}}%
\pgfpathlineto{\pgfqpoint{3.401823in}{0.627777in}}%
\pgfpathquadraticcurveto{\pgfqpoint{3.440712in}{0.627777in}}{\pgfqpoint{3.440712in}{0.666666in}}%
\pgfpathlineto{\pgfqpoint{3.440712in}{1.472220in}}%
\pgfpathquadraticcurveto{\pgfqpoint{3.440712in}{1.511109in}}{\pgfqpoint{3.401823in}{1.511109in}}%
\pgfpathlineto{\pgfqpoint{0.818053in}{1.511109in}}%
\pgfpathquadraticcurveto{\pgfqpoint{0.779164in}{1.511109in}}{\pgfqpoint{0.779164in}{1.472220in}}%
\pgfpathlineto{\pgfqpoint{0.779164in}{0.666666in}}%
\pgfpathquadraticcurveto{\pgfqpoint{0.779164in}{0.627777in}}{\pgfqpoint{0.818053in}{0.627777in}}%
\pgfpathlineto{\pgfqpoint{0.818053in}{0.627777in}}%
\pgfpathclose%
\pgfusepath{stroke,fill}%
\end{pgfscope}%
\begin{pgfscope}%
\pgfsetrectcap%
\pgfsetroundjoin%
\pgfsetlinewidth{1.505625pt}%
\definecolor{currentstroke}{rgb}{0.121569,0.466667,0.705882}%
\pgfsetstrokecolor{currentstroke}%
\pgfsetdash{}{0pt}%
\pgfpathmoveto{\pgfqpoint{0.856942in}{1.362498in}}%
\pgfpathlineto{\pgfqpoint{1.051386in}{1.362498in}}%
\pgfpathlineto{\pgfqpoint{1.245831in}{1.362498in}}%
\pgfusepath{stroke}%
\end{pgfscope}%
\begin{pgfscope}%
\pgfsetbuttcap%
\pgfsetroundjoin%
\definecolor{currentfill}{rgb}{0.121569,0.466667,0.705882}%
\pgfsetfillcolor{currentfill}%
\pgfsetlinewidth{1.003750pt}%
\definecolor{currentstroke}{rgb}{0.121569,0.466667,0.705882}%
\pgfsetstrokecolor{currentstroke}%
\pgfsetdash{}{0pt}%
\pgfsys@defobject{currentmarker}{\pgfqpoint{-0.041667in}{-0.041667in}}{\pgfqpoint{0.041667in}{0.041667in}}{%
\pgfpathmoveto{\pgfqpoint{-0.041667in}{-0.041667in}}%
\pgfpathlineto{\pgfqpoint{0.041667in}{0.041667in}}%
\pgfpathmoveto{\pgfqpoint{-0.041667in}{0.041667in}}%
\pgfpathlineto{\pgfqpoint{0.041667in}{-0.041667in}}%
\pgfusepath{stroke,fill}%
}%
\begin{pgfscope}%
\pgfsys@transformshift{1.051386in}{1.362498in}%
\pgfsys@useobject{currentmarker}{}%
\end{pgfscope}%
\end{pgfscope}%
\begin{pgfscope}%
\definecolor{textcolor}{rgb}{0.000000,0.000000,0.000000}%
\pgfsetstrokecolor{textcolor}%
\pgfsetfillcolor{textcolor}%
\pgftext[x=1.401386in,y=1.294443in,left,base]{\color{textcolor}\sffamily\fontsize{14.000000}{16.800000}\selectfont Optimum}%
\end{pgfscope}%
\begin{pgfscope}%
\pgfsetrectcap%
\pgfsetroundjoin%
\pgfsetlinewidth{1.505625pt}%
\definecolor{currentstroke}{rgb}{1.000000,0.498039,0.054902}%
\pgfsetstrokecolor{currentstroke}%
\pgfsetdash{}{0pt}%
\pgfpathmoveto{\pgfqpoint{0.856942in}{1.087499in}}%
\pgfpathlineto{\pgfqpoint{1.051386in}{1.087499in}}%
\pgfpathlineto{\pgfqpoint{1.245831in}{1.087499in}}%
\pgfusepath{stroke}%
\end{pgfscope}%
\begin{pgfscope}%
\pgfsetbuttcap%
\pgfsetmiterjoin%
\definecolor{currentfill}{rgb}{1.000000,0.498039,0.054902}%
\pgfsetfillcolor{currentfill}%
\pgfsetlinewidth{1.003750pt}%
\definecolor{currentstroke}{rgb}{1.000000,0.498039,0.054902}%
\pgfsetstrokecolor{currentstroke}%
\pgfsetdash{}{0pt}%
\pgfsys@defobject{currentmarker}{\pgfqpoint{-0.041667in}{-0.041667in}}{\pgfqpoint{0.041667in}{0.041667in}}{%
\pgfpathmoveto{\pgfqpoint{0.000000in}{0.041667in}}%
\pgfpathlineto{\pgfqpoint{-0.041667in}{-0.041667in}}%
\pgfpathlineto{\pgfqpoint{0.041667in}{-0.041667in}}%
\pgfpathlineto{\pgfqpoint{0.000000in}{0.041667in}}%
\pgfpathclose%
\pgfusepath{stroke,fill}%
}%
\begin{pgfscope}%
\pgfsys@transformshift{1.051386in}{1.087499in}%
\pgfsys@useobject{currentmarker}{}%
\end{pgfscope}%
\end{pgfscope}%
\begin{pgfscope}%
\definecolor{textcolor}{rgb}{0.000000,0.000000,0.000000}%
\pgfsetstrokecolor{textcolor}%
\pgfsetfillcolor{textcolor}%
\pgftext[x=1.401386in,y=1.019443in,left,base]{\color{textcolor}\sffamily\fontsize{14.000000}{16.800000}\selectfont Utility without incentive}%
\end{pgfscope}%
\begin{pgfscope}%
\pgfsetrectcap%
\pgfsetroundjoin%
\pgfsetlinewidth{1.505625pt}%
\definecolor{currentstroke}{rgb}{0.172549,0.627451,0.172549}%
\pgfsetstrokecolor{currentstroke}%
\pgfsetdash{}{0pt}%
\pgfpathmoveto{\pgfqpoint{0.856942in}{0.812499in}}%
\pgfpathlineto{\pgfqpoint{1.051386in}{0.812499in}}%
\pgfpathlineto{\pgfqpoint{1.245831in}{0.812499in}}%
\pgfusepath{stroke}%
\end{pgfscope}%
\begin{pgfscope}%
\pgfsetbuttcap%
\pgfsetmiterjoin%
\definecolor{currentfill}{rgb}{0.172549,0.627451,0.172549}%
\pgfsetfillcolor{currentfill}%
\pgfsetlinewidth{1.003750pt}%
\definecolor{currentstroke}{rgb}{0.172549,0.627451,0.172549}%
\pgfsetstrokecolor{currentstroke}%
\pgfsetdash{}{0pt}%
\pgfsys@defobject{currentmarker}{\pgfqpoint{-0.035355in}{-0.058926in}}{\pgfqpoint{0.035355in}{0.058926in}}{%
\pgfpathmoveto{\pgfqpoint{-0.000000in}{-0.058926in}}%
\pgfpathlineto{\pgfqpoint{0.035355in}{0.000000in}}%
\pgfpathlineto{\pgfqpoint{0.000000in}{0.058926in}}%
\pgfpathlineto{\pgfqpoint{-0.035355in}{0.000000in}}%
\pgfpathlineto{\pgfqpoint{-0.000000in}{-0.058926in}}%
\pgfpathclose%
\pgfusepath{stroke,fill}%
}%
\begin{pgfscope}%
\pgfsys@transformshift{1.051386in}{0.812499in}%
\pgfsys@useobject{currentmarker}{}%
\end{pgfscope}%
\end{pgfscope}%
\begin{pgfscope}%
\definecolor{textcolor}{rgb}{0.000000,0.000000,0.000000}%
\pgfsetstrokecolor{textcolor}%
\pgfsetfillcolor{textcolor}%
\pgftext[x=1.401386in,y=0.744444in,left,base]{\color{textcolor}\sffamily\fontsize{14.000000}{16.800000}\selectfont Utility with incentive}%
\end{pgfscope}%
\end{pgfpicture}%
\makeatother%
\endgroup%

%% file: plots/poav2.pgf
\begingroup%
\makeatletter%
\begin{pgfpicture}%
\pgfpathrectangle{\pgfpointorigin}{\pgfqpoint{5.540783in}{4.295999in}}%
\pgfusepath{use as bounding box, clip}%
\begin{pgfscope}%
\pgfsetbuttcap%
\pgfsetmiterjoin%
\definecolor{currentfill}{rgb}{1.000000,1.000000,1.000000}%
\pgfsetfillcolor{currentfill}%
\pgfsetlinewidth{0.000000pt}%
\definecolor{currentstroke}{rgb}{1.000000,1.000000,1.000000}%
\pgfsetstrokecolor{currentstroke}%
\pgfsetdash{}{0pt}%
\pgfpathmoveto{\pgfqpoint{0.000000in}{0.000000in}}%
\pgfpathlineto{\pgfqpoint{5.540783in}{0.000000in}}%
\pgfpathlineto{\pgfqpoint{5.540783in}{4.295999in}}%
\pgfpathlineto{\pgfqpoint{0.000000in}{4.295999in}}%
\pgfpathlineto{\pgfqpoint{0.000000in}{0.000000in}}%
\pgfpathclose%
\pgfusepath{fill}%
\end{pgfscope}%
\begin{pgfscope}%
\pgfsetbuttcap%
\pgfsetmiterjoin%
\definecolor{currentfill}{rgb}{1.000000,1.000000,1.000000}%
\pgfsetfillcolor{currentfill}%
\pgfsetlinewidth{0.000000pt}%
\definecolor{currentstroke}{rgb}{0.000000,0.000000,0.000000}%
\pgfsetstrokecolor{currentstroke}%
\pgfsetstrokeopacity{0.000000}%
\pgfsetdash{}{0pt}%
\pgfpathmoveto{\pgfqpoint{0.580783in}{0.530555in}}%
\pgfpathlineto{\pgfqpoint{5.540783in}{0.530555in}}%
\pgfpathlineto{\pgfqpoint{5.540783in}{4.226555in}}%
\pgfpathlineto{\pgfqpoint{0.580783in}{4.226555in}}%
\pgfpathlineto{\pgfqpoint{0.580783in}{0.530555in}}%
\pgfpathclose%
\pgfusepath{fill}%
\end{pgfscope}%
\begin{pgfscope}%
\pgfpathrectangle{\pgfqpoint{0.580783in}{0.530555in}}{\pgfqpoint{4.960000in}{3.696000in}}%
\pgfusepath{clip}%
\pgfsetrectcap%
\pgfsetroundjoin%
\pgfsetlinewidth{0.803000pt}%
\definecolor{currentstroke}{rgb}{0.690196,0.690196,0.690196}%
\pgfsetstrokecolor{currentstroke}%
\pgfsetdash{}{0pt}%
\pgfpathmoveto{\pgfqpoint{0.806238in}{0.530555in}}%
\pgfpathlineto{\pgfqpoint{0.806238in}{4.226555in}}%
\pgfusepath{stroke}%
\end{pgfscope}%
\begin{pgfscope}%
\pgfsetbuttcap%
\pgfsetroundjoin%
\definecolor{currentfill}{rgb}{0.000000,0.000000,0.000000}%
\pgfsetfillcolor{currentfill}%
\pgfsetlinewidth{0.803000pt}%
\definecolor{currentstroke}{rgb}{0.000000,0.000000,0.000000}%
\pgfsetstrokecolor{currentstroke}%
\pgfsetdash{}{0pt}%
\pgfsys@defobject{currentmarker}{\pgfqpoint{0.000000in}{-0.048611in}}{\pgfqpoint{0.000000in}{0.000000in}}{%
\pgfpathmoveto{\pgfqpoint{0.000000in}{0.000000in}}%
\pgfpathlineto{\pgfqpoint{0.000000in}{-0.048611in}}%
\pgfusepath{stroke,fill}%
}%
\begin{pgfscope}%
\pgfsys@transformshift{0.806238in}{0.530555in}%
\pgfsys@useobject{currentmarker}{}%
\end{pgfscope}%
\end{pgfscope}%
\begin{pgfscope}%
\definecolor{textcolor}{rgb}{0.000000,0.000000,0.000000}%
\pgfsetstrokecolor{textcolor}%
\pgfsetfillcolor{textcolor}%
\pgftext[x=0.806238in,y=0.433333in,,top]{\color{textcolor}\sffamily\fontsize{14.000000}{16.800000}\selectfont \(\displaystyle {0.0}\)}%
\end{pgfscope}%
\begin{pgfscope}%
\pgfpathrectangle{\pgfqpoint{0.580783in}{0.530555in}}{\pgfqpoint{4.960000in}{3.696000in}}%
\pgfusepath{clip}%
\pgfsetrectcap%
\pgfsetroundjoin%
\pgfsetlinewidth{0.803000pt}%
\definecolor{currentstroke}{rgb}{0.690196,0.690196,0.690196}%
\pgfsetstrokecolor{currentstroke}%
\pgfsetdash{}{0pt}%
\pgfpathmoveto{\pgfqpoint{1.933511in}{0.530555in}}%
\pgfpathlineto{\pgfqpoint{1.933511in}{4.226555in}}%
\pgfusepath{stroke}%
\end{pgfscope}%
\begin{pgfscope}%
\pgfsetbuttcap%
\pgfsetroundjoin%
\definecolor{currentfill}{rgb}{0.000000,0.000000,0.000000}%
\pgfsetfillcolor{currentfill}%
\pgfsetlinewidth{0.803000pt}%
\definecolor{currentstroke}{rgb}{0.000000,0.000000,0.000000}%
\pgfsetstrokecolor{currentstroke}%
\pgfsetdash{}{0pt}%
\pgfsys@defobject{currentmarker}{\pgfqpoint{0.000000in}{-0.048611in}}{\pgfqpoint{0.000000in}{0.000000in}}{%
\pgfpathmoveto{\pgfqpoint{0.000000in}{0.000000in}}%
\pgfpathlineto{\pgfqpoint{0.000000in}{-0.048611in}}%
\pgfusepath{stroke,fill}%
}%
\begin{pgfscope}%
\pgfsys@transformshift{1.933511in}{0.530555in}%
\pgfsys@useobject{currentmarker}{}%
\end{pgfscope}%
\end{pgfscope}%
\begin{pgfscope}%
\definecolor{textcolor}{rgb}{0.000000,0.000000,0.000000}%
\pgfsetstrokecolor{textcolor}%
\pgfsetfillcolor{textcolor}%
\pgftext[x=1.933511in,y=0.433333in,,top]{\color{textcolor}\sffamily\fontsize{14.000000}{16.800000}\selectfont \(\displaystyle {0.5}\)}%
\end{pgfscope}%
\begin{pgfscope}%
\pgfpathrectangle{\pgfqpoint{0.580783in}{0.530555in}}{\pgfqpoint{4.960000in}{3.696000in}}%
\pgfusepath{clip}%
\pgfsetrectcap%
\pgfsetroundjoin%
\pgfsetlinewidth{0.803000pt}%
\definecolor{currentstroke}{rgb}{0.690196,0.690196,0.690196}%
\pgfsetstrokecolor{currentstroke}%
\pgfsetdash{}{0pt}%
\pgfpathmoveto{\pgfqpoint{3.060783in}{0.530555in}}%
\pgfpathlineto{\pgfqpoint{3.060783in}{4.226555in}}%
\pgfusepath{stroke}%
\end{pgfscope}%
\begin{pgfscope}%
\pgfsetbuttcap%
\pgfsetroundjoin%
\definecolor{currentfill}{rgb}{0.000000,0.000000,0.000000}%
\pgfsetfillcolor{currentfill}%
\pgfsetlinewidth{0.803000pt}%
\definecolor{currentstroke}{rgb}{0.000000,0.000000,0.000000}%
\pgfsetstrokecolor{currentstroke}%
\pgfsetdash{}{0pt}%
\pgfsys@defobject{currentmarker}{\pgfqpoint{0.000000in}{-0.048611in}}{\pgfqpoint{0.000000in}{0.000000in}}{%
\pgfpathmoveto{\pgfqpoint{0.000000in}{0.000000in}}%
\pgfpathlineto{\pgfqpoint{0.000000in}{-0.048611in}}%
\pgfusepath{stroke,fill}%
}%
\begin{pgfscope}%
\pgfsys@transformshift{3.060783in}{0.530555in}%
\pgfsys@useobject{currentmarker}{}%
\end{pgfscope}%
\end{pgfscope}%
\begin{pgfscope}%
\definecolor{textcolor}{rgb}{0.000000,0.000000,0.000000}%
\pgfsetstrokecolor{textcolor}%
\pgfsetfillcolor{textcolor}%
\pgftext[x=3.060783in,y=0.433333in,,top]{\color{textcolor}\sffamily\fontsize{14.000000}{16.800000}\selectfont \(\displaystyle {1.0}\)}%
\end{pgfscope}%
\begin{pgfscope}%
\pgfpathrectangle{\pgfqpoint{0.580783in}{0.530555in}}{\pgfqpoint{4.960000in}{3.696000in}}%
\pgfusepath{clip}%
\pgfsetrectcap%
\pgfsetroundjoin%
\pgfsetlinewidth{0.803000pt}%
\definecolor{currentstroke}{rgb}{0.690196,0.690196,0.690196}%
\pgfsetstrokecolor{currentstroke}%
\pgfsetdash{}{0pt}%
\pgfpathmoveto{\pgfqpoint{4.188056in}{0.530555in}}%
\pgfpathlineto{\pgfqpoint{4.188056in}{4.226555in}}%
\pgfusepath{stroke}%
\end{pgfscope}%
\begin{pgfscope}%
\pgfsetbuttcap%
\pgfsetroundjoin%
\definecolor{currentfill}{rgb}{0.000000,0.000000,0.000000}%
\pgfsetfillcolor{currentfill}%
\pgfsetlinewidth{0.803000pt}%
\definecolor{currentstroke}{rgb}{0.000000,0.000000,0.000000}%
\pgfsetstrokecolor{currentstroke}%
\pgfsetdash{}{0pt}%
\pgfsys@defobject{currentmarker}{\pgfqpoint{0.000000in}{-0.048611in}}{\pgfqpoint{0.000000in}{0.000000in}}{%
\pgfpathmoveto{\pgfqpoint{0.000000in}{0.000000in}}%
\pgfpathlineto{\pgfqpoint{0.000000in}{-0.048611in}}%
\pgfusepath{stroke,fill}%
}%
\begin{pgfscope}%
\pgfsys@transformshift{4.188056in}{0.530555in}%
\pgfsys@useobject{currentmarker}{}%
\end{pgfscope}%
\end{pgfscope}%
\begin{pgfscope}%
\definecolor{textcolor}{rgb}{0.000000,0.000000,0.000000}%
\pgfsetstrokecolor{textcolor}%
\pgfsetfillcolor{textcolor}%
\pgftext[x=4.188056in,y=0.433333in,,top]{\color{textcolor}\sffamily\fontsize{14.000000}{16.800000}\selectfont \(\displaystyle {1.5}\)}%
\end{pgfscope}%
\begin{pgfscope}%
\pgfpathrectangle{\pgfqpoint{0.580783in}{0.530555in}}{\pgfqpoint{4.960000in}{3.696000in}}%
\pgfusepath{clip}%
\pgfsetrectcap%
\pgfsetroundjoin%
\pgfsetlinewidth{0.803000pt}%
\definecolor{currentstroke}{rgb}{0.690196,0.690196,0.690196}%
\pgfsetstrokecolor{currentstroke}%
\pgfsetdash{}{0pt}%
\pgfpathmoveto{\pgfqpoint{5.315329in}{0.530555in}}%
\pgfpathlineto{\pgfqpoint{5.315329in}{4.226555in}}%
\pgfusepath{stroke}%
\end{pgfscope}%
\begin{pgfscope}%
\pgfsetbuttcap%
\pgfsetroundjoin%
\definecolor{currentfill}{rgb}{0.000000,0.000000,0.000000}%
\pgfsetfillcolor{currentfill}%
\pgfsetlinewidth{0.803000pt}%
\definecolor{currentstroke}{rgb}{0.000000,0.000000,0.000000}%
\pgfsetstrokecolor{currentstroke}%
\pgfsetdash{}{0pt}%
\pgfsys@defobject{currentmarker}{\pgfqpoint{0.000000in}{-0.048611in}}{\pgfqpoint{0.000000in}{0.000000in}}{%
\pgfpathmoveto{\pgfqpoint{0.000000in}{0.000000in}}%
\pgfpathlineto{\pgfqpoint{0.000000in}{-0.048611in}}%
\pgfusepath{stroke,fill}%
}%
\begin{pgfscope}%
\pgfsys@transformshift{5.315329in}{0.530555in}%
\pgfsys@useobject{currentmarker}{}%
\end{pgfscope}%
\end{pgfscope}%
\begin{pgfscope}%
\definecolor{textcolor}{rgb}{0.000000,0.000000,0.000000}%
\pgfsetstrokecolor{textcolor}%
\pgfsetfillcolor{textcolor}%
\pgftext[x=5.315329in,y=0.433333in,,top]{\color{textcolor}\sffamily\fontsize{14.000000}{16.800000}\selectfont \(\displaystyle {2.0}\)}%
\end{pgfscope}%
\begin{pgfscope}%
\definecolor{textcolor}{rgb}{0.000000,0.000000,0.000000}%
\pgfsetstrokecolor{textcolor}%
\pgfsetfillcolor{textcolor}%
\pgftext[x=3.060783in,y=0.200000in,,top]{\color{textcolor}\sffamily\fontsize{14.000000}{16.800000}\selectfont Cost (\(\displaystyle c\))}%
\end{pgfscope}%
\begin{pgfscope}%
\pgfpathrectangle{\pgfqpoint{0.580783in}{0.530555in}}{\pgfqpoint{4.960000in}{3.696000in}}%
\pgfusepath{clip}%
\pgfsetrectcap%
\pgfsetroundjoin%
\pgfsetlinewidth{0.803000pt}%
\definecolor{currentstroke}{rgb}{0.690196,0.690196,0.690196}%
\pgfsetstrokecolor{currentstroke}%
\pgfsetdash{}{0pt}%
\pgfpathmoveto{\pgfqpoint{0.580783in}{0.866555in}}%
\pgfpathlineto{\pgfqpoint{5.540783in}{0.866555in}}%
\pgfusepath{stroke}%
\end{pgfscope}%
\begin{pgfscope}%
\pgfsetbuttcap%
\pgfsetroundjoin%
\definecolor{currentfill}{rgb}{0.000000,0.000000,0.000000}%
\pgfsetfillcolor{currentfill}%
\pgfsetlinewidth{0.803000pt}%
\definecolor{currentstroke}{rgb}{0.000000,0.000000,0.000000}%
\pgfsetstrokecolor{currentstroke}%
\pgfsetdash{}{0pt}%
\pgfsys@defobject{currentmarker}{\pgfqpoint{-0.048611in}{0.000000in}}{\pgfqpoint{-0.000000in}{0.000000in}}{%
\pgfpathmoveto{\pgfqpoint{-0.000000in}{0.000000in}}%
\pgfpathlineto{\pgfqpoint{-0.048611in}{0.000000in}}%
\pgfusepath{stroke,fill}%
}%
\begin{pgfscope}%
\pgfsys@transformshift{0.580783in}{0.866555in}%
\pgfsys@useobject{currentmarker}{}%
\end{pgfscope}%
\end{pgfscope}%
\begin{pgfscope}%
\definecolor{textcolor}{rgb}{0.000000,0.000000,0.000000}%
\pgfsetstrokecolor{textcolor}%
\pgfsetfillcolor{textcolor}%
\pgftext[x=0.233333in, y=0.797111in, left, base]{\color{textcolor}\sffamily\fontsize{14.000000}{16.800000}\selectfont \(\displaystyle {1.0}\)}%
\end{pgfscope}%
\begin{pgfscope}%
\pgfpathrectangle{\pgfqpoint{0.580783in}{0.530555in}}{\pgfqpoint{4.960000in}{3.696000in}}%
\pgfusepath{clip}%
\pgfsetrectcap%
\pgfsetroundjoin%
\pgfsetlinewidth{0.803000pt}%
\definecolor{currentstroke}{rgb}{0.690196,0.690196,0.690196}%
\pgfsetstrokecolor{currentstroke}%
\pgfsetdash{}{0pt}%
\pgfpathmoveto{\pgfqpoint{0.580783in}{1.538555in}}%
\pgfpathlineto{\pgfqpoint{5.540783in}{1.538555in}}%
\pgfusepath{stroke}%
\end{pgfscope}%
\begin{pgfscope}%
\pgfsetbuttcap%
\pgfsetroundjoin%
\definecolor{currentfill}{rgb}{0.000000,0.000000,0.000000}%
\pgfsetfillcolor{currentfill}%
\pgfsetlinewidth{0.803000pt}%
\definecolor{currentstroke}{rgb}{0.000000,0.000000,0.000000}%
\pgfsetstrokecolor{currentstroke}%
\pgfsetdash{}{0pt}%
\pgfsys@defobject{currentmarker}{\pgfqpoint{-0.048611in}{0.000000in}}{\pgfqpoint{-0.000000in}{0.000000in}}{%
\pgfpathmoveto{\pgfqpoint{-0.000000in}{0.000000in}}%
\pgfpathlineto{\pgfqpoint{-0.048611in}{0.000000in}}%
\pgfusepath{stroke,fill}%
}%
\begin{pgfscope}%
\pgfsys@transformshift{0.580783in}{1.538555in}%
\pgfsys@useobject{currentmarker}{}%
\end{pgfscope}%
\end{pgfscope}%
\begin{pgfscope}%
\definecolor{textcolor}{rgb}{0.000000,0.000000,0.000000}%
\pgfsetstrokecolor{textcolor}%
\pgfsetfillcolor{textcolor}%
\pgftext[x=0.233333in, y=1.469111in, left, base]{\color{textcolor}\sffamily\fontsize{14.000000}{16.800000}\selectfont \(\displaystyle {1.2}\)}%
\end{pgfscope}%
\begin{pgfscope}%
\pgfpathrectangle{\pgfqpoint{0.580783in}{0.530555in}}{\pgfqpoint{4.960000in}{3.696000in}}%
\pgfusepath{clip}%
\pgfsetrectcap%
\pgfsetroundjoin%
\pgfsetlinewidth{0.803000pt}%
\definecolor{currentstroke}{rgb}{0.690196,0.690196,0.690196}%
\pgfsetstrokecolor{currentstroke}%
\pgfsetdash{}{0pt}%
\pgfpathmoveto{\pgfqpoint{0.580783in}{2.210555in}}%
\pgfpathlineto{\pgfqpoint{5.540783in}{2.210555in}}%
\pgfusepath{stroke}%
\end{pgfscope}%
\begin{pgfscope}%
\pgfsetbuttcap%
\pgfsetroundjoin%
\definecolor{currentfill}{rgb}{0.000000,0.000000,0.000000}%
\pgfsetfillcolor{currentfill}%
\pgfsetlinewidth{0.803000pt}%
\definecolor{currentstroke}{rgb}{0.000000,0.000000,0.000000}%
\pgfsetstrokecolor{currentstroke}%
\pgfsetdash{}{0pt}%
\pgfsys@defobject{currentmarker}{\pgfqpoint{-0.048611in}{0.000000in}}{\pgfqpoint{-0.000000in}{0.000000in}}{%
\pgfpathmoveto{\pgfqpoint{-0.000000in}{0.000000in}}%
\pgfpathlineto{\pgfqpoint{-0.048611in}{0.000000in}}%
\pgfusepath{stroke,fill}%
}%
\begin{pgfscope}%
\pgfsys@transformshift{0.580783in}{2.210555in}%
\pgfsys@useobject{currentmarker}{}%
\end{pgfscope}%
\end{pgfscope}%
\begin{pgfscope}%
\definecolor{textcolor}{rgb}{0.000000,0.000000,0.000000}%
\pgfsetstrokecolor{textcolor}%
\pgfsetfillcolor{textcolor}%
\pgftext[x=0.233333in, y=2.141111in, left, base]{\color{textcolor}\sffamily\fontsize{14.000000}{16.800000}\selectfont \(\displaystyle {1.4}\)}%
\end{pgfscope}%
\begin{pgfscope}%
\pgfpathrectangle{\pgfqpoint{0.580783in}{0.530555in}}{\pgfqpoint{4.960000in}{3.696000in}}%
\pgfusepath{clip}%
\pgfsetrectcap%
\pgfsetroundjoin%
\pgfsetlinewidth{0.803000pt}%
\definecolor{currentstroke}{rgb}{0.690196,0.690196,0.690196}%
\pgfsetstrokecolor{currentstroke}%
\pgfsetdash{}{0pt}%
\pgfpathmoveto{\pgfqpoint{0.580783in}{2.882555in}}%
\pgfpathlineto{\pgfqpoint{5.540783in}{2.882555in}}%
\pgfusepath{stroke}%
\end{pgfscope}%
\begin{pgfscope}%
\pgfsetbuttcap%
\pgfsetroundjoin%
\definecolor{currentfill}{rgb}{0.000000,0.000000,0.000000}%
\pgfsetfillcolor{currentfill}%
\pgfsetlinewidth{0.803000pt}%
\definecolor{currentstroke}{rgb}{0.000000,0.000000,0.000000}%
\pgfsetstrokecolor{currentstroke}%
\pgfsetdash{}{0pt}%
\pgfsys@defobject{currentmarker}{\pgfqpoint{-0.048611in}{0.000000in}}{\pgfqpoint{-0.000000in}{0.000000in}}{%
\pgfpathmoveto{\pgfqpoint{-0.000000in}{0.000000in}}%
\pgfpathlineto{\pgfqpoint{-0.048611in}{0.000000in}}%
\pgfusepath{stroke,fill}%
}%
\begin{pgfscope}%
\pgfsys@transformshift{0.580783in}{2.882555in}%
\pgfsys@useobject{currentmarker}{}%
\end{pgfscope}%
\end{pgfscope}%
\begin{pgfscope}%
\definecolor{textcolor}{rgb}{0.000000,0.000000,0.000000}%
\pgfsetstrokecolor{textcolor}%
\pgfsetfillcolor{textcolor}%
\pgftext[x=0.233333in, y=2.813111in, left, base]{\color{textcolor}\sffamily\fontsize{14.000000}{16.800000}\selectfont \(\displaystyle {1.6}\)}%
\end{pgfscope}%
\begin{pgfscope}%
\pgfpathrectangle{\pgfqpoint{0.580783in}{0.530555in}}{\pgfqpoint{4.960000in}{3.696000in}}%
\pgfusepath{clip}%
\pgfsetrectcap%
\pgfsetroundjoin%
\pgfsetlinewidth{0.803000pt}%
\definecolor{currentstroke}{rgb}{0.690196,0.690196,0.690196}%
\pgfsetstrokecolor{currentstroke}%
\pgfsetdash{}{0pt}%
\pgfpathmoveto{\pgfqpoint{0.580783in}{3.554555in}}%
\pgfpathlineto{\pgfqpoint{5.540783in}{3.554555in}}%
\pgfusepath{stroke}%
\end{pgfscope}%
\begin{pgfscope}%
\pgfsetbuttcap%
\pgfsetroundjoin%
\definecolor{currentfill}{rgb}{0.000000,0.000000,0.000000}%
\pgfsetfillcolor{currentfill}%
\pgfsetlinewidth{0.803000pt}%
\definecolor{currentstroke}{rgb}{0.000000,0.000000,0.000000}%
\pgfsetstrokecolor{currentstroke}%
\pgfsetdash{}{0pt}%
\pgfsys@defobject{currentmarker}{\pgfqpoint{-0.048611in}{0.000000in}}{\pgfqpoint{-0.000000in}{0.000000in}}{%
\pgfpathmoveto{\pgfqpoint{-0.000000in}{0.000000in}}%
\pgfpathlineto{\pgfqpoint{-0.048611in}{0.000000in}}%
\pgfusepath{stroke,fill}%
}%
\begin{pgfscope}%
\pgfsys@transformshift{0.580783in}{3.554555in}%
\pgfsys@useobject{currentmarker}{}%
\end{pgfscope}%
\end{pgfscope}%
\begin{pgfscope}%
\definecolor{textcolor}{rgb}{0.000000,0.000000,0.000000}%
\pgfsetstrokecolor{textcolor}%
\pgfsetfillcolor{textcolor}%
\pgftext[x=0.233333in, y=3.485111in, left, base]{\color{textcolor}\sffamily\fontsize{14.000000}{16.800000}\selectfont \(\displaystyle {1.8}\)}%
\end{pgfscope}%
\begin{pgfscope}%
\pgfpathrectangle{\pgfqpoint{0.580783in}{0.530555in}}{\pgfqpoint{4.960000in}{3.696000in}}%
\pgfusepath{clip}%
\pgfsetrectcap%
\pgfsetroundjoin%
\pgfsetlinewidth{0.803000pt}%
\definecolor{currentstroke}{rgb}{0.690196,0.690196,0.690196}%
\pgfsetstrokecolor{currentstroke}%
\pgfsetdash{}{0pt}%
\pgfpathmoveto{\pgfqpoint{0.580783in}{4.226555in}}%
\pgfpathlineto{\pgfqpoint{5.540783in}{4.226555in}}%
\pgfusepath{stroke}%
\end{pgfscope}%
\begin{pgfscope}%
\pgfsetbuttcap%
\pgfsetroundjoin%
\definecolor{currentfill}{rgb}{0.000000,0.000000,0.000000}%
\pgfsetfillcolor{currentfill}%
\pgfsetlinewidth{0.803000pt}%
\definecolor{currentstroke}{rgb}{0.000000,0.000000,0.000000}%
\pgfsetstrokecolor{currentstroke}%
\pgfsetdash{}{0pt}%
\pgfsys@defobject{currentmarker}{\pgfqpoint{-0.048611in}{0.000000in}}{\pgfqpoint{-0.000000in}{0.000000in}}{%
\pgfpathmoveto{\pgfqpoint{-0.000000in}{0.000000in}}%
\pgfpathlineto{\pgfqpoint{-0.048611in}{0.000000in}}%
\pgfusepath{stroke,fill}%
}%
\begin{pgfscope}%
\pgfsys@transformshift{0.580783in}{4.226555in}%
\pgfsys@useobject{currentmarker}{}%
\end{pgfscope}%
\end{pgfscope}%
\begin{pgfscope}%
\definecolor{textcolor}{rgb}{0.000000,0.000000,0.000000}%
\pgfsetstrokecolor{textcolor}%
\pgfsetfillcolor{textcolor}%
\pgftext[x=0.233333in, y=4.157111in, left, base]{\color{textcolor}\sffamily\fontsize{14.000000}{16.800000}\selectfont \(\displaystyle {2.0}\)}%
\end{pgfscope}%
\begin{pgfscope}%
\definecolor{textcolor}{rgb}{0.000000,0.000000,0.000000}%
\pgfsetstrokecolor{textcolor}%
\pgfsetfillcolor{textcolor}%
\pgftext[x=0.177777in,y=2.378555in,,bottom,rotate=90.000000]{\color{textcolor}\sffamily\fontsize{14.000000}{16.800000}\selectfont Price of Anarchy}%
\end{pgfscope}%
\begin{pgfscope}%
\pgfpathrectangle{\pgfqpoint{0.580783in}{0.530555in}}{\pgfqpoint{4.960000in}{3.696000in}}%
\pgfusepath{clip}%
\pgfsetrectcap%
\pgfsetroundjoin%
\pgfsetlinewidth{1.505625pt}%
\definecolor{currentstroke}{rgb}{0.121569,0.466667,0.705882}%
\pgfsetstrokecolor{currentstroke}%
\pgfsetdash{}{0pt}%
\pgfpathmoveto{\pgfqpoint{0.806238in}{1.794119in}}%
\pgfpathlineto{\pgfqpoint{0.833320in}{1.799693in}}%
\pgfpathlineto{\pgfqpoint{0.950673in}{1.818221in}}%
\pgfpathlineto{\pgfqpoint{0.959700in}{1.819182in}}%
\pgfpathlineto{\pgfqpoint{0.968728in}{1.843883in}}%
\pgfpathlineto{\pgfqpoint{0.977755in}{1.862073in}}%
\pgfpathlineto{\pgfqpoint{0.986782in}{1.875150in}}%
\pgfpathlineto{\pgfqpoint{0.995809in}{1.882276in}}%
\pgfpathlineto{\pgfqpoint{1.022891in}{1.893404in}}%
\pgfpathlineto{\pgfqpoint{1.040945in}{1.899857in}}%
\pgfpathlineto{\pgfqpoint{1.059000in}{1.905101in}}%
\pgfpathlineto{\pgfqpoint{1.140245in}{1.924899in}}%
\pgfpathlineto{\pgfqpoint{1.203435in}{1.937054in}}%
\pgfpathlineto{\pgfqpoint{1.212462in}{1.951180in}}%
\pgfpathlineto{\pgfqpoint{1.221489in}{1.980184in}}%
\pgfpathlineto{\pgfqpoint{1.230517in}{1.996909in}}%
\pgfpathlineto{\pgfqpoint{1.239544in}{2.011110in}}%
\pgfpathlineto{\pgfqpoint{1.248571in}{2.019908in}}%
\pgfpathlineto{\pgfqpoint{1.266626in}{2.028566in}}%
\pgfpathlineto{\pgfqpoint{1.302734in}{2.042524in}}%
\pgfpathlineto{\pgfqpoint{1.374952in}{2.085425in}}%
\pgfpathlineto{\pgfqpoint{1.402034in}{2.099465in}}%
\pgfpathlineto{\pgfqpoint{1.411061in}{2.110348in}}%
\pgfpathlineto{\pgfqpoint{1.420088in}{2.152259in}}%
\pgfpathlineto{\pgfqpoint{1.429115in}{2.175832in}}%
\pgfpathlineto{\pgfqpoint{1.438143in}{2.194254in}}%
\pgfpathlineto{\pgfqpoint{1.447170in}{2.209712in}}%
\pgfpathlineto{\pgfqpoint{1.456197in}{2.236437in}}%
\pgfpathlineto{\pgfqpoint{1.465224in}{2.281375in}}%
\pgfpathlineto{\pgfqpoint{1.474251in}{2.307367in}}%
\pgfpathlineto{\pgfqpoint{1.483279in}{2.329479in}}%
\pgfpathlineto{\pgfqpoint{1.501333in}{2.349034in}}%
\pgfpathlineto{\pgfqpoint{1.519387in}{2.366198in}}%
\pgfpathlineto{\pgfqpoint{1.537442in}{2.381770in}}%
\pgfpathlineto{\pgfqpoint{1.555496in}{2.396569in}}%
\pgfpathlineto{\pgfqpoint{1.564523in}{2.425375in}}%
\pgfpathlineto{\pgfqpoint{1.573551in}{2.465666in}}%
\pgfpathlineto{\pgfqpoint{1.591605in}{2.513087in}}%
\pgfpathlineto{\pgfqpoint{1.609659in}{2.550342in}}%
\pgfpathlineto{\pgfqpoint{1.618687in}{2.565074in}}%
\pgfpathlineto{\pgfqpoint{1.627714in}{2.615250in}}%
\pgfpathlineto{\pgfqpoint{1.636741in}{2.648721in}}%
\pgfpathlineto{\pgfqpoint{1.645768in}{2.675115in}}%
\pgfpathlineto{\pgfqpoint{1.654796in}{2.697953in}}%
\pgfpathlineto{\pgfqpoint{1.663823in}{2.717224in}}%
\pgfpathlineto{\pgfqpoint{1.672850in}{2.763047in}}%
\pgfpathlineto{\pgfqpoint{1.681877in}{2.802483in}}%
\pgfpathlineto{\pgfqpoint{1.690904in}{2.831169in}}%
\pgfpathlineto{\pgfqpoint{1.708959in}{2.875022in}}%
\pgfpathlineto{\pgfqpoint{1.717986in}{2.892878in}}%
\pgfpathlineto{\pgfqpoint{1.727013in}{2.956510in}}%
\pgfpathlineto{\pgfqpoint{1.736040in}{2.994616in}}%
\pgfpathlineto{\pgfqpoint{1.745068in}{3.064866in}}%
\pgfpathlineto{\pgfqpoint{1.754095in}{3.104427in}}%
\pgfpathlineto{\pgfqpoint{1.763122in}{3.136747in}}%
\pgfpathlineto{\pgfqpoint{1.772149in}{3.222673in}}%
\pgfpathlineto{\pgfqpoint{1.781176in}{3.286401in}}%
\pgfpathlineto{\pgfqpoint{1.790204in}{3.328613in}}%
\pgfpathlineto{\pgfqpoint{1.799231in}{3.357950in}}%
\pgfpathlineto{\pgfqpoint{1.808258in}{3.381718in}}%
\pgfpathlineto{\pgfqpoint{1.817285in}{3.401446in}}%
\pgfpathlineto{\pgfqpoint{1.826312in}{3.414656in}}%
\pgfpathlineto{\pgfqpoint{1.844367in}{3.434945in}}%
\pgfpathlineto{\pgfqpoint{1.853394in}{3.443040in}}%
\pgfpathlineto{\pgfqpoint{1.862421in}{3.447738in}}%
\pgfpathlineto{\pgfqpoint{1.871449in}{3.450133in}}%
\pgfpathlineto{\pgfqpoint{1.871542in}{4.236555in}}%
\pgfpathlineto{\pgfqpoint{1.871542in}{4.236555in}}%
\pgfusepath{stroke}%
\end{pgfscope}%
\begin{pgfscope}%
\pgfpathrectangle{\pgfqpoint{0.580783in}{0.530555in}}{\pgfqpoint{4.960000in}{3.696000in}}%
\pgfusepath{clip}%
\pgfsetbuttcap%
\pgfsetmiterjoin%
\definecolor{currentfill}{rgb}{0.121569,0.466667,0.705882}%
\pgfsetfillcolor{currentfill}%
\pgfsetlinewidth{1.003750pt}%
\definecolor{currentstroke}{rgb}{0.121569,0.466667,0.705882}%
\pgfsetstrokecolor{currentstroke}%
\pgfsetdash{}{0pt}%
\pgfsys@defobject{currentmarker}{\pgfqpoint{-0.041667in}{-0.041667in}}{\pgfqpoint{0.041667in}{0.041667in}}{%
\pgfpathmoveto{\pgfqpoint{0.000000in}{0.041667in}}%
\pgfpathlineto{\pgfqpoint{-0.041667in}{-0.041667in}}%
\pgfpathlineto{\pgfqpoint{0.041667in}{-0.041667in}}%
\pgfpathlineto{\pgfqpoint{0.000000in}{0.041667in}}%
\pgfpathclose%
\pgfusepath{stroke,fill}%
}%
\begin{pgfscope}%
\pgfsys@transformshift{0.806238in}{1.794119in}%
\pgfsys@useobject{currentmarker}{}%
\end{pgfscope}%
\begin{pgfscope}%
\pgfsys@transformshift{1.618687in}{2.565074in}%
\pgfsys@useobject{currentmarker}{}%
\end{pgfscope}%
\begin{pgfscope}%
\pgfsys@transformshift{1.871449in}{3.450133in}%
\pgfsys@useobject{currentmarker}{}%
\end{pgfscope}%
\begin{pgfscope}%
\pgfsys@transformshift{1.880476in}{79.694417in}%
\pgfsys@useobject{currentmarker}{}%
\end{pgfscope}%
\begin{pgfscope}%
\pgfsys@transformshift{2.539462in}{79.340280in}%
\pgfsys@useobject{currentmarker}{}%
\end{pgfscope}%
\begin{pgfscope}%
\pgfsys@transformshift{3.640782in}{78.756506in}%
\pgfsys@useobject{currentmarker}{}%
\end{pgfscope}%
\begin{pgfscope}%
\pgfsys@transformshift{4.733074in}{78.187301in}%
\pgfsys@useobject{currentmarker}{}%
\end{pgfscope}%
\begin{pgfscope}%
\pgfsys@transformshift{5.834393in}{77.623042in}%
\pgfsys@useobject{currentmarker}{}%
\end{pgfscope}%
\begin{pgfscope}%
\pgfsys@transformshift{6.935713in}{77.068265in}%
\pgfsys@useobject{currentmarker}{}%
\end{pgfscope}%
\begin{pgfscope}%
\pgfsys@transformshift{8.046060in}{76.518333in}%
\pgfsys@useobject{currentmarker}{}%
\end{pgfscope}%
\begin{pgfscope}%
\pgfsys@transformshift{9.156406in}{75.977629in}%
\pgfsys@useobject{currentmarker}{}%
\end{pgfscope}%
\end{pgfscope}%
\begin{pgfscope}%
\pgfpathrectangle{\pgfqpoint{0.580783in}{0.530555in}}{\pgfqpoint{4.960000in}{3.696000in}}%
\pgfusepath{clip}%
\pgfsetrectcap%
\pgfsetroundjoin%
\pgfsetlinewidth{1.505625pt}%
\definecolor{currentstroke}{rgb}{1.000000,0.498039,0.054902}%
\pgfsetstrokecolor{currentstroke}%
\pgfsetdash{}{0pt}%
\pgfpathmoveto{\pgfqpoint{0.806238in}{0.868034in}}%
\pgfpathlineto{\pgfqpoint{1.456197in}{0.868784in}}%
\pgfpathlineto{\pgfqpoint{1.681877in}{0.870788in}}%
\pgfpathlineto{\pgfqpoint{1.717986in}{0.871758in}}%
\pgfpathlineto{\pgfqpoint{1.763122in}{0.873724in}}%
\pgfpathlineto{\pgfqpoint{1.880476in}{0.876918in}}%
\pgfpathlineto{\pgfqpoint{2.006857in}{0.879663in}}%
\pgfpathlineto{\pgfqpoint{2.178374in}{0.886042in}}%
\pgfpathlineto{\pgfqpoint{2.250591in}{0.889063in}}%
\pgfpathlineto{\pgfqpoint{2.268646in}{0.889683in}}%
\pgfpathlineto{\pgfqpoint{2.286700in}{0.893037in}}%
\pgfpathlineto{\pgfqpoint{2.340863in}{0.898548in}}%
\pgfpathlineto{\pgfqpoint{2.413081in}{0.904352in}}%
\pgfpathlineto{\pgfqpoint{2.584598in}{0.916204in}}%
\pgfpathlineto{\pgfqpoint{2.647789in}{0.919764in}}%
\pgfpathlineto{\pgfqpoint{2.855414in}{0.928087in}}%
\pgfpathlineto{\pgfqpoint{2.882496in}{0.934495in}}%
\pgfpathlineto{\pgfqpoint{2.918605in}{0.939736in}}%
\pgfpathlineto{\pgfqpoint{3.026931in}{0.949753in}}%
\pgfpathlineto{\pgfqpoint{3.243584in}{0.963693in}}%
\pgfpathlineto{\pgfqpoint{3.342884in}{0.966095in}}%
\pgfpathlineto{\pgfqpoint{3.378992in}{0.966650in}}%
\pgfpathlineto{\pgfqpoint{3.406074in}{0.975225in}}%
\pgfpathlineto{\pgfqpoint{3.460237in}{0.983964in}}%
\pgfpathlineto{\pgfqpoint{3.541482in}{0.993038in}}%
\pgfpathlineto{\pgfqpoint{3.577591in}{0.996010in}}%
\pgfpathlineto{\pgfqpoint{3.604673in}{0.998456in}}%
\pgfpathlineto{\pgfqpoint{3.649809in}{1.001227in}}%
\pgfpathlineto{\pgfqpoint{3.758135in}{1.004497in}}%
\pgfpathlineto{\pgfqpoint{3.812298in}{1.006040in}}%
\pgfpathlineto{\pgfqpoint{3.830353in}{1.014185in}}%
\pgfpathlineto{\pgfqpoint{3.857435in}{1.021396in}}%
\pgfpathlineto{\pgfqpoint{3.911598in}{1.030308in}}%
\pgfpathlineto{\pgfqpoint{3.938679in}{1.033847in}}%
\pgfpathlineto{\pgfqpoint{4.028952in}{1.043674in}}%
\pgfpathlineto{\pgfqpoint{4.146305in}{1.047865in}}%
\pgfpathlineto{\pgfqpoint{4.155332in}{1.051642in}}%
\pgfpathlineto{\pgfqpoint{4.164360in}{1.057271in}}%
\pgfpathlineto{\pgfqpoint{4.182414in}{1.063824in}}%
\pgfpathlineto{\pgfqpoint{4.254632in}{1.079025in}}%
\pgfpathlineto{\pgfqpoint{4.299768in}{1.085614in}}%
\pgfpathlineto{\pgfqpoint{4.326849in}{1.090220in}}%
\pgfpathlineto{\pgfqpoint{4.408094in}{1.093974in}}%
\pgfpathlineto{\pgfqpoint{4.417121in}{1.102417in}}%
\pgfpathlineto{\pgfqpoint{4.426149in}{1.107253in}}%
\pgfpathlineto{\pgfqpoint{4.453230in}{1.116938in}}%
\pgfpathlineto{\pgfqpoint{4.507394in}{1.129023in}}%
\pgfpathlineto{\pgfqpoint{4.552530in}{1.137760in}}%
\pgfpathlineto{\pgfqpoint{4.570584in}{1.140407in}}%
\pgfpathlineto{\pgfqpoint{4.588638in}{1.142040in}}%
\pgfpathlineto{\pgfqpoint{4.597666in}{1.142475in}}%
\pgfpathlineto{\pgfqpoint{4.606693in}{1.144696in}}%
\pgfpathlineto{\pgfqpoint{4.615720in}{1.154625in}}%
\pgfpathlineto{\pgfqpoint{4.633775in}{1.163001in}}%
\pgfpathlineto{\pgfqpoint{4.660856in}{1.171874in}}%
\pgfpathlineto{\pgfqpoint{4.696965in}{1.181247in}}%
\pgfpathlineto{\pgfqpoint{4.724047in}{1.187588in}}%
\pgfpathlineto{\pgfqpoint{4.751128in}{1.195812in}}%
\pgfpathlineto{\pgfqpoint{4.760155in}{1.198119in}}%
\pgfpathlineto{\pgfqpoint{4.769183in}{1.209516in}}%
\pgfpathlineto{\pgfqpoint{4.787237in}{1.221991in}}%
\pgfpathlineto{\pgfqpoint{4.814319in}{1.234748in}}%
\pgfpathlineto{\pgfqpoint{4.841400in}{1.244860in}}%
\pgfpathlineto{\pgfqpoint{4.868482in}{1.254484in}}%
\pgfpathlineto{\pgfqpoint{4.895564in}{1.262778in}}%
\pgfpathlineto{\pgfqpoint{4.913618in}{1.291184in}}%
\pgfpathlineto{\pgfqpoint{4.922645in}{1.307917in}}%
\pgfpathlineto{\pgfqpoint{4.931672in}{1.317042in}}%
\pgfpathlineto{\pgfqpoint{4.958754in}{1.334233in}}%
\pgfpathlineto{\pgfqpoint{4.985836in}{1.348719in}}%
\pgfpathlineto{\pgfqpoint{5.021944in}{1.364218in}}%
\pgfpathlineto{\pgfqpoint{5.030972in}{1.380335in}}%
\pgfpathlineto{\pgfqpoint{5.039999in}{1.389232in}}%
\pgfpathlineto{\pgfqpoint{5.058053in}{1.401911in}}%
\pgfpathlineto{\pgfqpoint{5.076108in}{1.412865in}}%
\pgfpathlineto{\pgfqpoint{5.130271in}{1.439026in}}%
\pgfpathlineto{\pgfqpoint{5.139298in}{1.442251in}}%
\pgfpathlineto{\pgfqpoint{5.148325in}{1.450230in}}%
\pgfpathlineto{\pgfqpoint{5.166380in}{1.497132in}}%
\pgfpathlineto{\pgfqpoint{5.175407in}{1.514015in}}%
\pgfpathlineto{\pgfqpoint{5.184434in}{1.525992in}}%
\pgfpathlineto{\pgfqpoint{5.202489in}{1.541581in}}%
\pgfpathlineto{\pgfqpoint{5.220543in}{1.555165in}}%
\pgfpathlineto{\pgfqpoint{5.247625in}{1.574258in}}%
\pgfpathlineto{\pgfqpoint{5.265679in}{1.581444in}}%
\pgfpathlineto{\pgfqpoint{5.292761in}{1.622854in}}%
\pgfpathlineto{\pgfqpoint{5.310815in}{1.642452in}}%
\pgfpathlineto{\pgfqpoint{5.328870in}{1.657463in}}%
\pgfpathlineto{\pgfqpoint{5.355951in}{1.724420in}}%
\pgfpathlineto{\pgfqpoint{5.364978in}{1.741466in}}%
\pgfpathlineto{\pgfqpoint{5.374006in}{1.752453in}}%
\pgfpathlineto{\pgfqpoint{5.392060in}{1.765060in}}%
\pgfpathlineto{\pgfqpoint{5.401087in}{1.794780in}}%
\pgfpathlineto{\pgfqpoint{5.410114in}{1.813408in}}%
\pgfpathlineto{\pgfqpoint{5.428169in}{1.841138in}}%
\pgfpathlineto{\pgfqpoint{5.437196in}{1.851716in}}%
\pgfpathlineto{\pgfqpoint{5.446223in}{1.864242in}}%
\pgfpathlineto{\pgfqpoint{5.464278in}{1.884808in}}%
\pgfpathlineto{\pgfqpoint{5.482332in}{1.949900in}}%
\pgfpathlineto{\pgfqpoint{5.500387in}{1.985802in}}%
\pgfpathlineto{\pgfqpoint{5.518441in}{2.059138in}}%
\pgfpathlineto{\pgfqpoint{5.527468in}{2.083694in}}%
\pgfpathlineto{\pgfqpoint{5.536495in}{2.102705in}}%
\pgfpathlineto{\pgfqpoint{5.550783in}{2.127000in}}%
\pgfpathlineto{\pgfqpoint{5.550783in}{2.127000in}}%
\pgfusepath{stroke}%
\end{pgfscope}%
\begin{pgfscope}%
\pgfpathrectangle{\pgfqpoint{0.580783in}{0.530555in}}{\pgfqpoint{4.960000in}{3.696000in}}%
\pgfusepath{clip}%
\pgfsetbuttcap%
\pgfsetmiterjoin%
\definecolor{currentfill}{rgb}{1.000000,0.498039,0.054902}%
\pgfsetfillcolor{currentfill}%
\pgfsetlinewidth{1.003750pt}%
\definecolor{currentstroke}{rgb}{1.000000,0.498039,0.054902}%
\pgfsetstrokecolor{currentstroke}%
\pgfsetdash{}{0pt}%
\pgfsys@defobject{currentmarker}{\pgfqpoint{-0.035355in}{-0.058926in}}{\pgfqpoint{0.035355in}{0.058926in}}{%
\pgfpathmoveto{\pgfqpoint{-0.000000in}{-0.058926in}}%
\pgfpathlineto{\pgfqpoint{0.035355in}{0.000000in}}%
\pgfpathlineto{\pgfqpoint{0.000000in}{0.058926in}}%
\pgfpathlineto{\pgfqpoint{-0.035355in}{0.000000in}}%
\pgfpathlineto{\pgfqpoint{-0.000000in}{-0.058926in}}%
\pgfpathclose%
\pgfusepath{stroke,fill}%
}%
\begin{pgfscope}%
\pgfsys@transformshift{0.806238in}{0.868034in}%
\pgfsys@useobject{currentmarker}{}%
\end{pgfscope}%
\begin{pgfscope}%
\pgfsys@transformshift{2.042966in}{0.880500in}%
\pgfsys@useobject{currentmarker}{}%
\end{pgfscope}%
\begin{pgfscope}%
\pgfsys@transformshift{3.279693in}{0.964596in}%
\pgfsys@useobject{currentmarker}{}%
\end{pgfscope}%
\begin{pgfscope}%
\pgfsys@transformshift{4.489339in}{1.125397in}%
\pgfsys@useobject{currentmarker}{}%
\end{pgfscope}%
\begin{pgfscope}%
\pgfsys@transformshift{5.419142in}{1.827750in}%
\pgfsys@useobject{currentmarker}{}%
\end{pgfscope}%
\begin{pgfscope}%
\pgfsys@transformshift{5.726067in}{3.052806in}%
\pgfsys@useobject{currentmarker}{}%
\end{pgfscope}%
\begin{pgfscope}%
\pgfsys@transformshift{6.673924in}{3.393095in}%
\pgfsys@useobject{currentmarker}{}%
\end{pgfscope}%
\begin{pgfscope}%
\pgfsys@transformshift{7.910651in}{3.364291in}%
\pgfsys@useobject{currentmarker}{}%
\end{pgfscope}%
\begin{pgfscope}%
\pgfsys@transformshift{9.147379in}{3.325852in}%
\pgfsys@useobject{currentmarker}{}%
\end{pgfscope}%
\end{pgfscope}%
\begin{pgfscope}%
\pgfsetrectcap%
\pgfsetmiterjoin%
\pgfsetlinewidth{0.803000pt}%
\definecolor{currentstroke}{rgb}{0.000000,0.000000,0.000000}%
\pgfsetstrokecolor{currentstroke}%
\pgfsetdash{}{0pt}%
\pgfpathmoveto{\pgfqpoint{0.580783in}{0.530555in}}%
\pgfpathlineto{\pgfqpoint{0.580783in}{4.226555in}}%
\pgfusepath{stroke}%
\end{pgfscope}%
\begin{pgfscope}%
\pgfsetrectcap%
\pgfsetmiterjoin%
\pgfsetlinewidth{0.803000pt}%
\definecolor{currentstroke}{rgb}{0.000000,0.000000,0.000000}%
\pgfsetstrokecolor{currentstroke}%
\pgfsetdash{}{0pt}%
\pgfpathmoveto{\pgfqpoint{5.540783in}{0.530555in}}%
\pgfpathlineto{\pgfqpoint{5.540783in}{4.226555in}}%
\pgfusepath{stroke}%
\end{pgfscope}%
\begin{pgfscope}%
\pgfsetrectcap%
\pgfsetmiterjoin%
\pgfsetlinewidth{0.803000pt}%
\definecolor{currentstroke}{rgb}{0.000000,0.000000,0.000000}%
\pgfsetstrokecolor{currentstroke}%
\pgfsetdash{}{0pt}%
\pgfpathmoveto{\pgfqpoint{0.580783in}{0.530555in}}%
\pgfpathlineto{\pgfqpoint{5.540783in}{0.530555in}}%
\pgfusepath{stroke}%
\end{pgfscope}%
\begin{pgfscope}%
\pgfsetrectcap%
\pgfsetmiterjoin%
\pgfsetlinewidth{0.803000pt}%
\definecolor{currentstroke}{rgb}{0.000000,0.000000,0.000000}%
\pgfsetstrokecolor{currentstroke}%
\pgfsetdash{}{0pt}%
\pgfpathmoveto{\pgfqpoint{0.580783in}{4.226555in}}%
\pgfpathlineto{\pgfqpoint{5.540783in}{4.226555in}}%
\pgfusepath{stroke}%
\end{pgfscope}%
\begin{pgfscope}%
\pgfsetbuttcap%
\pgfsetmiterjoin%
\definecolor{currentfill}{rgb}{1.000000,1.000000,1.000000}%
\pgfsetfillcolor{currentfill}%
\pgfsetfillopacity{0.800000}%
\pgfsetlinewidth{1.003750pt}%
\definecolor{currentstroke}{rgb}{0.800000,0.800000,0.800000}%
\pgfsetstrokecolor{currentstroke}%
\pgfsetstrokeopacity{0.800000}%
\pgfsetdash{}{0pt}%
\pgfpathmoveto{\pgfqpoint{2.971595in}{3.521000in}}%
\pgfpathlineto{\pgfqpoint{5.404672in}{3.521000in}}%
\pgfpathquadraticcurveto{\pgfqpoint{5.443561in}{3.521000in}}{\pgfqpoint{5.443561in}{3.559889in}}%
\pgfpathlineto{\pgfqpoint{5.443561in}{4.090444in}}%
\pgfpathquadraticcurveto{\pgfqpoint{5.443561in}{4.129333in}}{\pgfqpoint{5.404672in}{4.129333in}}%
\pgfpathlineto{\pgfqpoint{2.971595in}{4.129333in}}%
\pgfpathquadraticcurveto{\pgfqpoint{2.932706in}{4.129333in}}{\pgfqpoint{2.932706in}{4.090444in}}%
\pgfpathlineto{\pgfqpoint{2.932706in}{3.559889in}}%
\pgfpathquadraticcurveto{\pgfqpoint{2.932706in}{3.521000in}}{\pgfqpoint{2.971595in}{3.521000in}}%
\pgfpathlineto{\pgfqpoint{2.971595in}{3.521000in}}%
\pgfpathclose%
\pgfusepath{stroke,fill}%
\end{pgfscope}%
\begin{pgfscope}%
\pgfsetrectcap%
\pgfsetroundjoin%
\pgfsetlinewidth{1.505625pt}%
\definecolor{currentstroke}{rgb}{0.121569,0.466667,0.705882}%
\pgfsetstrokecolor{currentstroke}%
\pgfsetdash{}{0pt}%
\pgfpathmoveto{\pgfqpoint{3.010484in}{3.980722in}}%
\pgfpathlineto{\pgfqpoint{3.204928in}{3.980722in}}%
\pgfpathlineto{\pgfqpoint{3.399373in}{3.980722in}}%
\pgfusepath{stroke}%
\end{pgfscope}%
\begin{pgfscope}%
\pgfsetbuttcap%
\pgfsetmiterjoin%
\definecolor{currentfill}{rgb}{0.121569,0.466667,0.705882}%
\pgfsetfillcolor{currentfill}%
\pgfsetlinewidth{1.003750pt}%
\definecolor{currentstroke}{rgb}{0.121569,0.466667,0.705882}%
\pgfsetstrokecolor{currentstroke}%
\pgfsetdash{}{0pt}%
\pgfsys@defobject{currentmarker}{\pgfqpoint{-0.041667in}{-0.041667in}}{\pgfqpoint{0.041667in}{0.041667in}}{%
\pgfpathmoveto{\pgfqpoint{0.000000in}{0.041667in}}%
\pgfpathlineto{\pgfqpoint{-0.041667in}{-0.041667in}}%
\pgfpathlineto{\pgfqpoint{0.041667in}{-0.041667in}}%
\pgfpathlineto{\pgfqpoint{0.000000in}{0.041667in}}%
\pgfpathclose%
\pgfusepath{stroke,fill}%
}%
\begin{pgfscope}%
\pgfsys@transformshift{3.204928in}{3.980722in}%
\pgfsys@useobject{currentmarker}{}%
\end{pgfscope}%
\end{pgfscope}%
\begin{pgfscope}%
\definecolor{textcolor}{rgb}{0.000000,0.000000,0.000000}%
\pgfsetstrokecolor{textcolor}%
\pgfsetfillcolor{textcolor}%
\pgftext[x=3.554928in,y=3.912666in,left,base]{\color{textcolor}\sffamily\fontsize{14.000000}{16.800000}\selectfont PoA without incentive}%
\end{pgfscope}%
\begin{pgfscope}%
\pgfsetrectcap%
\pgfsetroundjoin%
\pgfsetlinewidth{1.505625pt}%
\definecolor{currentstroke}{rgb}{1.000000,0.498039,0.054902}%
\pgfsetstrokecolor{currentstroke}%
\pgfsetdash{}{0pt}%
\pgfpathmoveto{\pgfqpoint{3.010484in}{3.705722in}}%
\pgfpathlineto{\pgfqpoint{3.204928in}{3.705722in}}%
\pgfpathlineto{\pgfqpoint{3.399373in}{3.705722in}}%
\pgfusepath{stroke}%
\end{pgfscope}%
\begin{pgfscope}%
\pgfsetbuttcap%
\pgfsetmiterjoin%
\definecolor{currentfill}{rgb}{1.000000,0.498039,0.054902}%
\pgfsetfillcolor{currentfill}%
\pgfsetlinewidth{1.003750pt}%
\definecolor{currentstroke}{rgb}{1.000000,0.498039,0.054902}%
\pgfsetstrokecolor{currentstroke}%
\pgfsetdash{}{0pt}%
\pgfsys@defobject{currentmarker}{\pgfqpoint{-0.035355in}{-0.058926in}}{\pgfqpoint{0.035355in}{0.058926in}}{%
\pgfpathmoveto{\pgfqpoint{-0.000000in}{-0.058926in}}%
\pgfpathlineto{\pgfqpoint{0.035355in}{0.000000in}}%
\pgfpathlineto{\pgfqpoint{0.000000in}{0.058926in}}%
\pgfpathlineto{\pgfqpoint{-0.035355in}{0.000000in}}%
\pgfpathlineto{\pgfqpoint{-0.000000in}{-0.058926in}}%
\pgfpathclose%
\pgfusepath{stroke,fill}%
}%
\begin{pgfscope}%
\pgfsys@transformshift{3.204928in}{3.705722in}%
\pgfsys@useobject{currentmarker}{}%
\end{pgfscope}%
\end{pgfscope}%
\begin{pgfscope}%
\definecolor{textcolor}{rgb}{0.000000,0.000000,0.000000}%
\pgfsetstrokecolor{textcolor}%
\pgfsetfillcolor{textcolor}%
\pgftext[x=3.554928in,y=3.637667in,left,base]{\color{textcolor}\sffamily\fontsize{14.000000}{16.800000}\selectfont PoA with incentive}%
\end{pgfscope}%
\end{pgfpicture}%
\makeatother%
\endgroup%